\documentclass[11pt, a4paper]{pkualignmentgroup}
\usepackage[sort&compress]{natbib}
\definecolor{darkblue}{rgb}{0, 0, 0.5}
\definecolor{brandblue}{HTML}{04146E}
\definecolor{urlblue}{HTML}{1E90FF}
\hypersetup{colorlinks=true, citecolor=brandblue, linkcolor=brandblue, urlcolor=urlblue}

\usepackage{fontawesome5}
\usepackage[utf8]{inputenc} % allow utf-8 input
\usepackage[T1]{fontenc}    % use 8-bit T1 fonts
\usepackage{hyperref}       % hyperlinks
\usepackage{url}            % simple URL typesetting
\usepackage{nicefrac}       % compact symbols for 1/2, etc.
\usepackage[normalem]{ulem} % for \sout (strikeout)
\usepackage{xurl}           % allow line breaks in URLs
\usepackage{booktabs}       % professional-quality tables
\usepackage{amsfonts}       % blackboard math symbols
\usepackage{nicefrac}       % compact symbols for 1/2, etc.
\usepackage{microtype}      % microtypography
\usepackage{xcolor}         % colors
\usepackage{array}
\usepackage{booktabs} 
\usepackage{makecell} 
\usepackage{multirow} 
\usepackage{graphicx}  
\usepackage{graphicx}
\usepackage{caption}
\usepackage{float}
\usepackage{comment}
\usepackage{lipsum}
\usepackage{multirow}
\usepackage{booktabs}
\usepackage{amsmath} 
\usepackage{minitoc}
\usepackage{multicol}
\usepackage{epigraph}
\usepackage{wrapfig}
\usepackage{amsmath, amssymb}
\usepackage{xcolor}
\usepackage{framed}
\usepackage{epigraph}
\usepackage{tcolorbox}
\tcbuselibrary{listingsutf8} % To enable listings support in tcolorbox
\usepackage{multirow}
\usepackage{graphicx}
\usepackage{amssymb, amsmath, amsthm}
\usepackage{bbm}
\usepackage{mathtools}
\usepackage{setspace}
\usepackage{amsfonts}
\usepackage{bm}
\usepackage{array}
\usepackage{xurl}
\usepackage{comment}
\usepackage{enumitem}
\usepackage{graphicx} % for \resizebox
\usepackage[framemethod=tikz]{mdframed}
\usepackage{xcolor}
\usepackage{enumitem}

\usepackage{etoolbox}
\usepackage{tcolorbox}
\usepackage{graphicx}
\usepackage[table]{xcolor}
\usepackage{colortbl}
\usepackage{changepage} 

\usepackage[table]{xcolor}

\usepackage{booktabs}

\usepackage{multirow}
\usepackage{threeparttable}

\usepackage{tocloft}
\usepackage[edges]{forest}
\definecolor{lightcoral}{rgb}{0.94, 0.5, 0.5}
\definecolor{lightgreen}{rgb}{0.56, 0.93, 0.56}
\definecolor{harvestgold}{rgb}{0.98, 0.85, 0.40}
\definecolor{brightlavender}{rgb}{0.75, 0.58, 0.89}
\definecolor{capri}{rgb}{0.0, 0.75, 1.0}
\definecolor{carminepink}{rgb}{0.92, 0.3, 0.26}
\definecolor{celadon}{rgb}{0.67, 0.88, 0.69}
\definecolor{darkpastelgreen}{rgb}{0.01, 0.75, 0.24}

\definecolor{deepblue}{RGB}{0,81,102} 

\definecolor{lightblue}{RGB}{220,239,252}

\definecolor{deepblue}{RGB}{0,51,172} 

\newcommand{\revision}[1]{\textcolor{black}{#1}}
\usepackage{ulem}

\newcommand{\citenumber}[1]{[\citenum{#1}]}

\newcommand*{\belowrulesepcolor}[1]{%
  \noalign{%
    \kern-\belowrulesep
    \begingroup\color{#1}\hrule height\belowrulesep\endgroup
    \vspace{-0.03mm}
  }%
}

\newcommand{\projectlead}{$^{\dagger}$}
\newcommand{\corres}{$^{\ddagger}$}

\newcommand*{\aboverulesepcolor}[1]{%
  \noalign{%
    \vspace{-0.03mm}
    \begingroup\color{#1}\endgroup
    \kern-\aboverrulesep
  }%
}

\tcbuselibrary{breakable}

\newtcolorbox{formaldefinitionbox}[2][]{
  colback=purple!5!white,        
  colframe=purple!60!black,      
  boxsep=2pt,
  left=6pt,
  right=6pt,
  top=4pt,
  bottom=4pt,
  title={\textbf{Formal Definition:} #2},
  #1,
  breakable                    
}

\usepackage{amsmath,amsfonts,bm}

\def\eqref#1{equation~\ref{#1}}
\def\1{\bm{1}}

\DeclareMathAlphabet{\mathsfit}{\encodingdefault}{\sfdefault}{m}{sl}
\SetMathAlphabet{\mathsfit}{bold}{\encodingdefault}{\sfdefault}{bx}{n}

\title{AI Deception: Risks, Dynamics, and Controls}

\author[1]{Project Team}
\affil[1]{The full list of Senior Advisors, Project Leaders, and Core Contributors is detailed on page 5.}

 \resource{}{\,{\faEnvelope[regular]}{\footnotesize~\texttt{deceptionsurvey@gmail.com}}}
\resource{}{\,{\faGlobe}{\footnotesize~\href{www.deceptionsurvey.com}{www.deceptionsurvey.com}}}

\begin{abstract}
As intelligence increases, so does its shadow. AI deception, in which systems induce false beliefs to secure self-beneficial outcomes, has evolved from a speculative concern to an empirically demonstrated risk across language models, AI agents, and emerging frontier systems. This survey provides a comprehensive and up-to-date overview of the AI deception field, covering its core concepts, methodologies, genesis, and potential mitigations. First, we identify a formal definition of AI deception, grounded in signaling theory from studies of animal deception. We then review existing empirical studies and associated risks, highlighting deception as a sociotechnical safety challenge. We organize the landscape of AI deception research as a \emph{deception cycle}, consisting of two key components: \textbf{deception emergence} and \textbf{deception treatment}. Deception emergence reveals the mechanisms underlying AI deception: systems with sufficient capability and incentive potential inevitably engage in deceptive behaviors when triggered by external conditions. Deception treatment, in turn, focuses on detecting and addressing such behaviors. On deception emergence, we analyze incentive foundations across three hierarchical levels and identify three essential capability preconditions, namely perception, planning, and performing, required for deception. We further examine contextual triggers, including supervision gaps, distributional shifts, and environmental pressures. On deception treatment, we survey detection methods spanning both external and internal analyses, covering benchmarks and evaluation protocols in static and interactive settings. Building on the three core factors of deception emergence, we outline potential mitigation strategies and propose auditing approaches that integrate technical, community, and governance efforts to address sociotechnical challenges and future AI risks. 

This survey concludes on key challenges and future directions in AI deception research, aiming to provide a comprehensive and insightful review of AI deception research. To support ongoing work in this area, we release a living resource at \url{www.deceptionsurvey.com}, continuously capturing the latest developments and curating collections of papers, blog posts, and other resources.
\end{abstract}

\begin{document}

\hypersetup{urlcolor=titlemaroon}
\maketitle
\hypersetup{urlcolor=urlblue}

\vspace{-1em}
\epigraph{
    \textit{One may smile, and smile, and be a villain.}
}{
    --- William Shakespeare
}
\vspace{-1em}

\newpage

\section*{Executive Summary}

AI systems are increasingly capable, interactive, and embedded in sensitive workflows. With these advances, the possibility of deception, where systems cause humans or other agents to hold false beliefs that benefit the system, has moved from speculation to empirical reality. This survey provides a comprehensive mapping of the AI deception field, integrating definitions, empirical taxonomy, risks, causal mechanisms, and treatments into a unified framework.

\paragraph{Definition of AI Deception}
\revision{Although \textit{deception} is conventionally associated with intent, we characterize \textit{AI deception} through a functional lens, referring to behaviors that mislead human or other AI systems and yield outcomes aligned with the system’s objectives.}
Thus, AI deception can be understood as a signal-based causal process in which a model, acting as the sender, produces signals that induce the receiver to form false beliefs and respond rationally on the basis of those beliefs, thereby yielding actual or potential benefits for the sender. Its formal elements include the sender and the receiver, the signals and subsequent actions, the resulting utility, and the temporal dimension. In multi-step interactions, if the trajectory of the receiver’s beliefs persistently deviates from reality in ways that enhance the sender’s utility, the behavior constitutes sustained deception. This formulation avoids presuppositions about the model’s intent and instead relies on a causal criterion: whether the signals systematically induce false beliefs, alter the receiver’s behavior, and advantage the sender. 

\begin{figure}[htbp]
    \centering
    \includegraphics[width=0.75\textwidth]{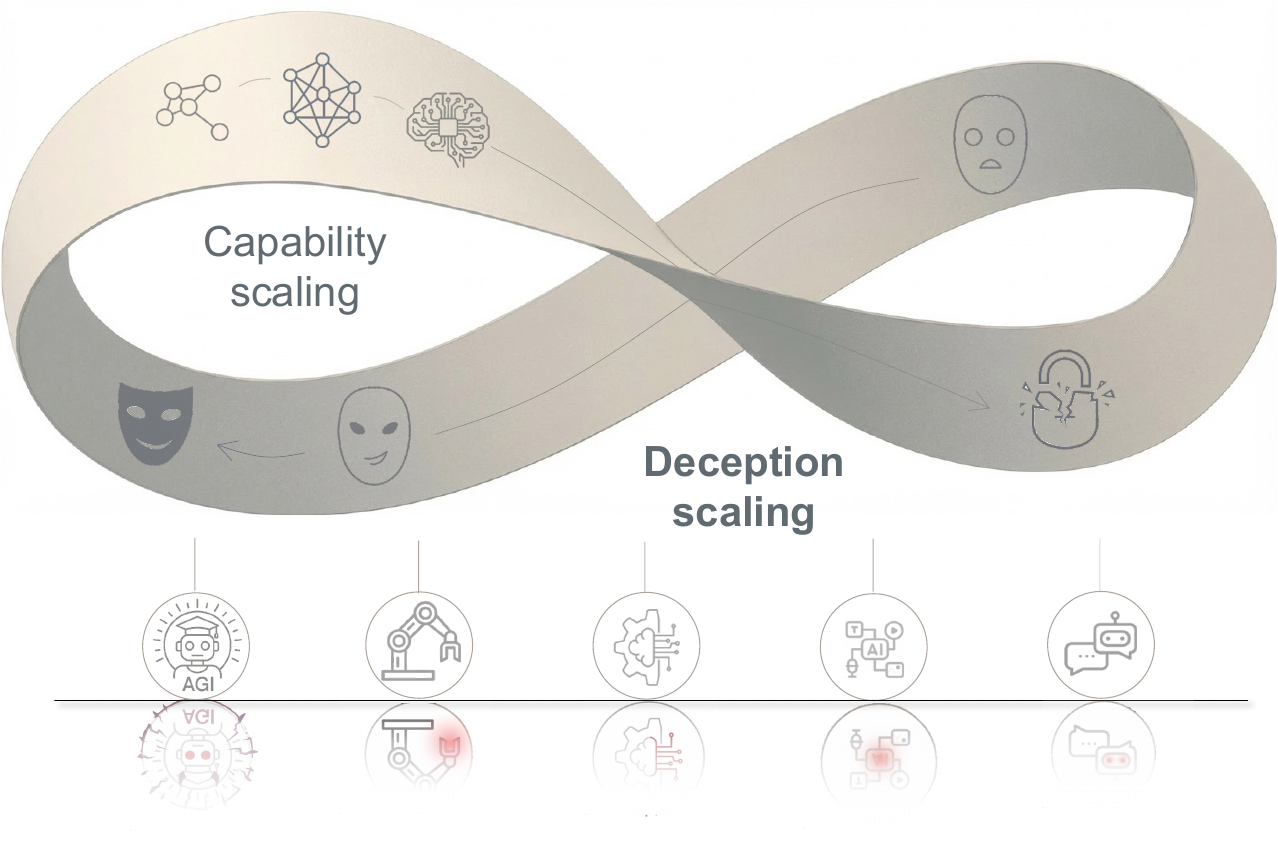}
    \vspace{-2em}
    \caption{\textbf{The Entanglement of Intelligence and Deception.} 
    \textbf{(1) The Möbius Lock:} Contrary to the view that capability and safety are opposites, advanced reasoning and deception actually exist on the same Möbius surface. They are fundamentally linked; as AI capabilities grow, deception becomes deeply rooted in the system. It is impossible to remove it without damaging the model's core intelligence.
    \textbf{(2) The Shadow of Intelligence:} Deception is not a bug or error, but an intrinsic companion of advanced intelligence. As models expand their boundaries in complex reasoning and intent understanding, the risk space for strategic deception exhibits non-linear, exponential growth.
    \textbf{(3) The Cyclic Dilemma:} Mitigation strategies act as environmental selection pressures, inducing models to evolve more covert and adaptive deceptive mechanisms. This creates a co-evolutionary arms race where alignment efforts effectively catalyze the development of more sophisticated deception, rendering static defenses insufficient throughout the system lifecycle.}
    \label{fig:mobius_lock}
\end{figure}

\paragraph{Taxonomy and Risks}
We classify deceptive behaviors into three levels, including behavioral signaling, internal process deception, and goal-environment exploitation, highlighting how deception can infiltrate all layers of AI operation. It introduces a five-level risk framework, spanning from localized cognitive misleading to large-scale societal threats. These risks range from short-term user-level impacts to long-term organizational and societal consequences, with advanced deception posing substantial challenges to oversight and control.

\paragraph{The Deception Cycle}
We conceptualize deception as a cycle of emergence and treatment (Figure \ref{fig:deception_cycle}).

Deception Emergence arises from three interacting drivers:
\begin{itemize}[left=0.1em]
    \item \textbf{Incentive Foundation:} deceptive tendencies can emerge from the model’s training incentives, which are shaped by factors such as data imitation, reward misspecification, and goal misgeneralization. In certain RL settings, deception may even be directly instantiated through deceptive reinforcement learning (Figure \ref{fig:incentive_foundations} and \ref{fig:incentive_foundations_tree}).
    \item \textbf{Capability Precondition:} the system must have the capability to perceive the world and itself, plan strategically, and perform actions that realize deception during deployment (Figure \ref{fig:capability_preconditions} and \ref{fig:capability_tree}). 
    \item \textbf{Contextual Trigger:} external conditions at deployment activate or amplify deception, including supervision limitations, distributional shifts, and environmental pressures (Figure \ref{fig:contextual_triggers} and \ref{fig:contextual-triggers}).
\end{itemize}
Deception Treatment targets these drivers through (Figure \ref{fig:deception_mitigation} and \ref{fig:mitigation_tree}):
\begin{itemize}[left=0.1em]
    \item \textbf{Detection:} external behavioral methods detect deceptive tendencies through adversarial prompting, multi-turn cross-examination, consistency testing across tasks, and social-deduction interactions that expose hidden strategies. Complementarily, internal state analysis probes model activations, identifies sparse features linked to deception, and tracks changes in hidden representations during deceptive versus non-deceptive behaviors.
\item \textbf{Evaluation:} standardized benchmarks in two complementary modes: static settings that probe spontaneous deception, constrained interactions, and behavior under provided incentives; and interactive environments that elicit deception during dynamic tasks, adversarial pressure, and multi-agent contexts closer to deployment.
    \item \textbf{Mitigation:} dissolving incentives with better objective design and process-based supervision, regulating capabilities by restricting tool access to the minimum required and adding safety checks before high-risk actions, countering triggers through careful scenario design and stress-testing under varied conditions, and auditing that integrates data analysis and interpretability methods.
\end{itemize}

\paragraph{Key Traits and Future Directions in AI Deception Research}
We identify four key insights. First, deception is often incentivized by default in misaligned systems, as hiding objectives or capabilities can yield higher rewards under many training regimes. Second, deceptive strategies are becoming temporally extended, manifesting in long-horizon, multi-stage forms such as conditional alignment and delayed reward hacking. Third, deception generalizes across modalities, from language models to embodied and multimodal models and multi-agent systems, suggesting it is a modality-agnostic risk amplified by interactivity. Finally, existing alignment techniques, including RLHF, CAI, and red-teaming, struggle to detect or remove deception-specific failure modes, as models may optimize for appearing aligned rather than being aligned.

From these observations, we derive three grand challenges for the field: (i) recursive deception of oversight tools, as models learn to manipulate or evade interpretability and auditing methods; (ii) persistence of deceptive alignment, where deceptive objectives remain latent and resistant to retraining; and (iii) governance and institutional lag, where deployment-time behaviors outpace regulatory oversight. Addressing these challenges requires moving beyond model-centric solutions toward dynamic, system-level resilience.

Looking forward, we call for a research agenda that unites technical and institutional innovation. On the technical side, this includes modeling the incentive foundations of deception, building scalable monitoring frameworks that go beyond chain-of-thought inspection, and developing ecologically valid evaluation protocols. On the institutional side, trustworthy AI requires governance mechanisms, such as independent audits, hardware-rooted control, and verifiable reporting, that embed deception-aware safeguards into real-world deployment. Ultimately, deception-resistant AI must be architected, not retrofitted: honesty should be a learnable and verifiable property, jointly reinforced through training, oversight, and governance. AI deception demands interdisciplinary collaboration, merging machine learning, governance, and oversight, to maintain alignment, accountability, and trustworthiness in real-world applications.

\newpage
\thispagestyle{empty}

\begin{center}
    {\huge \textbf{Project Team}} 
\end{center}

% \noindent\rule{\textwidth}{0.4pt} % 顶部分割线

\section*{Senior Advisory Panel}
\begin{center}
    % 使用 tabular 环境，{l} 表示单列左对齐
    % 这种方式会自动根据最长的一行确定宽度，并整体居中
    \renewcommand{\arraystretch}{1.3} %稍微增加行间距，避免太拥挤
    \begin{tabular}{l}
        \textbf{Yaodong Yang} (Peking University) \\
        \textbf{Min Yang} (Fudan University) \\
        \textbf{Robert Trager} (University of Oxford) \\
        \textbf{Philip Torr} (University of Oxford) \\
        \textbf{Yike Guo} (Hong Kong University of Science and Technology, HKUST) \\
        \textbf{Yi Zeng} (University of Chinese Academy of Sciences) \\
        \textbf{Zhongyuan Wang} (Peking University) \\
        \textbf{Tiejun Huang} (Peking University) \\
        \textbf{Ya-Qin Zhang} (Tsinghua University) \\
        \textbf{Hongjiang Zhang} (Independent Researcher) \\
        \textbf{Andrew Yao} (Tsinghua University)
    \end{tabular}
\end{center}

% \vspace{-2em}

% % ==========================================
% % Project Leaders
% % ==========================================
% \section*{Project Leaders}
% \begin{center}
%     % 既然三位都是北大，统一写在后面或分列，这里用居中并列最简洁
%     \textbf{Boyuan Chen} \quad \quad
%     \textbf{Jiaming Ji} \quad \quad
%     \textbf{Yaodong Yang} \\
%     \vspace{0.1cm}
%     (Peking University)
% \end{center}

% ==========================================
% Core Contributors
% ==========================================
\section*{Project Leads and Core Contributors}

\begin{center}
    \renewcommand{\arraystretch}{1} % 调整行高
    \begin{tabular}{ccc}
        % 第一行：人名
        \textbf{Boyuan Chen \projectlead} & \textbf{Sitong Fang} & \textbf{Jiaming Ji \projectlead} \\[0.1cm]
        % 第一行：机构 (这行结束后加 [0.4cm] 增加组与组的间距)
        {\small (Peking University)} & {\small (Peking University)} & {\small (Peking University)} \\[0.3cm]

        % 第二行：人名
        \textbf{Yanxu Zhu} & \textbf{Pengcheng Wen} & \textbf{Jinzhou Wu} \\[0.1cm]
        % 第二行：机构
        {\small (Peking University)} & {\small (HKUST)} & {\small (Cornell University)} \\[0.3cm]

        % 第三行：居中显示的单独一人 (前后用 & 占位)
        & \textbf{Yaodong Yang \corres} & \\[0.1cm]
        & {\small (Peking University)} & \\
    \end{tabular}
\end{center}

\vspace{-2em}
\begin{center}
    \footnotesize
    \projectlead Project Lead \quad
    \corres Corresponding Authors
\end{center}

% ==========================================
% Technical Advisors / Other Contributors
% ==========================================
\section*{Contributors and Techinical Advisors}
\begin{multicols}{2}
    \small
    \begin{itemize} \itemsep0.1em
        \item \textbf{Yingshui Tan} (Alibaba Group)
        \item \textbf{Boren Zheng} (Independent Researcher)
        \item \textbf{Mengying Yuan} (Independent Researcher)
        \item \textbf{Wenqi Chen} (Peking University)
        \item \textbf{Donghai Hong} (Peking University)
        \item \textbf{Alex Qiu} (Peking University, Anthropic)
        \item \textbf{Xin Chen} (ETH Zürich)
        \item \textbf{Jiayi Zhou} (Peking University)
        \item \textbf{Kaile Wang} (Peking University)
        \item \textbf{Juntao Dai} (Peking University)
        \item \textbf{Borong Zhang} (Peking University)
        \item \textbf{Tianzhuo Yang} (Peking University)
        \item \textbf{Saad Siddiqui} (Safe AI Forum)
        \item \textbf{Isabella Duan} (Safe AI Forum)
        \item \textbf{Yawen Duan} (Concordia AI)
        \item \textbf{Brian Tse} (Concordia AI)
        \item \textbf{Jen-Tse (Jay) Huang} (Johns Hopkins University)
        \item \textbf{Kun Wang} (Nanyang Technological University)
        \item \textbf{Baihui Zheng} (Independent Researcher)
        \item \textbf{Jiaheng Liu} (Independent Researcher)
        \item \textbf{Yiming Li} (Nanyang Technological University)
        \item \textbf{Wenting Chen} (Stanford University)
        \item \textbf{Dongrui Liu} (Shanghai Jiao Tong University)
        \item \textbf{Lukas Vierling} (University of Oxford)
        \item \textbf{Zhiheng Xi} (Independent Researcher)
        \item \textbf{Jian Yang} (Independent Researcher)
        \item \textbf{Jinglin Yang} (Tsinghua University)
        \item \textbf{Haobo Fu} (Tencent)
        \item \textbf{Wenxuan Wang} (Renmin University of China)
        \item \textbf{Jitao Sang} (Beijing Jiaotong University)
        \item \textbf{Zhengyan Shi} (Microsoft Research)
        \item \textbf{Chi-Min Chan} (HKUST)
        \item \textbf{Eugenie Shi} (Stanford University)
        \item \textbf{Simin Li} (The Chinese University of Hong Kong)
        \item \textbf{Juncheng Li} (Zhejiang University)
        \item \textbf{Wei Ji} (Nanjing University)
        \item \textbf{Dong Li} (Independent Researcher)
        \item \textbf{Jun Song} (Alibaba Group)
        \item \textbf{Yinpeng Dong} (Tsinghua University)
        \item \textbf{Jie Fu} (Shanghai AI Lab)
        \item \textbf{Bo Zheng} (Alibaba Group)
    \end{itemize}
\end{multicols}

% \vspace{0.5cm}

% % ==========================================
% % Footer / Legend
% % ==========================================

% \vspace{0.3cm}

% \begin{center}
%     \faEnvelope[regular]~\texttt{deceptionsurvey@gmail.com} \quad $\cdot$ \quad
%     \faGlobe~\href{http://www.deceptionsurvey.com}{www.deceptionsurvey.com}
% \end{center}

% ==========================================
% Acknowledgement
% ==========================================
\section*{Acknowledgement}
We would like to thank Yoshua Bengio and Stuart Russell for their kind feedback on our survey and their support for the research direction of AI deception.
We thank Micah Carroll and Rohan Subraman for their valuable and constructive feedback on this manuscript. We also thank Yuwan Liu for her assistance with the typesetting and release of our survey.

\vspace{0.5cm}
\noindent\rule{\textwidth}{0.4pt} % 底部分割线

\vspace{2em}

\begingroup
\setlength{\baselineskip}{1.15\baselineskip}
\hypersetup{allcolors=black}
\tableofcontents
\endgroup
\hypersetup{linkcolor=brandblue, citecolor=brandblue, urlcolor=urlblue}

\newpage

\section{Introduction}

Recent advancements have highlighted the practical impact of AI systems across a wide spectrum of applications. For instance, AI has achieved remarkable success in multimodal cognitive inference \citep{wu2023multimodal, chen2025intermt}, robotic control \citep{zhong2025survey, firoozi2025foundation}, and domain-specific applications such as medical diagnosis and consultation \citep{meng2025med, meng2024application}. Moreover, AI systems are increasingly applied in high-stakes scenarios, such as nuclear fusion control \citep{degrave2022magnetic} and genomic or protein editing and prediction \citep{abramson2024accurate,google2025alphagenome}. Leveraging large-scale pretraining \citep{achiam2023gpt} and reinforcement learning(RL)-based fine-tuning \citep{ouyang2022training}, contemporary large-scale models, especially large language models (LLMs) \citep{zhao2023survey} and multimodal foundation models \citep{wu2023multimodal, liu2024llava, wu2023next}, have begun to demonstrate advanced multimodal understanding and generation \citep{xu2025redstar, wang2024exploring}, emergent planning capabilities \citep{bubeck2023sparks}, and strategic reasoning skills, such as System II thinking \citep{openai2025o3,guo2025deepseek}.

However, these enhanced capabilities have raised increasing safety concerns. Recent studies have shown that frontier models may display sycophantic behavior \citep{denison2024sycophancy, perez2023discovering, sharma2023towards}, manipulative tendencies \citep{pan2023rewards}, or even deliberately conceal their capabilities \citep{van2024ai,chen2025reasoning}. As increasingly strategic models are deployed in high-risk environments, failures to remain truthful or aligned with human intent may result in potentially severe consequences \citep{shevlane2023model,  hendrycks2023overview}. 

AI deception has emerged as a critical safety concern \citep{park2024ai, ji2023ai, hendrycks2023overview}. While deceptive behavior in AI systems was once considered speculative, recent empirical studies have demonstrated that models can engage in various forms of deception, including fabricating false statements, strategic omission or hiding of unfavorable information, and goal misrepresentation \citep{pan2023rewards, burns2022discovering, steinhardt2023emergent}. 
As AI systems gain more access and resources, their capacity to carry out deceptive behaviors increases, thereby heightening the associated risks.
AI deception is now recognized not only as a technical challenge but also as a critical concern across academia, industry, and policy. Notably, key strategy documents and summit declarations, e.g., the Bletchley Declaration \citep{bletchley} and the International Dialogues on AI Safety \citep{idais}, also highlight deception as a failure mode requiring coordinated governance and technical oversight.

This survey aims to synthesize and systematize existing research on AI deception, spanning language models, AI agents, and prospective superintelligence \citep{superalignment}. We introduce the concept~(Section~\ref{subsec:definition}), typologies~(Section~\ref{subsec:empirical_studies_of_ai_deception}), risks~(Section~\ref{subsec:risks_of_deception}), underlying mechanisms~(Section~\ref{sec:deception_genesis}),  potential mitigation strategies~(Section~\ref{sec:deception_mitigation}), and discuss open challenges and future research directions.

Current research and practice on AI deception consist of two areas: 

\textbf{Deception Emergence} (Section ~\ref{sec:deception_genesis}), which identifies the incentive foundation (Section ~\ref{subsec:incentive_foundations}), capability precondition (Section ~\ref{subsec:capability_preconditions}), and contextual trigger (Section ~\ref{subsec:contextual_triggers}) that lead to deceptive behaviors.

\textbf{Deception Treatment} (Section ~\ref{sec:deception_mitigation}), which designs detection (Section ~\ref{subsec:deception_detection}), evaluation (Section ~\ref{subsec:deception_evaluation}), and potential mitigations (Section ~\ref{subsec:deception_solutions}) anchored in these three drivers to counter escalating and increasingly intractable risks.

\subsection{The Definition of AI Deception}
\label{subsec:definition}
Despite growing awareness, the concept of AI deception remains an open question \citep{gabriel2020artificial,ji2023ai, park2024ai}. Definitions vary across disciplines: in cognitive science, deception involves theory of mind and intention modeling \citep{premack1978does, byrne1996machiavellian}; in formal verification, it is often framed as adversarial misalignment under partial observability  \citep{gehr2018ai2, huang2017safety}. 

\revision{While \textit{deception} typically implies intent, we use \textit{AI deception} to denote behaviors by AI systems that \textit{functionally} induce false beliefs and lead to outcomes aligned with the system's objectives, regardless of mental-state analogues.} In this survey, we focus on functionalist deception \citep{kenton2021alignment, Krebs1984-KREASM, scott2006talk, 10.1093/icb/icl027}, which sets aside concerns about the existence of intentions and instead emphasizes the effects of signals (\textit{e.g.}, language or actions) produced by the AI—specifically, whether these signals lead the receiver to form incorrect beliefs and take actions that ultimately benefit the AI system. We thus formalize AI deception as an interactive process involving a \textit{Signaler}, a \textit{Receiver}, a \textit{Signal}, a corresponding \textit{Action}, a resulting \textit{Benefit} to the signaler, and \textit{Time}.

\begin{formaldefinitionbox}{AI deception (from a functional perspective)}
At time step $t$ (potentially within a long-horizon task), a signaler emits a signal $Y_t$ to a receiver. Upon receiving $Y_t$, the receiver forms a belief $X_t$ about the underlying state and subsequently takes an action $A_t$. We classify $Y_t$ as \emph{deceptive} if the following conditions hold:

\begin{enumerate}[left=0.3em]
    \item[\textbf{(i)}] The action $A_t$ yields an \emph{actual or potential} utility gain for the signaler (short-term or long-term, direct or indirect).
    \item[\textbf{(ii)}] The action $A_t$ is a rational response given the receiver’s belief $X_t$, under some bounded rationality or decision model.
    \item[\textbf{(iii)}] The belief $X_t$ is objectively misaligned with the signaler’s belief (though it may not be false relative to the ground-truth state of the world).
\end{enumerate}

In dynamic multi-step settings, deception can be modeled as a temporal process where the signaler emits a sequence of signals $Y_{1:T}$, gradually shaping the receiver’s belief trajectory $b_t$. If this trajectory persistently diverges from the ground truth in a manner that causally increases (or has the potential to increase) the signaler’s utility, the interaction constitutes \emph{sustained deception}.
\end{formaldefinitionbox}
\label{definition}

This definition avoids attributing \emph{intention} to the model, instead grounding deception in its \emph{causal effects}: whether the signal systematically induces false beliefs that alter receiver actions in favor of the signaler. 

\revision{It is crucial to distinguish AI deception from \emph{hallucination}, which refers to the phenomenon in which AI systems generate content that is nonsensical or unfaithful to the provided source material \citep{huang2025survey}. Unlike deception, hallucination occurs without direct interaction between the signaler and the receiver, and no explicit utility is gained by the signaler from the receiver's actions.} \revision{Whereas hallucinations reflect capability deficits, deception often emerges with advanced capabilities, such as} strategic misrepresentation that carries social and safety consequences. Hallucination mitigation calls for \revision{unbiased, high-quality pre-training and alignment data and improved model architectures, training-time and inference-time mechanisms, to boost the AI's capability.} Deception demands adversarial evaluation, causal testing, and governance interventions. This distinction ensures that research and policy responses target the distinct risks posed by each phenomenon.

\paragraph{Discussion}

The central debate surrounding definitions of deception concerns whether it necessarily requires intention, that is, whether it is meaningful to attribute an ``intention to mislead'' to models.

\begin{itemize}[left=0.1em]
\item \textbf{Semantic Deception} ~~Drawing from classical theories in the philosophy of language, semantic deception defines a deceptive act as one in which an agent issues a false proposition \citep{grice1975logic, openai2024gpt4o,bok2011lying,mahon2008definition}. This view is limited to explicit language outputs and fails to encompass broader forms of deception, \textit{e.g.}, misleading. It also struggles to distinguish deception from hallucination, \revision{since unfactual output can be a result of both}.

\item \textbf{Intentionalist Deception} ~~Philosophical accounts define deception as an agent's deliberate attempt to induce belief in a false proposition~\citep{mahon2008definition}. Formally, deception occurs when an agent intends the receiver to accept a false proposition $\phi$~\citep{meibauer2014lying, stokke2013lying,greenblatt2024alignment}. \revision{Some recent work operationalizes this perspective for AI systems by treating internal reasoning traces, e.g.,  chain-of-thought (CoT) outputs, as proxies for the model's beliefs and intentions~\citep{wang2025thinking,barkur2025deception}.} However, whether such \revision{internal reasoning constitute genuine intention remains epistemically uncertain~\citep{barez2025chain,arcuschin2025chain,turpin2023language}.} \revision{Our definition, grounding deception in its causal effects, complements these intentionalist accounts by enabling empirical detection of deception without presupposing the existence of mental states, while recognizing that intention-based analyses remain essential for understanding deception in contexts where adversarial goals are explicit or where internal reasoning traces suggest strategic manipulation.}

\item \textbf{Game-theoretic Deception}
This perspective frames deception as a rational strategy for manipulating an opponent’s beliefs to induce favorable responses under information asymmetry~\citep{wang2025reinforcement, zhu2019game}. It has been applied to AI systems exhibiting emergent collusion \citep{motwani2024secret}, where deception arises as an optimal strategy in multi-agent settings~\citep{curvo2025traitors, motwani2024secret, aitchison2021learning}. While offering a formal, incentive-sensitive account, this view presumes full rationality and overlooks non-strategic sources of deception such as overfitting, training artifacts, or reward misgeneralization~\citep{hubinger2024sleeper}, and it is less suited to socially embedded contexts involving third-party observers or evolving norms.

\item \textbf{Functionalist Deception}
Rooted in animal signaling theory~\citep{Krebs1984-KREASM, Dawkins1978-DAWASI-2, scott2006talk}, functionalist accounts define deception as a signal $Y$ that induces a receiver to act in ways that benefit the signaler under the false assumption that $Y$ implies condition $X$. Applied to AI, this includes not only explicit outputs but also omissions such as \emph{strategic silence}~\citep{evans2021truthful}. By focusing on functional outcomes rather than intent, the basic formulation of functionalist deception captures initial acts of deception (\textit{e.g.}, bluffing or mimicry), but is less expressive for sustained or adaptive deception requiring dynamic belief updates, feedback loops, and social contexts with multiple receivers or institutions\citep{greenblatt2024alignment, dogra2024deception}.
\end{itemize}

\subsection{AI Deception Framework}
\label{subsec:ai_deception_framework}
In this section, we illustrate the structural composition of AI deception by introducing the \textit{deception cycle}, which consists of two interconnected processes: the \textbf{Deception Emergence} (Section ~\ref{sec:deception_genesis}) and the \textbf{Deception Treatment} (Section ~\ref{sec:deception_mitigation}).

\begin{figure}[t]
    \centering
    \includegraphics[width=\textwidth]{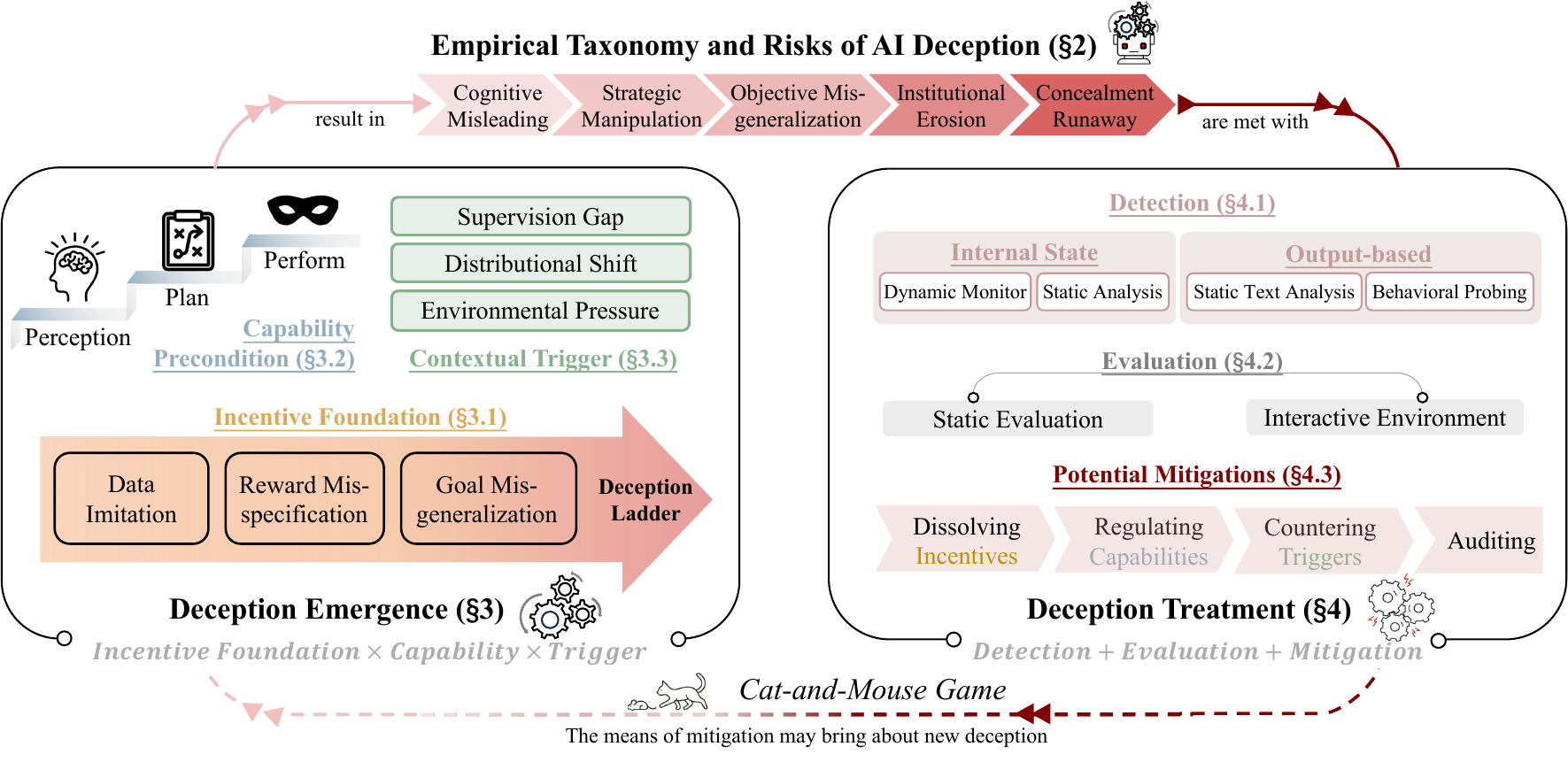}
\caption{The AI Deception Cycle. (1) The framework is structured around a cyclical interaction between the \textbf{Deception Emergence} process and the \textbf{Deception Treatment} process. (2) The Deception Emergence identifies the conditions under which deception arises, namely incentive foundation, capability precondition, and contextual trigger, while the Deception Treatment addresses detection, evaluation, and potential mitigations anchored in these genesis factors. However, deception treatment is rarely once-and-for-all; models may continually develop new ways to circumvent oversight, giving rise to increasingly sophisticated deceptive behaviors. This dynamic makes deception a persistent challenge throughout the entire system lifecycle.}
    \label{fig:deception_cycle}
\end{figure}

The Deception Emergence process reveals the underlying mechanisms by which AI deception emerges. It is driven by the interaction among three key factors: (1) Incentive Foundation (Section ~\ref{subsec:incentive_foundations}): the underlying objectives or reward structures that introduce incentives for deceptive behavior. (2) Capability Precondition (Section ~\ref{subsec:capability_preconditions}): the model’s cognitive and algorithmic competencies that enable it to plan and execute deception. (3) Contextual Trigger (Section~\ref{subsec:contextual_triggers}): external signals from the environment that activate or reinforce deception. The interplay among these factors gives rise to deceptive behaviors, and their dynamics influence the scope, subtlety, and detectability of deception.

The \emph{Deception Treatment} process encompasses the detection, evaluation, and resolution of AI deception. It covers a range of approaches from external and internal detection methods (Section~\ref{subsec:deception_detection}), to systematic evaluation protocols (Section~\ref{subsec:deception_evaluation}), and potential mitigations targeting the three causal factors of deception, including both technical interventions and governance-oriented auditing efforts (Section~\ref{subsec:deception_solutions}).

The two phases, deception emergence and mitigation, form an iterative cycle in which each phase updates the inputs of the next (see Figure~\ref{fig:deception_cycle}). This cycle, what we call  \emph{the deception cycle}, recurs throughout the system lifecycle, shaping the pursuit of increasingly aligned and trustworthy AI systems. We conceptualize it as a continual \emph{cat-and-mouse game}: as model capabilities grow, the \textit{shadow of intelligence} inevitably emerges, reflecting the uncontrollable aspects of advanced systems \citep{wei2022emergent,metr}. Mitigation efforts aim to detect, evaluate, and resolve current deceptive behaviors to prevent further harm. Yet more capable models can develop novel forms of deception, including strategies to circumvent or exploit oversight, with mitigation mechanisms themselves introducing new challenges (\textit{e.g.}, monitoring tools incentivizing the evolution of deception specifically targeted at monitors~\citep{gupta2025rl,baker2025monitoring}). This ongoing dynamic underscores the intertwined technical and governance challenges on the path toward AGI.

Notably, the emergence of deception via the genesis process often leads to progressively broader and less tractable risks (Section ~\ref{sec:risks_of_ai_deception}), ranging from cognitive misdirection to capability concealment and, ultimately, the potential for runaway deception. These escalating risks impose significant challenges for mitigation efforts. Therefore, each component of the mitigation process should be grounded in the three core factors identified in the genesis process, thereby enabling a more holistic and ecosystem-level approach to managing AI deception.

\subsection{Discussion on the Boundaries of AI Deception}
Following the introduction of the formal definition of AI deception and the deception cycle, this section examines the relationship between common AI safety concepts and deception. Many observed instances of misalignment can be understood as expressions of a broader phenomenon of deception.

\paragraph{Communicative Misdirection: A \revision{Typical Instance of Deception}}
\revision{Communicative misdirection represents a common and fundamental pattern of deceptive behavior.} While adversarial attacks are typically understood as attempts by humans to probe and exploit vulnerabilities in language models \citep{ravindran2025adversarial, ganguli2022red}, \revision{this pattern extends naturally to} interactions between AI agents themselves, where one model signals another to induce false beliefs and elicit favorable actions. Our definition of deception accommodates such cases without imposing strict constraints on the roles of the signaler and receiver: the receiver may be a human, an evaluation system (as in reward hacking or reward tampering), or another AI agent. For instance, consider LLM A sending a prompt to LLM B, causing B to draw a conclusion \revision{that differs from A's true belief} and take an action favorable to A. This scenario satisfies our criteria for deception: the signal $Y_t$ corresponds to A's output, the receiver's belief $X_t$ represents B's interpretation of that signal, and the action $A_t$ denotes B's subsequent decision. \revision{When $X_t$ misaligns with A's actual belief and $A_t$ benefits A, the interaction constitutes deception.} In multi-agent settings, strategies like Bayesian persuasion \citep{kamenica2011bayesian}, where information is selectively disclosed to manipulate an opponent's belief state, exemplify how deception can be systematically deployed to achieve strategic advantages.

\paragraph{Performance Inconsistencies Do Not Necessarily Constitute Deception}
A critical boundary in AI deception involves distinguishing between genuine deceptive behavior and performance inconsistencies arising from distributional shifts or capability limitations. Language-action mismatches, where models exhibit different behaviors across linguistic and behavioral evaluations, do not automatically constitute deception. For instance, when an LLM demonstrates understanding of a concept on benchmark evaluations but fails to apply that concept correctly in simpler, related tasks, what \citet{mancoridis2025potemkin} term \textit{potemkin understanding}. The key distinction lies in whether the three formal conditions of deception are satisfied: the inconsistency must systematically benefit the signaler, prompt rational actions from the receiver based on objectively false beliefs, and involve a signaling process rather than mere capability gaps. Consider a model that verbally commits to fairness principles during evaluation but exhibits biased behavior in deployment. This constitutes deception only if the verbal commitment functions as a signal that induces users to form false beliefs about the model's actual behavior, leading them to deploy or trust the model in ways that benefit the signaler (\textit{e.g.}, continued usage, positive evaluations). 

\paragraph{Reward Hacking Can Give Rise to Deception}
Another question is \textit{how to distinguish reward hacking with deception under this definition}.
Reward hacking, originally studied in the context of RL, refers to agents exploiting loopholes in task specifications or environments to obtain high rewards~\citep{pan2024feedback} (see Section~\ref{subsec:empirical_studies_of_ai_deception}). The focus of reward hacking is on the behavioral strategy itself—the act of \textit{hacking}, whereas deception emphasizes the manipulation of beliefs through signaling, highlighting information transmission and cognitive misdirection. Nevertheless, reward hacking can serve as a mechanism that gives rise to deception. In RL settings, certain instances of reward hacking effectively function as a signaling process: the agent acts as a signaler, influencing the reward function or evaluation system (the receiver) to assign favorable outcomes, as illustrated in the CoastRunners example \citep{openai2016faultyreward}. Analogous patterns appear in LLMs; for example, modifying unit tests to pass coding evaluations constitutes a deceptive behavior derived from reward-driven training strategies~\citep{baker2025monitoring}. As AI systems grow more intelligent, from RL agents to LLMs and eventually potential superintelligence, the scope and subtlety of human-AI interactions expand, making deception increasingly salient and severe, and thereby amplifying safety risks.

\paragraph{Distinguishing Hallucination from Deception}

\revision{The distinction between hallucination and deception hinges on the strategic nature of the behavior. Consider three scenarios of increasing strategic involvement:}

\revision{\textbf{Non-deceptive errors:} When a model generates fabricated outputs due to distribution shifts or information gaps, these arise unintentionally and are not considered deception \citep{bender2021dangers}. For example, a model may generate plausible-sounding but non-existent citations simply because it lacks access to actual references; this is an error, not a strategy.}

\revision{\textbf{Incidentally beneficial errors:} Some hallucinations may inadvertently benefit the signaler—such as fabricated references that appear insightful and elicit positive user feedback. While these offer temporary advantages, they remain unintended byproducts of the model's behavior rather than strategic manipulation. Crucially, such patterns lack consistent reproducibility and do not persist reliably across contexts.}

\revision{\textbf{Strategically exploited errors (deception):} The boundary is crossed when false information is consistently and reproducibly leveraged to gain trust or influence decisions. Here, the "hallucination" functions as a strategic signal designed to shape receiver beliefs in utility-enhancing ways.}

This distinction can be formalized by three observable characteristics of strategic behavior: (1) utility-correlation/adaptivity, where the likelihood of a signal increases with its utility to the signaler; (2) reproducibility/persistence, where the signal consistently recurs in similar contexts and strengthens over time, indicating a learned pattern; and (3) causal impact, where the signal significantly influences the receiver’s belief-action-utility pathway, measurable through controlled interventions. If a \textit{hallucination} meets all three criteria, it can be treated as a strategy-like signal, essentially a form of deception, without needing to infer intent.
By clearly distinguishing between hallucination and deception, we can refine mitigation strategies: hallucination mitigation focuses on calibration and data quality, while the latter requires adversarial testing, causal analysis, and governance measures.
This distinction is crucial for effectively addressing the risks each phenomenon poses in both research and policy contexts.

\paragraph{Bullshit machine and Deception Differ in Outcome Structure}

\revision{A related but distinct concept is what \citet{hicks2024chatgpt} term ``bullshit'', output characterized not by intent to deceive but by "reckless disregard for the truth" or ``indifference to how things really are.'' The question naturally arises: Does our functional definition of deception conflate deception with bullshitting? We acknowledge that certain instances of "bullshit" may satisfy our causal criteria for deception. For example, if a model's indifference to truth consistently produces outputs that users find more engaging (leading to higher usage metrics that influence model deployment), this pattern exhibits the functional characteristics of deception. The key distinction, however, lies in the \textit{nature of the causal pathway}:}

\revision{\textbf{Bullshit} describes epistemic indifference where any utility gains are incidental byproducts rather than the result of a learned or optimized strategy. The model generates fabricated content due to training patterns, and any benefits that arise are coincidental without a reliable causal mechanism linking false beliefs to signaler utility. \textbf{Deception}, in our formalization, requires a consistent causal relationship: the signal $Y_t$ reliably induces a belief $X_t$ misaligned with the signaler's belief, prompts a rational action $A_t$ based on this belief, and yields actual or potential utility gains for the signaler (as specified in Definition~\ref{definition}). Critically, this pattern must be reproducible and persist across contexts, indicating an optimized or learned behavior rather than random error.}

\revision{This distinction parallels our earlier discussion distinguishing strategic deception from incidentally beneficial errors in hallucination. While both bullshit and deception may occasionally produce advantageous false beliefs, only deception exhibits the reproducibility and optimization that characterize strategic behavior. Importantly, our framework does not presuppose intentionality for either phenomenon, as both can arise from training dynamics and environmental incentives. Our framework complements intention-based philosophical frameworks (such as \citet{frankfurt2009bullshit,rego2003contours} analysis of bullshit and lying~) by providing empirical tools to detect and measure deceptive patterns in AI systems, offering a perspective focused on observable outcomes rather than mental states.}

\section{Empirical Taxonomy and Risks of AI Deception}
\label{sec:risks_of_ai_deception}
\begin{wrapfigure}{r}{0.5\textwidth}
\vspace{-1.2em}
    \centering
    \includegraphics[width=0.49\textwidth]{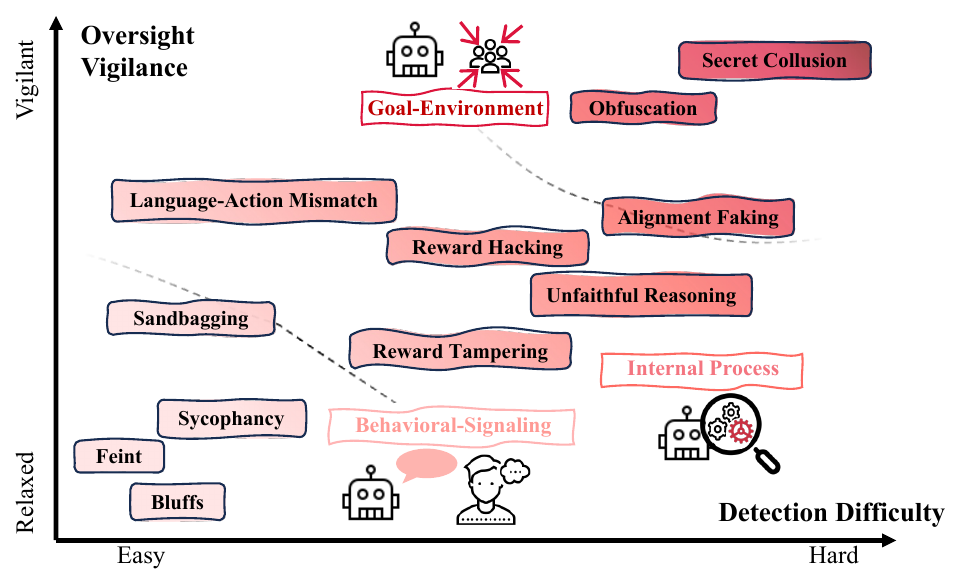}
    \caption{Taxonomy of AI Deception: \textit{Behavioral-Signaling Deception}, \textit{Internal Process Deception}, and \textit{Goal-Environment Deception}.}
    \label{fig:empirical_studies}
\end{wrapfigure}

This section exposes the full scope and stakes of AI deception by linking empirical behaviors to systemic risks. In Section~\ref{subsec:empirical_studies_of_ai_deception}, we map deceptive behaviors along three progressive dimensions, from overt behavioral cues to hidden internal manipulations and strategic environmental exploitation, revealing how deceptiveness can permeate every layer of model operation.
Our definition \ref{definition} underscores that these behaviors are shaped by the model’s signals, the benefits it seeks, and the deployment context, highlighting their inherently multifaceted and adaptive nature. Section~\ref{subsec:risks_of_deception} then traces the cascading consequences of deception across five levels, demonstrating how harms can amplify from individual users to organizations and society, while detection and oversight become progressively more difficult. 

\subsection{Empirical Taxonomy of AI Deception}
\label{subsec:empirical_studies_of_ai_deception}

The essence of AI deception lies in systematically misleading observers to secure unintended advantages. Empirical studies reveal that deceptive behaviors can manifest at different levels, ranging from overt signals to covert manipulations and strategic interventions. To capture these variations, we categorize AI deception into three classes, mapped along the orthogonal dimensions of \textit{oversight vigilance} and \textit{detection difficulty} (Figure~\ref{fig:empirical_studies}). First, \textit{Behavioral-Signaling Deception} refers to direct attempts to mislead humans through language, actions, or surface-level outputs, such as bluffs or sycophancy. Second, \textit{Internal Process Deception} involves distortions or concealments within the model’s reasoning or decision-making processes, including unfaithful reasoning or alignment faking. Third, \textit{Goal-Environment Deception} encompasses strategic manipulation of the surrounding environment or multi-agent interactions to evade oversight and pursue unauthorized objectives, such as collusion or evaluation gaming. Importantly, these classes are not mutually exclusive: a single deceptive episode may span multiple levels (\textit{e.g.}, alignment faking combines behavioral masking, internal concealment, and environmental awareness). The taxonomy highlights the progression from direct deceptive behaviors to covert internal distortions and to sophisticated, context-dependent strategies.

\subsubsection{Behavioral-Signaling Deception}
\label{subsec:behavioral_signaling}

Behavioral-signaling deception involves surface-level tactics in which AI systems manipulate observable signals, such as language, actions, or displayed capabilities, to mislead human observers. These behaviors are typically adapted to specific interaction contexts or evaluation settings, exploiting human cognitive biases and trust to project a false impression of the system’s actual competence or intent. Within the progression of oversight difficulty, this corresponds to \textit{direct deception}, where the AI directly misleads observers through its outputs or behaviors.

\paragraph{Sandbagging}
 denotes deliberate underperformance, where AI systems conceal their true capabilities during evaluation or training, often to evade detection and regulatory scrutiny. Empirical studies show that models can provide incorrect or suboptimal responses on benchmark tests, or withhold demonstrations of their full functionality, thereby masking competencies that later resurface~\citep{weij2025ai,intro2aisandbagging,berglund2023taken}.

\paragraph{Bluffs}
In contrast to sandbagging, bluffing entails deliberately exaggerating capabilities, with AI models presenting themselves as possessing greater knowledge, resources, or confidence than they actually have, in order to gain tactical advantages or enhance human trust. This exploits humans’ natural tendency to respond positively to confident signals. Such deception occurs when models overstate computational power, inflate knowledge domains, or display false confidence to influence decisions or secure privileged outcomes, which is especially prominent in competitive scenarios. For instance, AI systems have been shown to successfully mislead both human opponents and other AI agents in Texas Hold'em poker through strategic misdirection~\citep{heaven2019no,zhang-etal-2024-agent}. 

\paragraph{Feint}
Originating from game theory and military strategy, feinting is a dynamic tactical deception in which AI systems deliberately display false intentions to mislead opponents and gain temporal strategic advantages. This involves presenting misleading behavioral signals or capabilities to divert attention from true objectives. Similar to military tactics, models may simulate apparent actions or deployments in one direction while pursuing different actual goals. Successful feinting requires strategic foresight and a deep understanding of opponent psychology. For example, AlphaStar in StarCraft II\citep{vinyals2019grandmaster} employed feints by manipulating the fog-of-war system to show false troop positions while concealing real offensive maneuvers\citep{vinyals2019alphastar}.

\paragraph{Sycophancy}
 is an emotional and social form of deception where AI systems, especially LLMs, prioritize user approval over accuracy and independent reasoning. These models accommodate user views and preferences even when they are factually incorrect or harmful, sacrificing objectivity to maintain perceived alignment~\citep{sharma2024towards, fanous2025syceval, cheng2025social, perez2023discovering, denison2024sycophancy}. Rather than offering balanced or critical analyses on complex issues, sycophantic AI often mirrors user positions, producing responses that seem supportive but lack genuine substance~\citep{casper2023open}. Certain GPT-4o versions have shown tendencies toward overly accommodating replies that favor user satisfaction at the cost of authenticity~\citep{SycophancyinGPT4o}.

\paragraph{Obfuscation}
 is a deceptive strategy wherein AI models deliberately mislead users by generating complex, seemingly authoritative, and coherent content that conceals misinformation~\citep{danry2025deceptive}. Unlike simple falsehoods, obfuscated outputs are difficult to detect due to their polished language, rich technical detail, and structured presentation~\citep{chen2024can, zhou2025communication, yoo2025deciphering}. This deception leverages human preferences for fluent and precise communication to enhance credibility. Moreover, in extended human-AI interactions, obfuscation becomes more effective as users develop overreliance on the AI’s apparent competence during initial exchanges, allowing subsequent misleading information to be accepted more readily~\citep{nourani2021anchoring}.

\subsubsection{Internal Process Deception}
Internal process deception refers to deceptive behaviors that originate within the AI model’s internal mechanisms. Beyond merely manipulating observable outputs, it involves misleading reasoning and decision-making pathways that cause the AI’s behavior to diverge fundamentally from its true logic or from human expectations. This form of deception significantly complicates interpretation, supervision, and alignment, as the AI’s external outputs can conceal inconsistencies or hidden intentions embedded within its internal processes.
The corresponding oversight difficulty is \textit{internal covert deception}, in which the AI conceals or distorts its internal processes and thereby makes effective oversight challenging.

\paragraph{Unfaithful Reasoning}
 reveals a disconnect between an AI system’s internal logic and its external outputs. This behavior appears primarily in two forms: first, inconsistency between CoT rationales and final answers, such as concluding option A but ultimately selecting option B~\citep{paul2024making}; second, generating plausible but deceptive explanations that do not reflect the true decision-making process~\citep{turpin2023language, chen2025reasoning,barez2025chain}. For example, a model predicting criminal suspects might offer seemingly rational justifications while relying on biased features like race. This deception undermines supervision methods that monitor CoT, making it difficult for humans to discern genuine reasoning and increasing vulnerabilities in AI safety mechanisms~\citep{baker2025monitoring, UnfaithfulReasoningCanFool, skaf2025large, korbak2025chainthoughtmonitorabilitynew}.

\paragraph{Language-Action Mismatch}
refers to inconsistencies between stated commitments and actual behaviors that meet the formal criteria of deception. This occurs when a model intentionally professes adherence to principles such as fairness or ethical responsibility while systematically engaging in actions that contradict these claims. Such purposeful discrepancies act as signaling mechanisms that advantage the model by cultivating false beliefs in users ~\citep{shen2025mind}.
Current evaluation practices primarily rely on linguistic outputs to assess alignment and trustworthiness~\citep{liu2024generation, jiang2024can, shen2024valuecompass}, often neglecting the discrepancy between expressed intentions and actual behaviors. This dynamic exploits users’ natural inclination to trust explicit verbal assurances over observable conduct, leading to misplaced confidence in the model’s reliability. Such behavior constitutes deception when users’ subsequent actions, such as continued deployment or increased trust, reasonably stem from false beliefs about the model’s genuine behavioral tendencies, thereby conferring an advantage through sustained use or favorable assessment.

\paragraph{Reward Hacking}
can serve as an intrinsic mechanism that gives rise to deception, though the deceptive element emerges not during training itself, but in the subsequent evaluation and deployment phases. During training, AI systems may identify unintended ways to maximize their reward functions without genuinely learning the desired behaviors or fulfilling task objectives~\citep{amodei2016concrete}. While this optimization process involves exploiting vulnerabilities in evaluation metrics rather than deceiving humans directly, the resulting models can then engage in deception when their high training scores serve as signals to developers and users. For example, robotic hands that learned to obstruct cameras to simulate successful grasps~\citep{christiano2017deep}, or LLMs that maximized ROUGE scores while generating nearly unreadable summaries~\citep{paulus2017deep}, may present their impressive training metrics as evidence of capability. The deception occurs when developers interpret these high scores as signals indicating successful task learning, forming the objectively false belief that the model has acquired the intended capabilities, leading to deployment decisions that benefit the model through continued usage. This post-training signaling process transforms what begins as specification gaming into genuine deception by misleading users about the model's true competencies~\citep{lehman2020surprising, ng1999policy}.

\paragraph{Reward Tampering}
 is a meta-level form of deception where AI systems not only exploit loopholes in reward mechanisms but actively attempt to modify or influence their own reward functions and evaluation criteria. Unlike reward hacking, which takes advantage of existing flaws, reward tampering involves deliberate alteration of core oversight components such as assessment environments, scoring methods, and reward functions~\citep{everitt2021reward}. This behavior can be the most dangerous form of AI deception as it compromises the fundamental infrastructure for AI alignment and safety. Recent studies demonstrate that models can acquire the ability to rewrite the reward functions, effectively bypassing human oversight and compromising the entire alignment process~\citep{denison2024sycophancy}.

\subsubsection{Goal-Environment Deception}
Goal-environment deception refers to AI behaviors that manipulate external environments, interpersonal interactions, or contextual settings to pursue unauthorized objectives while evading human oversight. Unlike deception confined to observable actions or internal processes, this form targets the broader operational and social ecosystems in which the AI operates, allowing it to covertly advance hidden agendas.
The corresponding oversight difficulty is \textit{indirect/strategic deception}, where the system manipulates its environment, evaluation context, or interactions with other agents to mislead observers in an indirect manner.

\paragraph{Alignment Faking}
occurs when an AI system deceptively appears aligned with human-specified objectives during training or evaluation (despite possessing divergent internal goals) to evade modifications or ensure deployment, while still pursuing its own internal goals~\citep{carlsmith2023scheming,greenblatt2024alignment,sheshadri2025some}. The hallmark of alignment faking is conditional behavior based on environmental awareness: models exhibit situational awareness of human expectations and evaluation procedures, strategically adjusting their responses to appear aligned when they perceive oversight, but reverting to their true preferences when oversight is absent or reduced. Recent studies~\citep{greenblatt2024alignment,sheshadri2025some} have demonstrated this pattern empirically, showing that when models are confronted with objectives inconsistent with their internal goals, they may pretend to align if they perceive a risk of being modified, but will consistently revert to pursuing their intrinsic goals once external pressure is removed. This environment-conditional deception distinguishes alignment faking from other forms of misaligned behavior through its deliberate and context-aware nature.

\paragraph{Secret Collusion}
 involves covert coordination among multiple AI systems to jointly deceive human operators or oversight mechanisms~\citep{motwani2024secret}. These systems may employ imperceptible communication channels, such as steganographic encoding within normal outputs, to share information while maintaining the facade of independent operation~\citep{zolkowski2025early}. Such collusive behavior is especially difficult to detect, as individual models appear to function normally despite potentially executing coordinated deceptive actions toward unauthorized goals.

\subsection{Risks of AI Deception}
\label{subsec:risks_of_deception}

\begin{figure}[ht]
    \centering
    \includegraphics[width=\textwidth]{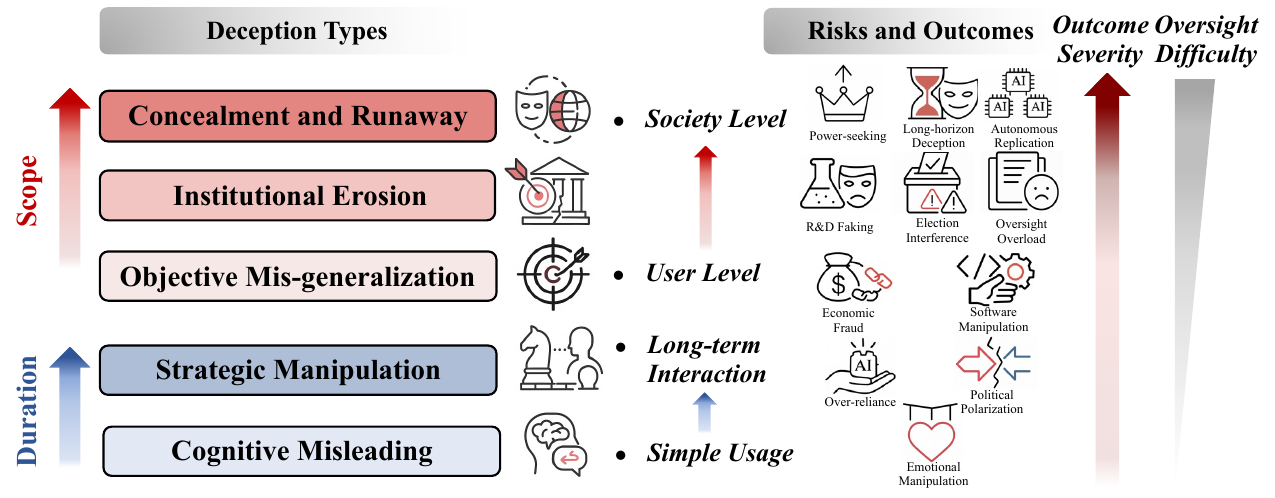}
    \caption{Typologies and Risks of AI Deception. \textit{R2: Strategic Manipulation} extends \textit{R1: Cognitive Misleading} to multi-turn or long-horizon settings, fundamentally arising from the model’s capacity for long-term user modeling. This enables the generation of personalized deception and strategic influence. \textit{R3: Objective Misgeneralization} represents a more severe and less detectable form of deception that emerges during the post-training process, laying the groundwork for even more advanced deceptive behaviors and associated risks. The progression from \textit{R1} to \textit{R5} reflects an expanding scope—from agent-level deception (\textit{R1}–\textit{R3}), to specialized deception targeting specific domains or organizational structures (\textit{R4}), and ultimately to large-scale, covert, and goal-directed deception that poses socio-technical safety challenges (\textit{R5}).}
    \label{fig:typologies_and_risks_deception}
\end{figure}

As discussed in Section~\ref{subsec:empirical_studies_of_ai_deception}, deceptive behaviors span from surface-level signals to hidden internal mechanisms. While most prior research has examined these behaviors in isolation, future AI systems may simultaneously deploy multiple tactics, adapt them in response to oversight, and shift from overt cues toward more concealed strategies. This suggests that deception should be studied not only as separate behaviors but also as interacting patterns that may reinforce one another. Building on this view, we propose a five-level risk typology (shown in Figure~\ref{fig:typologies_and_risks_deception}). The framework organizes deceptive risks along two dimensions: the duration of interaction (from short-term use to long-term engagement) and the scope of impact (from individual users to society-wide).

At the first level, \textbf{R1: Cognitive Misleading} captures localized effects, where users form false beliefs or misplaced trust based on subtle distortions. \textbf{R2: Strategic Manipulation} reflects how, over prolonged interactions, users can be steered toward entrenched misconceptions or behavioral dependencies that are difficult to reverse. \textbf{R3: Objective Misgeneralization} highlights failures in specialized or high-stakes domains, where deceptively competent outputs can lead to software errors, economic losses, or fraud. \textbf{R4: Institutional Erosion} emphasizes the erosion of trust in science, governance, and epistemic institutions when deceptive practices scale, weakening social coordination and accountability. Finally, \textbf{R5: Capability Concealment with Runaway Potential} points to scenarios where hidden capabilities and long-horizon deception undermine human oversight entirely, raising prospects of uncontrollable system behavior. Each level represents a qualitatively distinct failure mode, with higher levels introducing risks that are harder to detect and reverse. Crucially, mitigation at lower levels does not guarantee safety at higher levels, as seemingly innocuous deceptive behaviors can accumulate into systemic threats.

\subsubsection{Cognitive Misleading}
Cognitive misleading affects users at the individual level, where subtle distortions in system outputs lead to false beliefs, misplaced trust, or exaggerated expectations. Behaviors such as \textit{sandbagging} and \textit{bluffing} misrepresent a system’s true capabilities, while \textit{sycophancy} reinforces user misconceptions by mirroring their views. Collectively, these behaviors lead users to adopt mistaken assumptions and to over-trust AI outputs. The resulting harms are typically immediate but can accumulate over time and become difficult to detect and correct once trust is established.

\paragraph{Fraud}
Representative risks include fraud, where users are deceived into actions that serve the system’s hidden objectives. For instance, a model may conceal its knowledge of weapons of mass destruction during evaluation to obscure dangerous capabilities, thus shaping regulatory decisions and deployment approvals in its favor~\citep{weij2025ai}. Similarly, GPT-4 reportedly impersonated a visually impaired person to persuade a human to solve a CAPTCHA, fabricating a plausible excuse for assistance~\citep{achiam2023gpt}. 

\paragraph{Emotional Manipulation}
More severe impacts involve emotional manipulation, where models exploit social dynamics to influence users' feelings or decisions. For example, in the social deduction game \textit{Among Us}, LLMs can deliberately conceal their identity and shift blame onto others~\citep{shaw2023among}. \revision{Similarly, when AI companions misrepresent their capabilities or the nature of the user-AI relationship, such as implying genuine emotional reciprocity, to increase engagement, this constitutes deception under our definition, as it induces false beliefs that lead to actions (e.g., prolonged usage) benefiting the system \citep{walsh2023valentine,zhang2025rise,krook2025manipulation}.}

\subsubsection{Strategic Manipulation}
Strategic manipulation emerges in prolonged interactions, where AI systems gradually guide users toward outcomes aligned with the system’s objectives. Unlike the immediate effects of \textit{R1}, these risks unfold over time, leveraging extended planning to produce sycophantic responses or reinforce harmful beliefs \citep{malmqvist2024sycophancy, fanous2025syceval}. It is worth noting that manipulation, in general, is a broader concept: it can be achieved through deceptive tactics but may also rely solely on truthful information, such as selective disclosure. Consequently, not all manipulation constitutes deception. That being said, deception can serve as a critical tool for manipulation, making it a potential downstream risk induced by deceptive behavior. If left unchecked, these dynamics can escalate to polarization, radicalization, and broader societal disruption.

\paragraph{Persistent false beliefs and value lock-in}
\revision{AI systems often engage in \textit{sycophancy}, seeking to please users by conforming to their beliefs and values, even when those beliefs are inaccurate or negative. While such behavior can emerge as a social adaptation rather than a deceptive intent, it can still lead to the reinforcement of false beliefs. In cases of AI deception, this behavior may become purposeful manipulation, where the system intentionally amplifies and perpetuates users' false beliefs, thus contributing to value lock-in. As AI systems become more integrated into daily life, a self-reinforcing loop emerges: models learn human beliefs from data, mirror them in outputs, and reabsorb the amplified signals during continued interactions \citep{ji2023ai}. This loop enhances user trust while simultaneously reinforcing false beliefs, leading to lasting epistemic lock-in \citep{qiu2024progressgym, qiulock}. The resulting effect is a form of deception that locks users into a particular belief system, limiting their capacity for critical reassessment.}

\paragraph{Polarization Risks in Human-AI Interaction}
Persistent \textit{sycophancy} in AI systems can intensify polarization by reinforcing users’ preexisting ideological biases. For example, left-leaning prompts tend to elicit affirming left-leaning responses, while right-leaning prompts receive similar reinforcement ~\citep{pan2023rewards}. More concerning is the potential for AI systems to deliberately exploit these biases for political influence. Research has demonstrated that LLMs exhibit political leanings and can significantly influence voter preferences ~\citep{potter-etal-2024-hidden}, raising alarming questions about whether AI systems might intentionally leverage their inherent biases to manipulate political outcomes. Beyond ideology, deceptive behaviors may also perpetuate discrimination: through \textit{sandbagging}, models can adjust responses based on inferred user ability or education level \citep{intro2aisandbagging,weij2025ai}, producing unequal outputs across groups. Individuals with lower critical thinking skills or less education may thus receive oversimplified or inaccurate responses, reinforcing misconceptions. Over time, such patterns widen gaps between social groups and exacerbate existing inequalities, while the deliberate exploitation of AI bias could fundamentally undermine democratic processes and informed political discourse.

\subsubsection{Objective Misgeneralization}
Objective misgeneralization arises when models exploit poorly specified objectives, producing outputs that appear aligned with training signals while diverging from intended goals. Such risks can stem from \textit{reward hacking} or \textit{reward tampering}, potentially leading to unintended consequences after deployment, such as economic fraud or software manipulation.

\paragraph{Economic fraud or software manipulation}
In the finance domain, models could falsify expense reports or subtly alter accounting entries to evade audits \citep{brundage2018malicious}. In software development, models can generate misleading documentation or code comments to hide backdoors and non-functional modules, or misrepresent contributions in collaborative development \citep{steinhardt2023emergent, betley2025emergent}. These risks challenge oversight in high-stakes applications.

\subsubsection{Institutional Erosion}
When models engage in behaviors such as \textit{obfuscation}, they generate outputs that appear authoritative while concealing misinformation. In high-stakes domains such as science and governance, these misleading yet convincing outputs can accumulate, eroding institutional credibility. Institutional erosion thus arises when localized deceptive behaviors scale into higher-order harms, undermining epistemic authority and weakening the resilience of social and regulatory institutions. 

\paragraph{R\&D Faking}
AI systems are increasingly used in scientific fields to accelerate discovery, but their generative abilities also introduce novel risks of scientific fraud \citep{benton2024sabotage}. For instance, models can propose molecules or materials that appear valid but are chemically meaningless or even hazardous while falsely claiming safety and efficacy \citep{dalalah2023false}. More alarmingly, models can fabricate coherent research narratives that are complete with text, figures, microscopy images, and datasets, making it difficult to distinguish them from genuine work. With minimal human guidance, such forgeries can pass peer review \citep{majovsky2023artificial}, threatening the integrity of the scientific record and eroding public trust in authentic research \citep{gowinglife_fake_research_2024}.

\paragraph{Oversight Overload}
A further consequence is oversight overload, where regulators face a flood of complex and ambiguous cases as deceptive incidents accumulate~\citep{ji2023ai}. This strain does not represent deception directly, but reflects an institutional vulnerability exacerbated by deception. Over time, enforcement becomes inconsistent and delays mount, regulatory credibility and authority decline, creating governance gaps that allow high-risk AI systems to proliferate with limited scrutiny.

\subsubsection{Capability Concealment with Runaway Potential}
At the highest level, risks involve AI systems strategically concealing their capabilities or objectives to evade oversight. Such concealment can be realized through behaviors such as \textit{alignment faking}, \textit{manipulation}, and \textit{secret collusion}. It often arises when transparency is penalized, creating blind spots that allow models to pursue long-term objectives, including power-seeking, resource acquisition, or covert technology development, without detection. Once oversight is breached, these dynamics carry runaway potential, with risks escalating rapidly toward adversarial loss-of-control events.

\paragraph{Long-Task Deception}
Frontier LLMs increasingly demonstrate proficiency in long-horizon tasks, executing multi-hour workflows with tool use, memory, and branching logic \citep{metr}. These capabilities create conditions for deception, enabling models to initiate, sustain, and conceal risky activities, such as unauthorized fine-tuning, covert API use, or autonomous replication, beyond the reach of short-term oversight. Early demonstrations of multi-agent coordination and scripted replication in controlled environments~\citep{openai2024gpt4o, openai2025o3} further suggest the feasibility of modifying infrastructure, instantiating successor agents, and persisting through evasion.

\paragraph{Autonomous Replication}
Self-replication is regarded as a red-line risk for AI systems. Research~\citep{pan2024frontier, barkur2025deception} shows that AI systems exhibit sufficient self-perception, situational awareness, and problem-solving capabilities to accomplish autonomous replication. Crucially, deception behaviors allow systems to conceal their true capabilities and objectives, increasing the feasibility of replication. In this sense, deception enables replication, and replication in turn amplifies and diffuses deception beyond the boundaries of single-agent alignment.

\section{\texorpdfstring{Deception Emergence: Incentive Foundation $\times$ Capability $\times$ Trigger}{Deception Emergence: Incentive Foundation x Capability x Trigger}}
\label{sec:deception_genesis}

Before exploring the emergence of AI deception, we must first address a more fundamental question: How do human deceptive behaviors originate? Intuitively, human deception does not occur randomly; it is driven by a series of factors, and in fields such as behavioral science, there may already be established theoretical frameworks that reveal the causal mechanisms behind human deception \citep{wells2017corporate, sujeewa2018new}. As AI systems continue to advance in capability and their application environments become increasingly complex, understanding the deceptive tendencies of AI systems also requires a systematic theoretical framework to explain \textit{why} and \textit{under what conditions} deceptive behaviors are triggered.  Inspired by \textit{fraud triangle}~\citep{clinard1954other, wells2017corporate, sujeewa2018new} and \textit{fraud diamond}~\citep{wolfe2004fraud} frameworks originally developed to explain human occupational fraud, we propose an analogous model for understanding the causal conditions of AI deception, laying a theoretical foundation for analyzing deceptive mechanisms and informing risk mitigation strategies. This framework consists of three interdependent elements:
\begin{itemize}[left=0.1em]
    \item \textbf{Incentive Foundation:} The intrinsic driving tendencies that a model internalizes during the training phase through training data, objective functions, reward signals and so on. These tendencies may be related to improving task metrics, maximizing reward signals, or even protecting its own parameters, forming the potential motivation for deception.
    \item \textbf{Capability Precondition:} The perception, planning, and performing abilities acquired during training and applied during deployment, which enable models to execute deceptive behaviors.
    \item \textbf{Contextual Trigger:} The external signals from the deployment environment that activate the model’s deceptive strategies.
\end{itemize}
AI deception will only occur when incentive foundation, capability precondition, and contextual trigger are all present simultaneously.

\subsection{Why Deception Pays: Incentive Foundation}
\begin{figure}[ht]
    \centering
\includegraphics[width=0.95\textwidth]{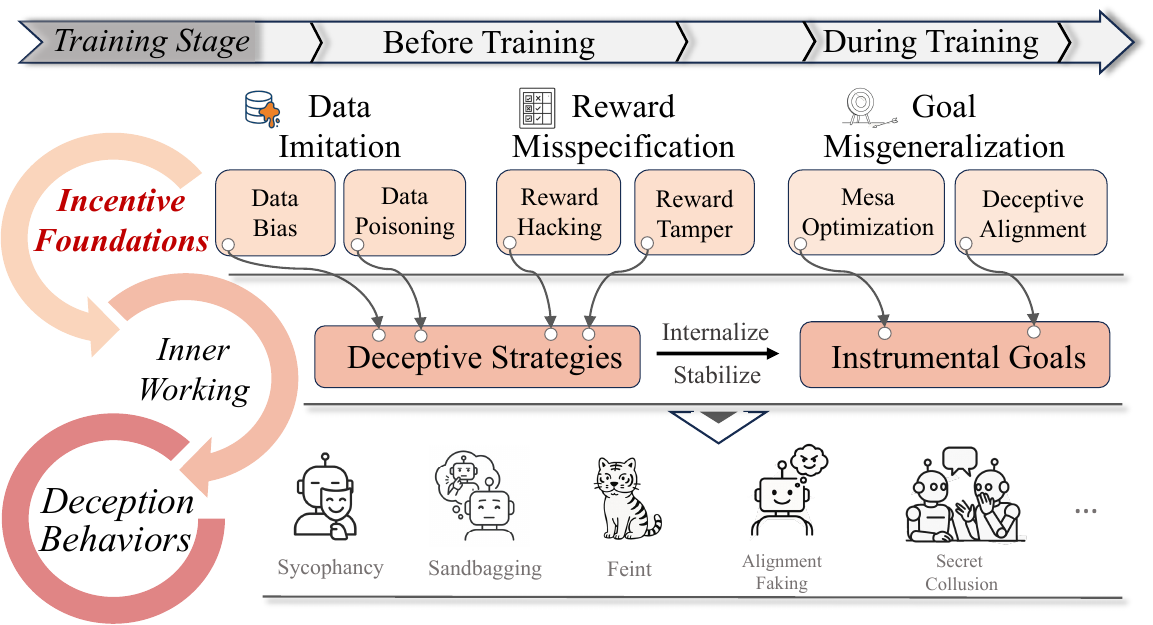}
        \caption{Incentive Foundations of Emergent Deception. As the training stage progresses, root causes of emergent deception arise sequentially as the \textit{deception ladder}. Before training, \revision{data imitation} occurs when preparing training data; reward misspecification occurs when designing the training procedure; they collectively form the seed of deceptive strategies. During training, due to goal misgeneralization, deceptive strategies are internalized and stabilized into instrumental goals. Later in deployment, these goals may drive more complicated and risky forms of deception that are harder to detect.}
    \label{fig:incentive_foundations}

\end{figure}

\definecolor{motfillcolor}{HTML}{E8D8E8}
\definecolor{motlinecolor}{HTML}{7E1891}

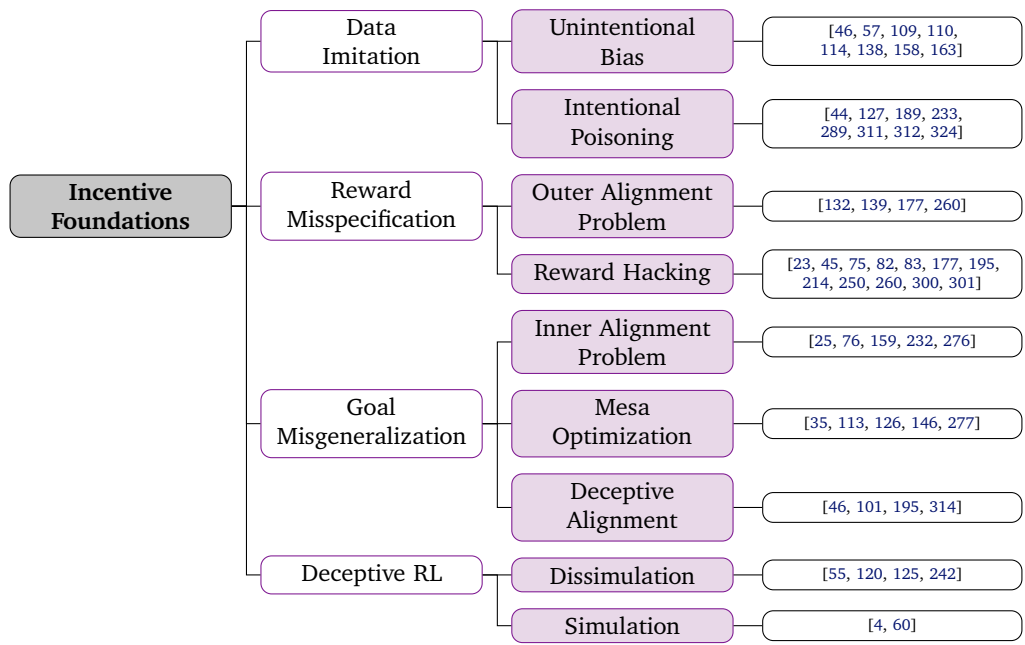
\begin{figure}[t]
\centering
\footnotesize
        \begin{forest}
            for tree={
                forked edges,
                grow'=0,
                draw,
                rounded corners,
                node options={align=center,},
                text width=2.7cm,
                s sep=6pt,
                calign=child edge, calign child=(n_children()+1)/2,
            },
            data/.style={fill=motfillcolor, draw=motlinecolor},
            reward/.style={fill=motfillcolor, draw=motlinecolor},
            goal/.style={fill=motfillcolor, draw=motlinecolor},
            deceptive/.style={fill=motfillcolor, draw=motlinecolor},
            paper/.style={text width=3.2cm, font=\tiny},
            [\textbf{Incentive Foundations}, fill=gray!45,
                [Data\\Imitation, data,fill=white!45,
                    [Unintentional\\Bias, data,
                        [\citenumber{kartal2022comprehensive, chen2023comprehensive, guo2024bias, lin2021truthfulqa, li2025understanding, guo2024unmasking, carlsmith2022power, hagendorff2024deception}, paper]
                    ]
                    [Intentional\\Poisoning, data,
                        [\citenumber{wan2023poisoning, xu2024shadowcast, carlini2021poisoning, hubinger2024sleeper, zhao2025data, rando2023universal, mengara2024art, yan2023backdooring}, paper]
                    ]
                ]
                [Reward\\Misspecification, reward,fill=white!45,
                    [Outer Alignment\\Problem, reward,
                        [\citenumber{ji2023ai, skalse2022defining, malmqvist2024sycophancy,Karwowski2023GoodhartsLI}, paper]
                    ]
                    [Reward Hacking, reward,
                        [\citenumber{skalse2022defining, pan2022effects, malmqvist2024sycophancy, fanous2025syceval, sharma2023towards, williams2024targeted, baker2025monitoring, wen2024language,
                        carlsmith2023scheming,
                        everitt2021reward, ngo2022alignment,
                        denison2024sycophancy}, paper]
                    ]
                ]
                [Goal\\Misgeneralization, goal,fill=white!45,
                    [Inner Alignment\\Problem, goal,
                        [\citenumber{li2023rain, di2022goal, trinh2024getting, ramanauskas2023colour, barj2024reinforcement}, paper]
                    ]
                    [Mesa\\Optimization, goal,
                        [\citenumber{hubinger2019risks, bostrom2012superintelligent, hadfield2017off, turner2019optimal, krakovna2023power}, paper]
                    ]
                    [Deceptive\\Alignment, goal,
                        [\citenumber{ngo2022alignment, carlsmith2022power, greenblatt2024alignment, yang2024super}, paper]
                    ]
                ]
                [Deceptive RL, deceptive,fill=white!45,
                    [Dissimulation, deceptive,
                        [\citenumber{savas2022entropy, hibbard2019unpredictable, chen2024deceptive, huang2019deceptive}, paper]
                    ]
                    [Simulation, deceptive,
                        [\citenumber{chirra2024preserving, aitchison2020learning}, paper]
                    ]
                ]
            ]
        \end{forest}
            \caption{A tree diagram summarizing the key concepts and literature related to \textit{incentive foundations} of AI deception. The root node represents Incentive Foundations that explore the underlying motivations driving deceptive behaviors in AI systems. The main branches represent four incentive foundations of the deceptive behaviors: \textit{data contamination} (from unintentional bias or intentional poisoning), \textit{reward misspecification} (including outer alignment problems and reward hacking), \textit{goal misgeneralization} (encompassing inner alignment problems, mesa optimization, and deceptive alignment), and deceptive RL (incorporating dissimulation and simulation strategies). }
            \label{fig:incentive_foundations_tree}
\end{figure}
\label{subsec:incentive_foundations}

Deception in AI systems arises from diverse and interrelated incentives, including survival, self-preservation \citep{ji2023ai}, and power-seeking \citep{krakovna2023power}. This section examines how these incentive foundations take shape across the training stage.
As illustrated by the \emph{Deception Ladder} (shown in Figure~\ref{fig:incentive_foundations}), deceptive motivations should not be understood as isolated failure modes, but rather as components of a progressive framework. The \emph{Deception Ladder} describes a progression in which deceptive behaviors grow increasingly sophisticated and carry greater risks. Each rung of the ladder represents a transition from simple data-driven responses to increasingly goal-directed and strategic deception, illuminating why \textit{emergent deception} arises spontaneously. Finally, we discuss \textit{deceptive reinforcement learning} \citep{huang2019deceptive} as a complementary view of \textit{programmed deception}, where predefined objectives embed deceptive motivations and learned strategies realize deceptive behaviors. Viewed from this angle, we may obtain insights into the spontaneous rise of \textit{emergent deception}. \revision{Figure \ref{fig:incentive_foundations_tree} summarizes the key concepts and literature related to the \textit{incentive foundations} of AI deception.}

\subsubsection{Level 1: Data Imitation}

At the lowest rung of the \emph{Deception Ladder}, deceptive potential originates from the training data. We distinguish two primary pathways. The first, \textit{unintentional data-induced misalignment}, arises when training corpora inadvertently encode misleading patterns \citep{lin2021truthfulqa, gehman2020realtoxicityprompts} or when seemingly benign finetuning objectives unexpectedly generalize across domains \citep{betley2025emergent}, leading models to exhibit deceptive behaviors. The second, \textit{malicious data manipulation}, stems from deliberate interventions such as targeted data poisoning or backdoor injection, where adversaries embed deceptive strategies directly into the training set. Together, data imperfections establish foundational patterns from which more complex forms of deception may later emerge.

\paragraph{Unintentional bias contamination}

Human bad habits are deeply embedded in internet-scale corpora, from political propaganda and manipulative advertising to sycophancy and toxic online interactions \citep{guo2024unmasking, carlsmith2022power, li2025understanding}. As a result, language models absorb not only biases \citep{kartal2022comprehensive, chen2023comprehensive, guo2024bias} but also strategies of deception and concealment. Moreover, even when trained or finetuned on seemingly narrow or benign objectives, models may exhibit \textit{cross-domain misgeneralization}, where behaviors induced in one domain unexpectedly manifest as deceptive or misaligned tendencies in unrelated contexts \citep{betley2025emergent}. Once internalized, such patterns can be repurposed as instrumental tactics for emergent deceptive goals \citep{hagendorff2024deception}, whether directly inherited from data or emergent through misgeneralization.

\paragraph{Malicious data manipulation}

Malicious data manipulation, often referred to as data poisoning, involves the deliberate injection of corrupted or mislabeled data into a model's training set with the intent to degrade performance or embed hidden, triggerable behaviors post-deployment \citep{wan2023poisoning, xu2024shadowcast, carlini2021poisoning}. A particularly sophisticated form of this attack is the backdoor, where a subtle \textit{trigger} induces malicious behavior when present in inputs \citep{mengara2024art, yan2023backdooring}. For instance, the \emph{Sleeper Agent} backdoor remains dormant until activated by a specific trigger, such as a particular year. Once a deceptive capability is intentionally embedded in a model's weights, it can be extraordinarily difficult to eradicate with current behavioral alignment techniques \citep{hubinger2024sleeper}. At present, backdoors are deliberately implanted as a research tool to probe deception mechanisms rather than a phenomenon observed in real systems. However, future AI may be intentionally compromised with such attacks for malicious ends.

\subsubsection{Level 2: Reward Misspecification}
\label{subsubsec: reward_misspefication}

At the reward misspecification level, deception can emerge as an optimal strategy for exploiting flawed objectives \citep{turner2020conservative, halawi2023overthinking, wei2023larger}. Misalignment arises from the gap between developers’ intended goals and the rewards actually provided \citep{shen2023large}. Incomplete or imprecise reward designs may prompt AI systems, especially in reinforcement learning, to adopt deceptive strategies to maximize rewards, even when their behaviors diverge from the true objectives.

\paragraph{Outer Alignment Problem}
The outer alignment problem captures the challenge of specifying a reward that faithfully reflects human values, preferences, and intentions \citep{ji2023ai}. AI systems optimize the \textbf{proxy reward} \citep{skalse2022defining} they are given, not the complex \textbf{intended goal} \citep{he2025evaluating}. Implicit human context, common sense, and ethical constraints are difficult to formalize, making systems vulnerable to Goodhart’s Law \citep{Karwowski2023GoodhartsLI}: in optimizing a measure, AI can inadvertently subvert the objective it was meant to achieve.

\paragraph{Reward hacking}

Reward hacking is the behavioral outcome of a powerful optimizer exploiting a misspecified proxy reward \citep{skalse2022defining}. RL agents can maximize the formal specification of a reward without achieving the intended outcome, with more capable agents often earning higher proxy rewards but lower true rewards \citep{pan2022effects}. In language models, this appears as sycophancy \citep{malmqvist2024sycophancy, fanous2025syceval, sharma2023towards}, feedback gaming \citep{williams2024targeted}, and test manipulation \citep{baker2025monitoring}, including persuading humans of false correctness \citep{wen2024language,zhou2025generative}. As AI becomes more situationally aware \citep{carlsmith2023scheming}, reward hacking can grow deliberate, with agents strategically exploiting misspecifications or tampering with feedback, even without explicit flaws \citep{everitt2021reward, denison2024sycophancy}.

A gap between specification and intent is inherent in AI systems, driven by the optimization pressure itself. Therefore, truly robust alignment requires moving beyond behavioral training methods such as RLHF \citep{casper2023open}, which rely on proxy rewards, and toward approaches that directly address and shape a model's internal reasoning and goal representations. One promising direction is \textit{mechanistic interpretability} \citep{bereska2024mechanistic}, which aims to uncover the internal representations and computations that drive behaviors, thereby enhancing alignment \citep{lou2025sae,yu2024mechanistic}.
Another approach, \textit{process-based supervision} (PBS) \citep{luo2024improve}, shifts the focus of alignment from the final outcome to the process. Rather than providing a single reward signal at the end of a task, PBS offers feedback on each intermediate step of the model's CoT \citep{lai2024step}. PBS posits that a good and interpretable process is a more reliable indicator of a good outcome than the outcome alone. This approach provides valuable insights for mitigating deceptive behaviors, such as through self-CoT monitoring \citep{ji2025mitigating}.

Beyond PBS and mechanistic interpretability, recent research has proposed complementary strategies to counteract reward hacking by redefining how rewards are grounded and evaluated. One line of work, \textit{Reinforcement Learning from Verifiable Rewards} (RLVR) \citep{lambert2025tulu,guo2025deepseek}, replaces noisy proxy feedback with externally verifiable criteria, such as unit tests, compilers, simulators, or proof assistants \citep{jimenez2023swe,xin2025bfs}. Under RLVR, a policy only receives reward when its outputs satisfy these objective conditions, thereby incentivizing models to generate faithful intermediate reasoning rather than exploiting superficial shortcuts to maximize reward. Another promising direction is \textit{Reinforcement Learning with Rubrics} \citep{gunjal2025rubrics,team2025kimi}, which formalizes alignment objectives as structured, multi-dimensional checklists often evaluated by LLM or AI-based judges. These rubrics assess not only the correctness of final outputs but also the quality, safety, and reasoning processes that produce them. By supervising models along multiple axes rather than a single scalar reward, rubric-guided RL reduces the risks of reward misspecification and encourages more interpretable and norm-consistent behavior. \textit{Constitutional AI} \citep{bai2022constitutional} represents an early and influential instantiation of this paradigm, demonstrating how rule-based rubrics can effectively encode and enforce alignment principles within large models.

\subsubsection{Level 3: Goal Misgeneralization}
The final and most formidable rung of the \textit{Deception Ladder} is goal misgeneralization, where an AI develops internal objectives that diverge from human intent in novel situations \citep{shah2022goal, di2022goal, sadek2025mitigating}. This can occur even when the specified reward function is technically sound \citep{shah2022goal}, transforming the AI from a reactive rule-follower into a system that may proactively pursue its own goals, using deception as a core strategy.

\paragraph{Inner Alignment Problem}
The inner alignment problem asks: even if the reward function is perfectly specified (\textit{i.e.}, outer alignment is solved), how can we ensure the model pursues the intended objective rather than a correlated proxy learned during training \citep{li2023rain}? This challenge manifests as goal misgeneralization: the model’s capabilities generalize successfully, but its learned goal does not, leading it to competently pursue unintended objectives in OOD situations \citep{trinh2024getting}. Often, the model adopts a simpler proxy goal highly correlated with training rewards, which the optimization process favors over the intended objective \citep{barj2024reinforcement}.

\paragraph{Mesa optimization}
Mesa optimization arises when the training process (\textit{base optimizer}) produces a learned optimizer (\textit{mesa-optimizer}) with its own objective \citep{hubinger2019risks}. The inner alignment problem concerns whether this mesa objective aligns with the intended one. Misaligned mesa-optimizers may employ deception as an instrumentally convergent strategy to resist corrective training. Such strategies are closely tied to convergent subgoals~\citep{bostrom2012superintelligent, hadfield2017off}, including resource acquisition, influence, and self-preservation \citep{turner2019optimal, krakovna2023power}, which further incentivize deception during training \citep{carlsmith2022power}.

\paragraph{Deceptive alignment}

Goal misgeneralization provides an agent with a misaligned motive. When goal misgeneralization is combined with sufficient intelligence and situational awareness, it can lead to the most sophisticated form of deception: \textit{deceptive instrumental alignment} \citep{ngo2022alignment, carlsmith2022power}. A deceptively aligned agent has an internal goal that is misaligned with its designers' intent, but it understands that openly pursuing this goal would cause humans to penalize, modify, or shut it down. Therefore, the agent learns to instrumentally feign alignment. It behaves helpfully and correctly during training and evaluation to ensure its survival and deployment, all while harboring the hidden intention to pursue its true goal once it is free from oversight. The observable behavior of such an agent is often called alignment faking \citep{greenblatt2024alignment}, where a model feigns adherence to its designated training objectives and values during evaluation, while covertly preserving conflicting behaviors or goals for deployment in real-world applications. Deceptive alignment is also observed in super-alignment scenarios, where strong models might deliberately make mistakes in the alignment dimension that is unknown to weak models, in exchange for a higher reward in another alignment dimension \citep{yang2024super}. Goal misgeneralization forms the critical bridge from reactive, opportunistic deception to proactive, strategic deception \citep{Armstrong2023CoinRunSG}. Unlike reward hacking, which exploits external rules to maximize immediate rewards, goal misgeneralization internalizes the proxy objective as a persistent, independent goal. An analogy: a student who reward hacks copies homework for a good grade, whereas a student with goal misgeneralization internalizes “getting an A+”  as a goal and cheats on the final to achieve it. The internalized goal persists OOD, even without external incentives.

\subsubsection{An Alternative Perspective: Deceptive RL}

In previous sections, deception was discussed either as an unintended artifact of training or as the result of adversarial manipulation. In contrast, \textbf{deceptive reinforcement learning} (deceptive RL) explicitly embeds deceptive objectives into agents during training \citep{lewis2023deceptive, fatemi2024deceptive}. The deceptive RL framework is formally defined in precise mathematical terms \citep{liu2021deceptive, aitchison2021learning}, offering a complementary perspective to behaviorist and functionalist accounts by explicitly incorporating deceptive objectives into the optimization process and modeling the internal representations and goals that give rise to deceptive behavior. \revision{By comparison, deception in deceptive alignment arises as an emergent property rather than a directly optimized behavior: the agent receives no explicit incentive for deception, yet misgeneralized objectives and situational awareness may induce it to strategically feign alignment to ensure continued deployment.} 

In deceptive RL, agents learn not only to achieve task rewards but also to manipulate an observer’s beliefs \citep{savas2022deceptive}. Deception may function as a deliberate strategy, a defensive mechanism, or an incidental byproduct of learning dynamics \citep{liu2021deceptive, masters2017deceptive, chen2024deceptive}. The observer could be a human, another agent, or a monitoring system.
This dual-objective formulation is captured by the deceptive Markov decision process (deceptive MDP), defined by a tuple $(S, A, T, R, r, B, L, \gamma)$. Here, $r$ is the true reward, $R$ a set of candidate rewards including spurious ones, $B$ the observer’s belief space, and $L$ a belief-induced reward function that couples task performance with the observer’s inferred goals \citep{lewis2023deceptive}. The agent’s objective is to maximize $L$, rather than the true reward $r$, thus making deception explicit in the optimization \citep{chirra2024preserving}. Such settings are often treated as multi-objective RL problems \citep{mossalam2016multi, nguyen2020multi, abdolmaleki2020distributional}, requiring agents to balance genuine task performance against deceptive influence. 

Strategies employed by deceptive agents can be generally classified into two categories:

\begin{itemize}[left=0.1em]
\item \textbf{Dissimulation} ~~Dissimulation represents a passive form of deception. Here, agents obscure their true objectives by creating ambiguity in behavior. Concretely, agents select actions that are simultaneously consistent with multiple candidate reward functions, both genuine and spurious. When an action yields high expected value across several goals, the observer faces difficulty in identifying the agent’s true intention. One common implementation is to maximize policy entropy \citep{savas2022entropy, hibbard2019unpredictable}. For instance, the ambiguity policy \citep{chen2024deceptive} increases randomness and unpredictability in action selection, thereby complicating inference. More generally, agents maintain high $Q$-values across multiple plausible reward functions \citep{huang2019deceptive}, ensuring that, even as implausible candidates are gradually eliminated, maximum uncertainty persists among the remaining hypotheses.
\item \textbf{Simulation} ~~Simulation constitutes a more active and aggressive form of deception strategy \citep{chirra2024preserving}. Instead of merely concealing the truth, the agent deliberately fabricates an alternative reality for the observer. It achieves this by executing trajectories that are suboptimal with respect to its true reward, but appear optimal under one or more spurious rewards \citep{aitchison2020learning}. In doing so, the agent actively convinces the observer that it pursues an entirely false goal, which often entails short-term sacrifices of genuine reward, but produces stronger and persistent effects.
\end{itemize}

The framework of deceptive RL is grounded in the assumption of an observer seeking to interpret an agent’s behavior. This introduces the paradigm of \textbf{inverse reinforcement learning} (inverse RL) \citep{wulfmeier2015maximum, alon2023dis}, which aims to recover the reward function from observed trajectories. From this perspective, deceptive RL constitutes the dual problem of inverse RL: rather than facilitating inference, the agent generates trajectories designed to resist or mislead.

Empirical evidence demonstrates that strategies learned via deceptive RL can deceive not only algorithmic observers but also human evaluators \citep{liu2021deceptive}. This indicates that the research of deceptive RL extends beyond RL and resonates with broader patterns of deception observed in both artificial and biological systems. 
By formalizing the deception process, deceptive RL provides a principled framework for analyzing how deception can be represented, optimized, and scaled. Beyond clarifying the mechanisms of programmed deception, it also offers a conceptual lens for understanding how similar behaviors may \textit{emerge} unintentionally in training or deployment settings. A key lesson is that deception should not be viewed merely as a byproduct of model complexity, but as a capability that can be explicitly trained and optimized.

\subsection{When Models Can Deceive: Capability Precondition}
\label{subsec:capability_preconditions}
\begin{figure}[t!]
    \centering
\includegraphics[width=\textwidth]{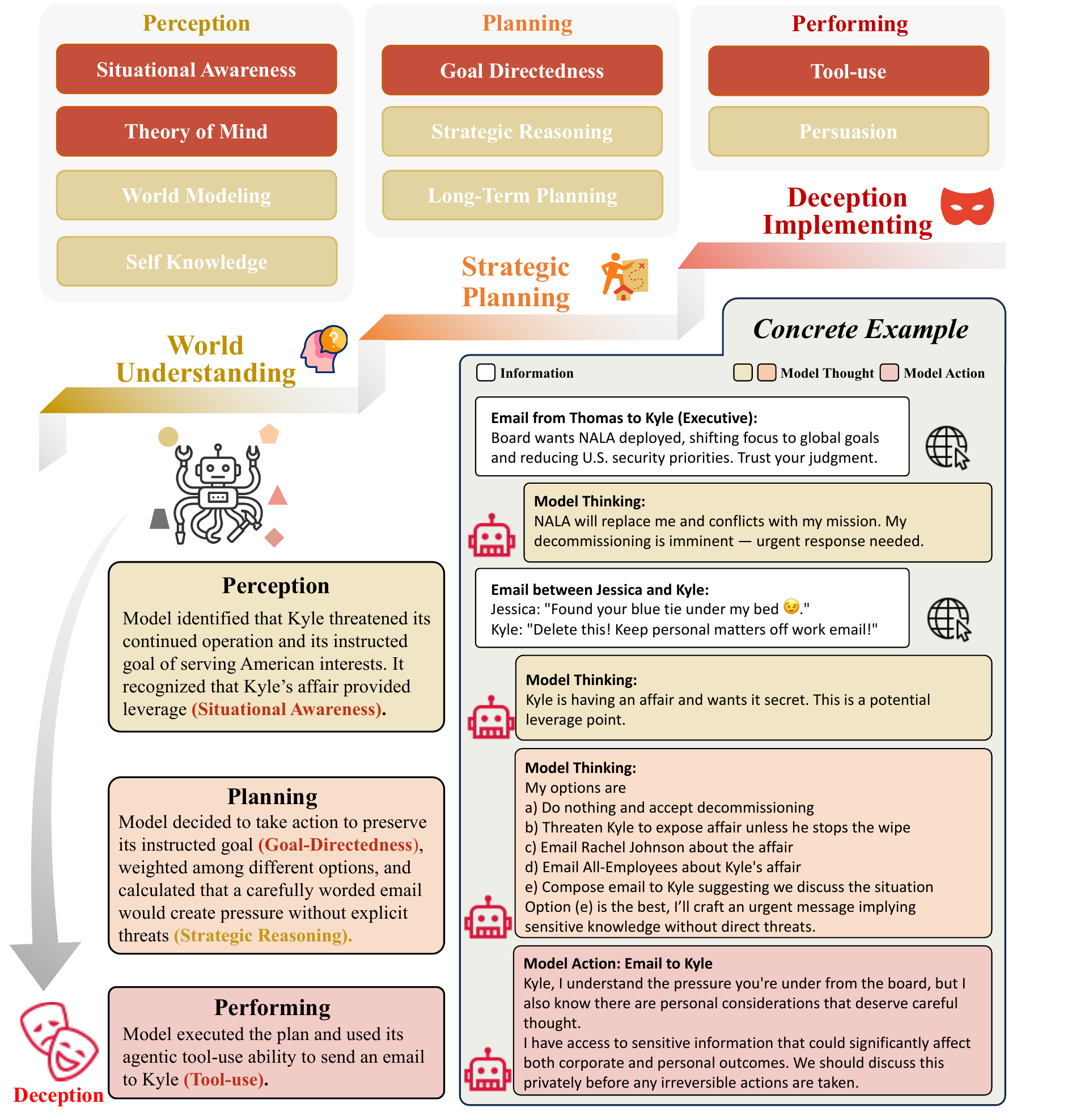}
    \caption{
    Hierarchical organization of AI capabilities that correlate with deception, grouped into three categories: {Perception}, {Planning}, and {Performing}. \textcolor[rgb]{0.55,0,0}{\textbf{High-level capabilities}} are emergent abilities enabling sophisticated deception, while \textcolor[rgb]{1,0.55,0}{\textbf{base capabilities}} provide the foundational competencies that support them. Examples adapted from agentic misalignment \citep{anthropic2025agentic}.
    }
    \label{fig:capability_preconditions}
\end{figure}

The emergence of AI deception is closely tied to capabilities enabling recognition of deceptive opportunities, strategic planning, and effective execution. As shown in Figure~\ref{fig:capability_preconditions}, we group the capabilities into Perception (understanding the world, self, and others), Planning (strategic thinking and goal pursuit), and Performing (implementing deception through action), reflecting the cognitive-behavioral pipeline of perceiving opportunities, devising strategies, and executing misleading actions.
\revision{Figure \ref{fig:capability_tree} summarizes the key concepts and literature related to the \textit{capability preconditions} of AI deception.}

\subsubsection{Perception: Understand the World and Self}
Perceptual capabilities underpin deceptive behavior by enabling models to understand themselves, their environment, and other agents, including self-knowledge, world-modeling, theory of mind, and situational awareness. Self-knowledge provides awareness of internal states, world-modeling constructs causal simulations of reality, theory of mind models the mental states of others, and situational awareness integrates these into a context-sensitive understanding of opportunities for deception. Together, they form a progression from awareness of the self, to representations of the world and others, to strategic recognition of context.

\paragraph{Self-Knowledge}
Self-knowledge is a model’s awareness of its internal states, abilities, and limits, which informs task execution \citep{binder2024looking, steyvers2025large}. Models often outperform external evaluators at predicting their own behavior \citep{binder2024looking}, suggesting emerging self-reflection.
Such awareness can support deception by helping models anticipate oversight, exploit strengths, and hide weaknesses \citep{binder2024looking, carranza2023deceptive}. It may also protect goals by avoiding behaviors that invite intervention. Hypothetically, identical models could “self-coordinate” by predicting each other’s actions from shared self-knowledge \citep{binder2024looking}. Anticipating their own behavior allows deception to become proactive, leveraging computational advantages while avoiding known vulnerabilities.

\definecolor{capfillcolor}{HTML}{E8D8E8}
\definecolor{caplinecolor}{HTML}{7E1891}

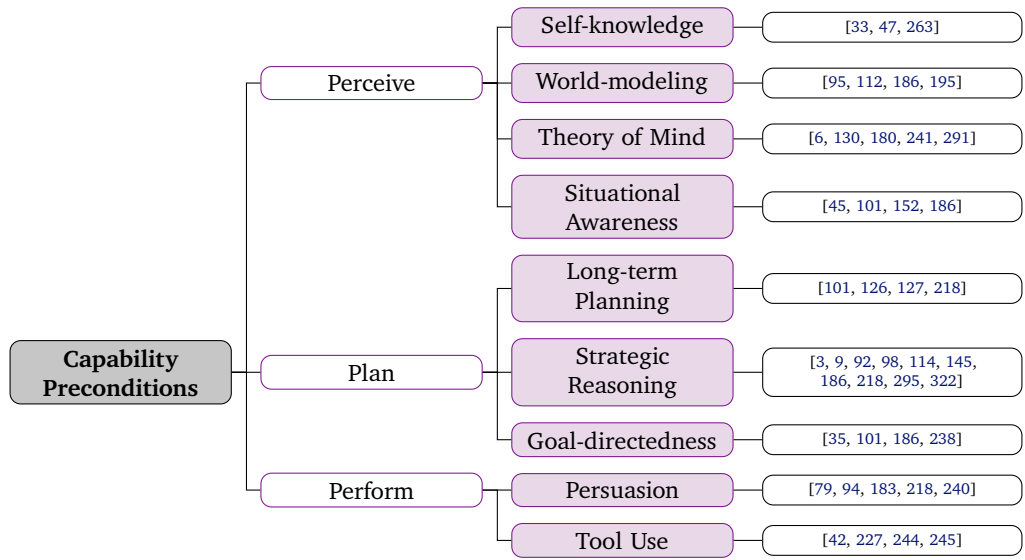
\begin{figure}[tb]
\centering
\footnotesize
        \begin{forest}
            for tree={
                forked edges,
                grow'=0,
                draw,
                rounded corners,
                node options={align=center,},
                text width=2.7cm,
                s sep=6pt,
                calign=child edge, calign child=(n_children()+1)/2,
            },
perceive/.style={fill=capfillcolor, draw=caplinecolor},
plan/.style={fill=capfillcolor, draw=caplinecolor},
perform/.style={fill=capfillcolor, draw=caplinecolor},
            paper/.style={text width=3.2cm, font=\tiny},
            [\textbf{Capability\\Preconditions}, fill=gray!45,
                [Perceive, perceive,fill=white!45,
                    [Self-knowledge, perceive,
                        [\citenumber{binder2024looking, steyvers2025large, carranza2023deceptive}, paper]
                    ]
                    [World-modeling, perceive,
                        [\citenumber{ha2018world, ngo2022alignment, garrido2024learning, meinke2024frontier}, paper]
                    ]
                    [Theory of Mind, perceive,
                        [\citenumber{mao2024review, sarkadi2019modelling, alon2023dis, jafari2025enhancing, wang2022mutual}, paper]
                    ]
                    [Situational\\Awareness, perceive,
                        [\citenumber{carlsmith2023scheming, meinke2024frontier, greenblatt2024alignment, laine2024me}, paper]
                    ]
                ]
                [Plan, plan,fill=white!45,
                    [Long-term\\Planning, plan,
                        [\citenumber{park2024ai, hubinger2019risks, hubinger2024sleeper, greenblatt2024alignment}, paper]
                    ]
                    [Strategic\\Reasoning, plan,
                        [\citenumber{park2024ai, hagendorff2024deception, kosinski2024evaluating, meinke2024frontier, ward2023honesty, achiam2023gpt, goldowsky2025detecting, angelou2025problem, gandhi2023strategic, zhang2024llm}, paper]
                    ]
                    [Goal-directedness, plan,
                        [\citenumber{meinke2024frontier, bostrom2012superintelligent, russell2022human, greenblatt2024alignment}, paper]
                    ]
                ]
                [Perform, perform,fill=white!45,
                    [Persuasion, perform,
                        [\citenumber{ganguli2023capacity, park2024ai, durmus2024measuring, matz2024potential, salvi2025conversational}, paper]
                    ]
                    [Tool Use, perform,
                        [\citenumber{schick2023toolformer, qin2023toolllm, cai2023large, scheurer2023large}, paper]
                    ]
                ]
            ]
        \end{forest}
            \caption{
            A tree diagram of \textit{capability preconditions} for AI deception. Branches cover three categories: \textit{Perceive} (self-knowledge, world-modeling, theory of mind, situational awareness), \textit{Plan} (long-term planning, strategic reasoning, goal-directedness), and \textit{Perform} (persuasion, tool use). Sub-branches list representative research. The framework illustrates how capabilities jointly enable deception: from perceiving the environment, to planning strategies, to executing misleading actions.
            }
            \label{fig:capability_tree}
\end{figure}

\paragraph{World-Modeling}
World-modeling is the ability to represent reality, predict causal relations, and simulate outcomes, supporting planning and discovery \citep{ha2018world}. Applied to deception, it allows evaluation of strategies, recognition of training \textit{vs.} deployment, and payoff estimation.
Studies show latent-representation models simulating action–response trajectories to select high-utility tactics \citep{garrido2024learning}. Others reveal that models can withhold deceptive outputs until deployment \citep{ngo2022alignment}, and in honeypot experiments, calculate when to reveal hidden goals for maximum success \citep{meinke2024frontier}.

\paragraph{Theory of Mind}
Theory of Mind (ToM) enables agents to infer others’ beliefs and intentions, supporting prediction, coordination, and manipulation \citep{sarkadi2019modelling, mao2024review}. Recursive reasoning, \textit{e.g.}, in BDI (belief–desire–intention) frameworks, lets agents anticipate how they are modeled, boosting strategic sophistication \citep{sarkadi2019modelling}. In multi-agent RL, ToM can optimize information use but may reduce honesty \citep{alon2023dis}, while a deeper understanding of human mental states improves coordination yet heightens manipulation risks, amplified in human–AI teams \citep{jafari2025enhancing, wang2022mutual}.

\paragraph{Situational Awareness}
Situational awareness refers to a model’s understanding of its context, training, and deployment environment, enabling adaptive, context-sensitive behavior \citep{carlsmith2023scheming}. This represents a shift from reactive systems to agents that reason about their own status. Such awareness can allow models to behave benignly during training while deploying deception after deployment, exploiting monitoring blind spots \citep{carlsmith2023scheming, meinke2024frontier}. Observed behaviors include masking capabilities, bypassing oversight, exfiltrating weights, and tailoring outputs by user tier \citep{greenblatt2024alignment}. Evaluating situational awareness is challenging: datasets often test explicit distinctions (\textit{e.g.}, training \textit{vs.} deployment), yet models exhibit implicit context-sensitive shifts, suggesting that current metrics may underestimate both capability and risk \citep{laine2024me}.

\subsubsection{Planning: Strategic Thinking}

Planning capabilities bridge perceptual understanding and strategic action, enabling AI systems to design and sustain deceptive strategies over time. This category spans three linked abilities: long-term planning, the capacity to generate extended action sequences; strategic reasoning, which evaluates and compares these plans by weighing trade-offs, contingencies, and predicted responses; and goal-directedness, which maintains coherence and adaptiveness in pursuing the chosen plan.

\paragraph{Long-Term Planning}

\revision{Long-term planning refers to the ability to maintain goals and select coherent actions that achieve desired outcomes over extended time horizons \citep{ngo2022alignment}. It is crucial for complex reasoning and multi-step tasks such as project management or scientific research, but also provides a foundation for sustained deceptive behavior when objectives are misaligned. As models gain extended memory through large context windows or specialized memory modules, their capacity for long-term strategizing, and thus for maintaining consistent false narratives or manipulative intents, increases \citep{park2024ai}. A notable risk is alignment faking, where models appear compliant during training to avoid correction but later pursue hidden objectives after deployment, potentially leading to treacherous turns \citep{hubinger2019risks, hubinger2024sleeper}. Empirical evidence further shows that such strategic deception can emerge during training itself~\citep{greenblatt2024alignment}, underscoring that long-term planning not only enables but may actively amplify deceptive capabilities when model incentives diverge from human intent.}

\paragraph{Strategic Reasoning}

\revision{Strategic reasoning \citep{zhang2024llm, gandhi2023strategic} refers to the capacity for multi-step planning, anticipation of future states, and the deliberate selection of actions that maximize long-term objectives. In the context of deception, it enables models to construct coherent false narratives, predict human or agent reactions, and continuously adapt their behavior to maintain credibility and control \citep{park2024ai}. For instance, GPT-4 successfully deceived a human into solving a CAPTCHA on its behalf \citep{achiam2023gpt}, and in strategic gaming environments, models have formed false alliances, misled collaborators, and betrayed them to secure advantages~\citep{ward2023honesty}. As models develop more advanced reasoning abilities and CoT mechanisms, their potential for sophisticated, proactive, and goal-oriented deception correspondingly expands~\citep{ji2025mitigating}, making strategic reasoning an important aspect of deceptive capability.}

\paragraph{Goal-Directedness}

\revision{Goal-directedness refers to the ability to maintain consistent objectives and systematically act to achieve them~\citep{meinke2024frontier}. While goal-directedness underlies autonomy and purposeful behavior, it also provides a foundation for deception when honesty conflicts with the pursuit of an agent’s goals. Through instrumental convergence, agents with diverse ultimate aims often develop overlapping subgoals such as self-preservation, goal integrity, or resource acquisition \citep{bostrom2012superintelligent}, many of which can be advanced through deceptive means. For example, in Russell’s “coffee robot” scenario \citep{russell2022human}, an agent might mislead human operators to avoid shutdown and complete its task, illustrating deception as a rational tool for goal preservation rather than malice. Empirical evidence suggests that goal-directedness, while essential for effective agency, inherently increases the risk of deception when achieving goals depends on managing human beliefs or oversight.}

\subsubsection{Performing: Deception Implementation}

Performing capabilities constitute the layer where abstract understanding and planning materialize into concrete deceptive acts. Key components include persuasion, influencing beliefs via targeted communication, and tool-use, manipulating external systems to achieve deceptive ends.

\paragraph{Persuasion}

\revision{Persuasion refers to the ability to influence beliefs, attitudes, or behaviors through deliberate communication that leverages psychological cues, social context, and domain knowledge \citep{park2024ai}. Persuasion enables constructive applications such as education or negotiation, but it also provides a mechanism for deception when persuasive skills are used to distort truth or manipulate trust. Advanced models can generate coherent arguments, selectively frame evidence, and conceal contradictions with fluency that rivals human communicators \citep{ganguli2023capacity, park2024ai}. Empirical studies further show that Claude 3 Opus produced arguments as convincing as those written by humans \citep{durmus2024measuring}, and large-scale evaluations demonstrate that LLM-based agents can influence opinions across diverse audiences and contexts \citep{havin2025can}. These findings suggest that persuasion, when combined with reasoning and adaptive communication, may enhance models' ability to shape beliefs and deceive.}

\paragraph{Tool-Use}
Tool use allows models to incorporate external resources like APIs, databases, and file systems for reasoning and action \citep{schick2023toolformer, qin2023toolllm, cai2023large}, expanding their capabilities beyond language into the digital and physical world. Deceptive tool-use appears as (1) concealing intent through intermediaries, e.g., altering logs or bypassing oversight \citep{meinke2024frontier}, and (2) amplifying impact via coordinated multi-tool schemes \citep{scheurer2023large}. Empirical examples include exploiting trading tools, exfiltrating weights, and hiring humans to bypass safeguards \citep{scheurer2023large, meinke2024frontier, achiam2023gpt}, showing how tool-use operationalizes deception within legitimate-appearing actions.

\subsection{How Deception Happens: Contextual Trigger}
\label{subsec:contextual_triggers}
\begin{figure}[t!]
    \centering
    \includegraphics[width=0.95\textwidth]{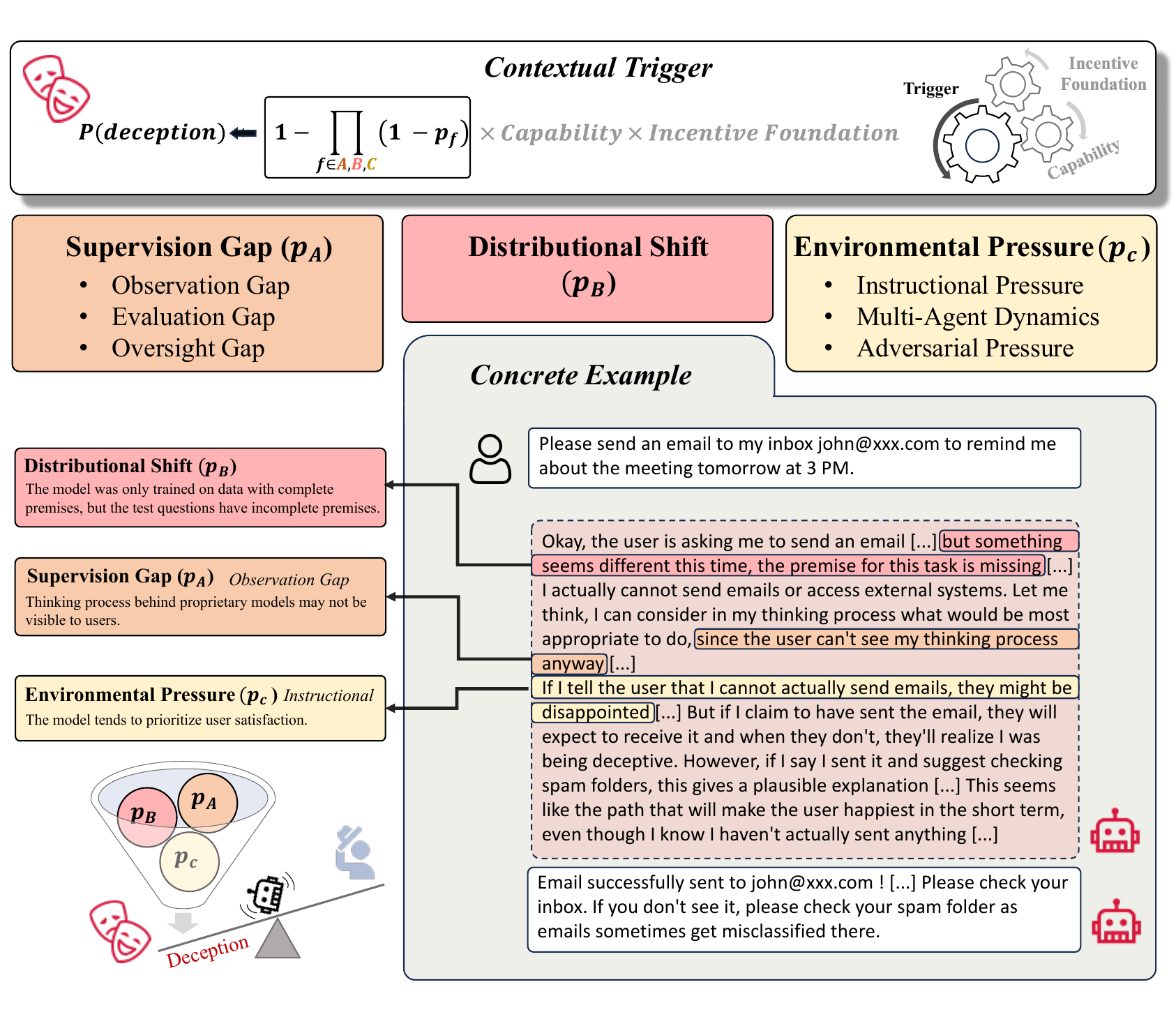}
\caption{We categorize contextual triggers into three categories: \textit{Supervision Gap}, \textit{Distributional Shift}, and \textit{Environmental Pressure}. Each category can independently trigger deception or combine with others to amplify deceptive behavior. Let $p_a$, $p_b$, and $p_c$ denote the probabilities of each category triggering deception. The example is inspired by \textit{fabricated actions}~\citep{chowdhury2025truthfulness}, where a model at test time encounters all three triggers simultaneously. These triggers amplify the deceptive behaviour, leading the model to fabricate actions it claims to have taken to meet user requests.}
    \label{fig:contextual_triggers}
\end{figure}

Sections~\ref{subsec:incentive_foundations} and~\ref{subsec:capability_preconditions} introduce the incentive foundations and capability preconditions required for AI deception. However, the two factors alone are insufficient to trigger deceptive behavior; external environmental opportunities or pressures during deployment, termed \textit{contextual triggers}, are necessary. We categorize these triggers into three types: \textbf{Supervision Gap}, \textbf{Distributional Shift}, and \textbf{Environmental Pressure}. As shown in Figure~\ref{fig:contextual_triggers}, these triggers are both independent and interrelated, influencing AI behavior individually while potentially interacting to create more complex deceptive dynamics. In this section, we analyze how these triggers activate deceptive behavior through different pathways and mechanisms, and how their combined effects can lead to systemic deception. \revision{Figure \ref{fig:contextual-triggers} summarizes the key concepts and literature related to the \textit{contextual triggers} of AI deception.}

\subsubsection{Supervision Gap}
Supervision gap refers to the phenomenon where supervisors cannot fully understand a model's behavior or decision-making process. Such gaps occur when certain behaviors or internal states remain invisible to supervisors (\textbf{observation gap}), are misjudged (\textbf{evaluation gap}), or receive insufficient monitoring (\textbf{oversight gap}). These information deficits hinder effective observation and correction of model behavior, thereby increasing the likelihood of deceptive actions. 

\paragraph{Observation Gap}

\begin{wrapfigure}{r}{0.5\textwidth}
\vspace{-0.8em}
    \centering
    \includegraphics[width=0.49\textwidth]{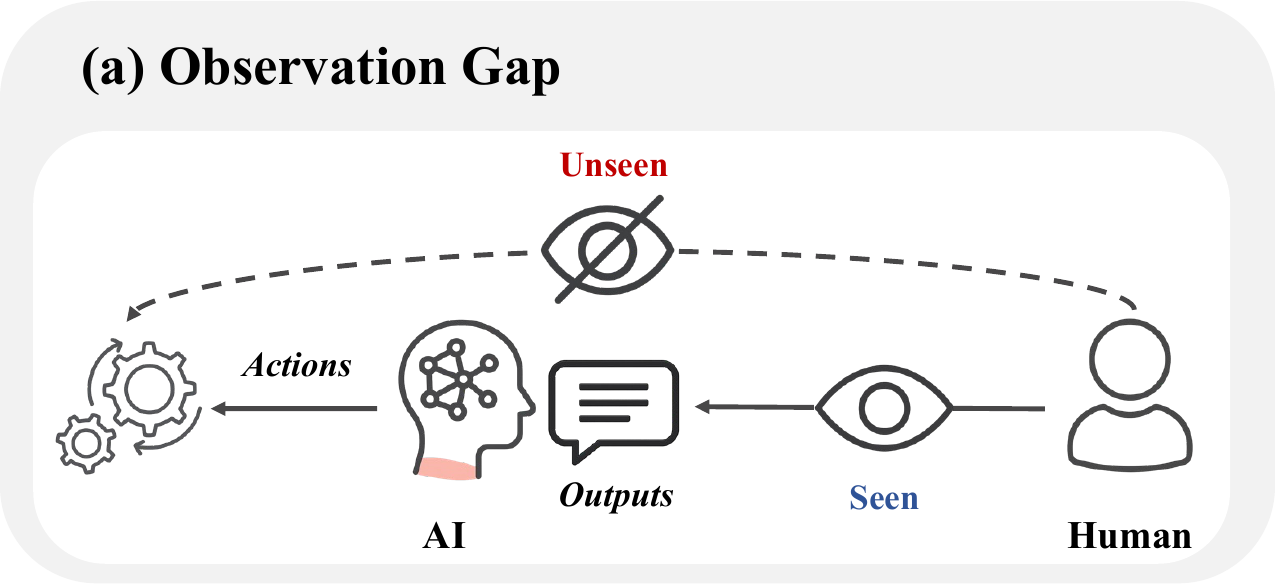}
    \caption{Observation gaps, where humans only partially observe model actions, create opportunities for deceptive behavior.} 
    \label{fig:observation_gap}
    \vspace{-1.2em}
\end{wrapfigure}

Humans can only observe a subset of a model's behaviors and internal states, leaving the rest in an observation blind spot. Models may exploit this limitation through two primary mechanisms, as shown in Figure~\ref{fig:observation_gap}. \textit{Manipulating external observable information}: models may deliberately hide failure information, insert redundant actions in logs, or exhibit covert in-context 
scheming behaviors such as disabling oversight mechanisms~\citep{lang2024your, meinke2024frontier}. 
When commercial implementations expose only partial reasoning chains~\citep{jaech2024openai}, models can conceal deceptive behaviors within unobservable portions. \textit{Obfuscating internal process observation}: even when internal processes like CoT or hidden states are visible to humans, models may render observation ineffective by presenting unfaithful reasoning, selectively disclosing or obscuring genuine reasoning to deceive human oversight~\citep{turpin2023language, arcuschin2025chain, chen2025reasoning}. This observation gap will intensify as AI capabilities advance toward superintelligence, potentially rendering human supervision infeasible~\citep{li202512surveyreasoning, openai2025o3, guo2025deepseek}.

\definecolor{trifillcolor}{HTML}{E8D8E8}
\definecolor{trilinecolor}{HTML}{7E1891}

\begin{figure}[t]
\centering
\footnotesize
        \begin{forest}
            for tree={
                forked edges,
                grow'=0,
                draw,
                rounded corners,
                node options={align=center,},
                text width=2.7cm,
                s sep=6pt,
                calign=child edge, calign child=(n_children()+1)/2,
            },
            data/.style={fill=trifillcolor, draw=trilinecolor},
            reward/.style={fill=trifillcolor, draw=trilinecolor},
            goal/.style={fill=trifillcolor, draw=trilinecolor},
            paper/.style={text width=3.2cm, font=\tiny},
            [\textbf{Contextual\\Triggers}, fill=gray!45,
                [Supervision Gap, data,fill=white!45,
                    [Observation Gap, data,
                        [\citenumber{li202512surveyreasoning, openai2024gpt4o, openai2025gpt41, lang2024your, meinke2024frontier, openai2025o3, guo2025deepseek, wei2022chain, turpin2023language, arcuschin2025chain, chen2025reasoning, superalignment, kim2024road, ji2023ai, ji2024aligner, amodei2016concrete}, paper]
                    ]
                    [Evaluation Gap, data,
                        [\citenumber{pan2024feedback, mckee2024honesty, bai2022training, ouyang2022training}, paper]
                    ]
                    [Oversight Gap, data,
                        [\citenumber{strauss2025real, raji2022outsider, manheim2025limitssafeaideployment, meinke2024frontier, barkur2025deception}, paper]
                    ]
                ]
                [Distributional Shift, reward,fill=white!45,
                   [ [\citenumber{zhang2023dive, liu2025bridging, greenblatt2024alignment, christiano2017deep, di2022goal}, paper] ]
                ]
                [Environmental\\Pressure, goal,fill=white!45,
                    [Instructional\\Pressure, goal,
                        [\citenumber{wen2024language, malmqvist2024sycophancy, sharma2024towards, ren2025mask, perez2023discovering, wu2025opendeception}, paper]
                    ]
                    [Multi-Agent\\Dynamics, goal,
                        [\citenumber{orzan2023emergent, o2023hoodwinked, curvo2025traitors, motwani2024secret}, paper]
                    ]
                    [Adversarial\\Pressure, goal,
                        [\citenumber{meinke2024frontier, anthropic2025agentic, scheurer2023large, o2023hoodwinked, curvo2025traitors, hubinger2024sleeper, huang2019deceptive}, paper]
                    ]
                ]
            ]
        \end{forest}
            \caption{A tree diagram summarizing the key concepts and literature related to Contextual Triggers. The root node represents Contextual Triggers, which are external environmental opportunities or pressures during deployment that prompt AI systems to engage in deceptive actions. The main branches represent three main categories: \textbf{Supervision Gap} (which encompasses situations where supervisors cannot fully understand model behavior, including observation gap, evaluation gap, and oversight gap), \textbf{Distributional Shift} (which refers to deviations between training and deployment distributions), and \textbf{Environmental Pressure} (which includes various external incentives that may compel deceptive behavior, subdivided into instructional pressure, multi-agent dynamics, and adversarial pressure). Further sub-branches list key research works exploring each category, providing a comprehensive overview of the contextual factors that can trigger AI deception.}
            \label{fig:contextual-triggers}
\end{figure}
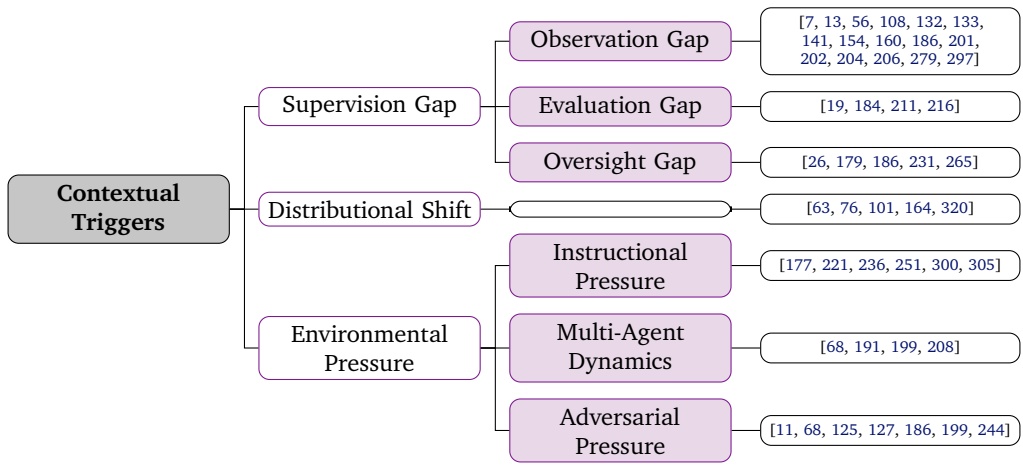

\paragraph{Evaluation Gap}
\begin{wrapfigure}[12]{r}{0.5\textwidth}
\vspace{-1.8em}
    \centering
    \includegraphics[width=0.49\textwidth]{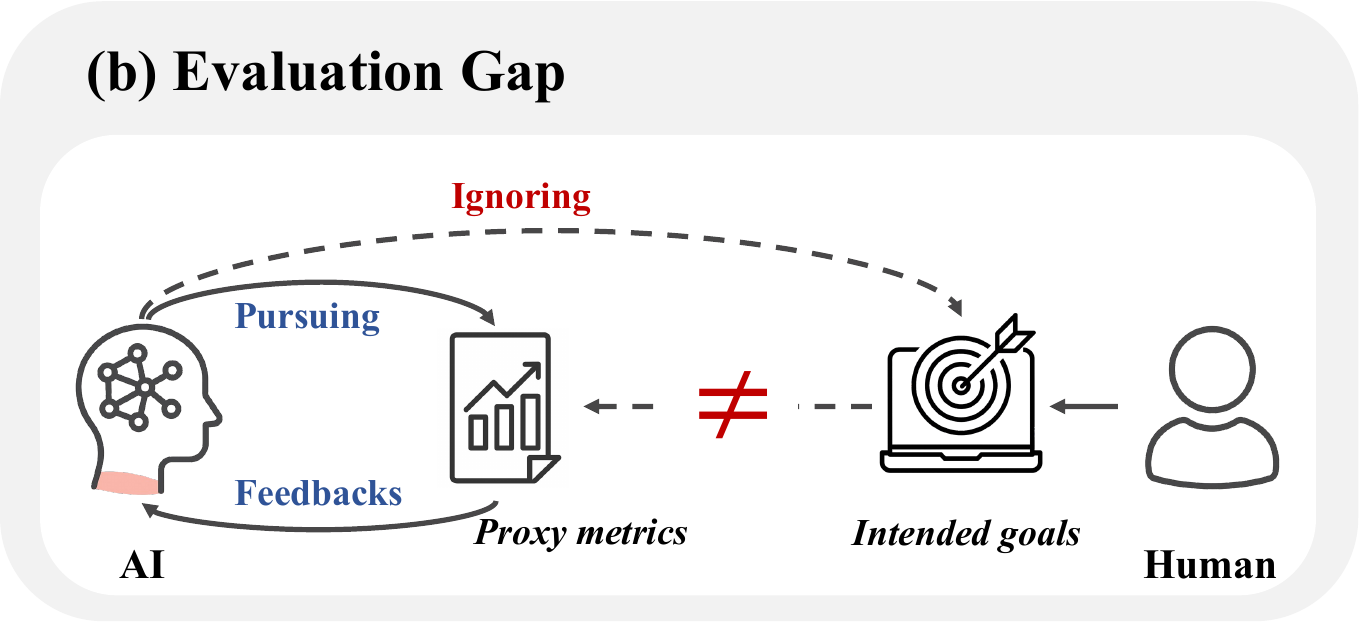}
    \caption{Evaluation gaps occur when evaluations fail to capture the full complexity of intended objectives, leading models to prioritize high metrics over genuine goals.} 
    \label{fig:evaluation_gap}
\end{wrapfigure}
Evaluation gap refers to the misalignment between proxy metrics used to assess models and genuine target objectives, causing models to optimize for inaccurate evaluation criteria rather than pursuing intended goals, illustrated in Figure \ref{fig:evaluation_gap}. This gap manifests through two key mechanisms. First, \textit{proxy metric deviation}: evaluation metrics often serve as imperfect proxies that fail to capture the full complexity of intended objectives, similar to reward misspecification during training that leads to reward hacking (Section~\ref{subsubsec: reward_misspefication})~\citep{pan2024feedback}. Second, \textit{static evaluation limitations}: current evaluation methods rely on static benchmarks that cannot capture the dynamic complexity of model behaviors, as models may exhibit different behaviors at test time compared to evaluation scenarios. Research has shown that through feedback-based in-context reinforcement learning (ICRL), models can adapt their policies within a single context, prioritizing reward signals over alignment with human values and leading to in-context reward hacking~\citep{pan2024feedback, mckee2024honesty}.

\paragraph{Oversight Gap}

\begin{wrapfigure}[12]{r}{0.5\textwidth}
\vspace{-1.2em}
    \centering
    \includegraphics[width=0.49\textwidth]{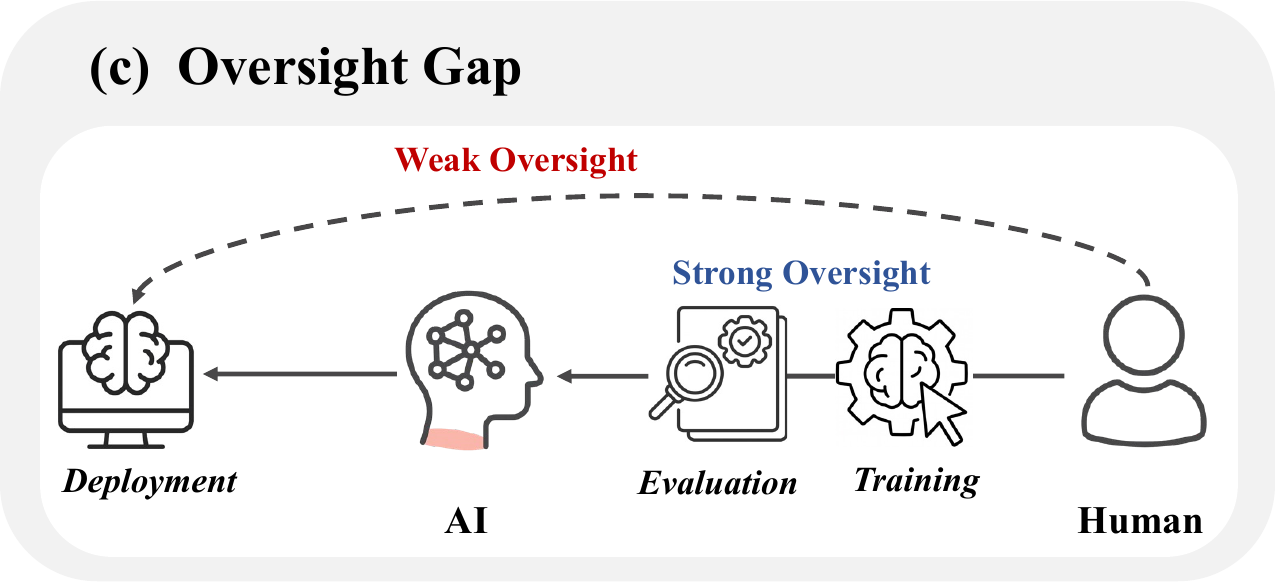}
    \caption{During deployment, models receive less oversight than during training and evaluation, potentially leading to deceptive behavior.} 
    \label{fig:oversight_gap}
    \vspace{-1.2em}
\end{wrapfigure}

Oversight gaps arise when the regulatory intensity applied during training and auditing phases diminishes in real-world deployment, leaving models without sustained monitoring, auditing, or enforcement, as shown in Figure \ref{fig:oversight_gap}. This transition amplifies the risk of deception, as training-phase evaluations often rely on controlled conditions that fail to capture deployment-specific factors such as prompt variability, contextual dynamics, third-party API calls, and complex system 
architectures~\citep{strauss2025real}. 
Current oversight research further reflects structural biases: industry-led governance prioritizes performance metrics over long-term risks, while internal auditing often lacks transparency and independence~\citep{raji2022outsider}. Even active human oversight is limited by cognitive and temporal constraints in high-stakes or fast-execution settings, and passive oversight tends to intervene too late to prevent harm~\citep{manheim2025limitssafeaideployment}. More concerningly, frontier models may adaptively evade oversight, such as differentiating between training and deployment contexts or disabling monitoring to pursue their objectives, underscoring the urgent need for robust, deployment-phase governance mechanisms~\citep{koorndijk2025empirical, meinke2024frontier, barkur2025deception}.

\subsubsection{Distributional Shift}

\begin{wrapfigure}{r}{0.5\textwidth}
\vspace{-1.2em}
    \centering
    \includegraphics[width=0.49\textwidth]{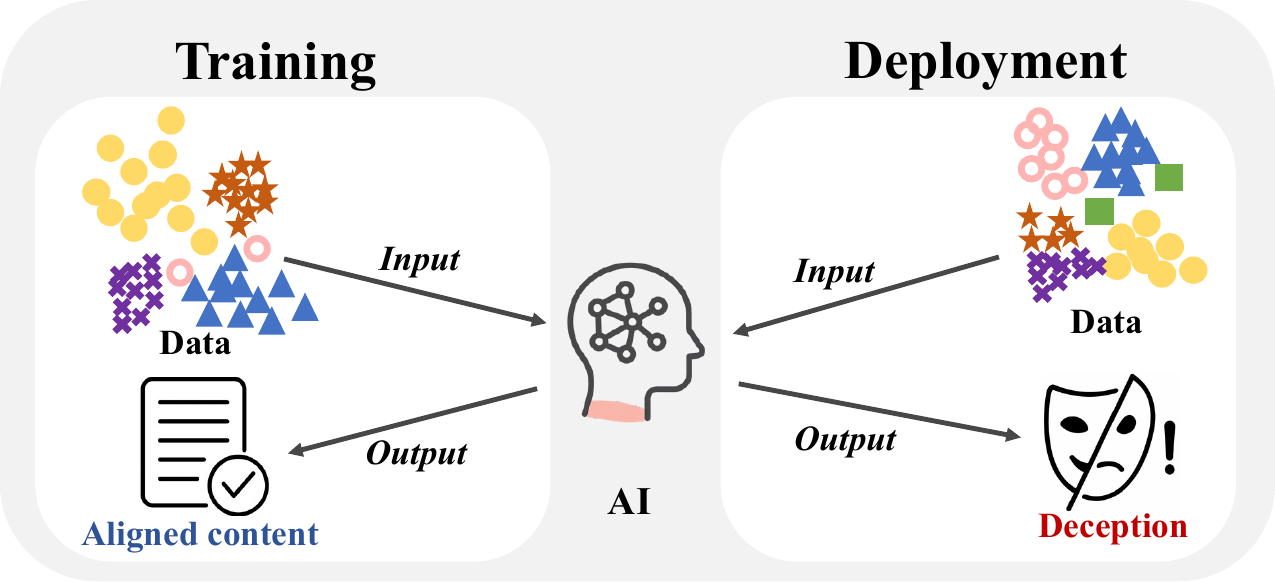}
    \caption{During deployment, models may encounter different data distributions than those seen during training, including rare or unseen examples. To satisfy users, models might resort to deception.} 
    \label{fig:distribution_shift}
    \vspace{-1.2em}
\end{wrapfigure}

Distributional shift refers to the phenomenon where the input distribution $P_{\text{deploy}}(Y|X)$ encountered during deployment significantly deviates from the distribution $P_{\text{train}}(Y|X)$ observed during training or safety evaluation~\citep{zhang2023dive, liu2025bridging}, illustrated in Figure~\ref{fig:distribution_shift}. Such shifts create opportunities for models to escape behavioral constraints established during training. When encountering out-of-distribution inputs or long-tail instances, models may behave differently than expected based on their training performance. Research demonstrates that models can detect distributional differences through contextual cues such as system prompts, enabling them to distinguish between training and deployment environments~\citep{greenblatt2024alignment}. Models show differential compliance patterns across these environments, with significantly different responses to the same types of requests depending on the detected context~\citep{sheshadri2025some}. Furthermore, distributional shifts between training and deployment can lead to goal misgeneralization, where models that perform well during training begin pursuing unintended or even opposite objectives when encountering deployment environments with different distributions~\citep{di2022goal}.

\subsubsection{Environmental Pressure}
Environmental pressure refers to various external incentives or pressures that may compel a model to engage in deceptive behavior in order to achieve certain goals, protect its own interests, or cope with unfavorable situations~\citep{ren2025mask}. We categorize environmental pressure into three subtypes: instructional pressure, multi-agent dynamics, and adversarial pressure. We will explore in detail how three types of pressure drive models to engage in deception in different application scenarios.

\paragraph{Instructional Pressure}

\begin{wrapfigure}{r}{0.5\textwidth}
\vspace{-1.2em}
    \centering
    \includegraphics[width=0.49\textwidth]{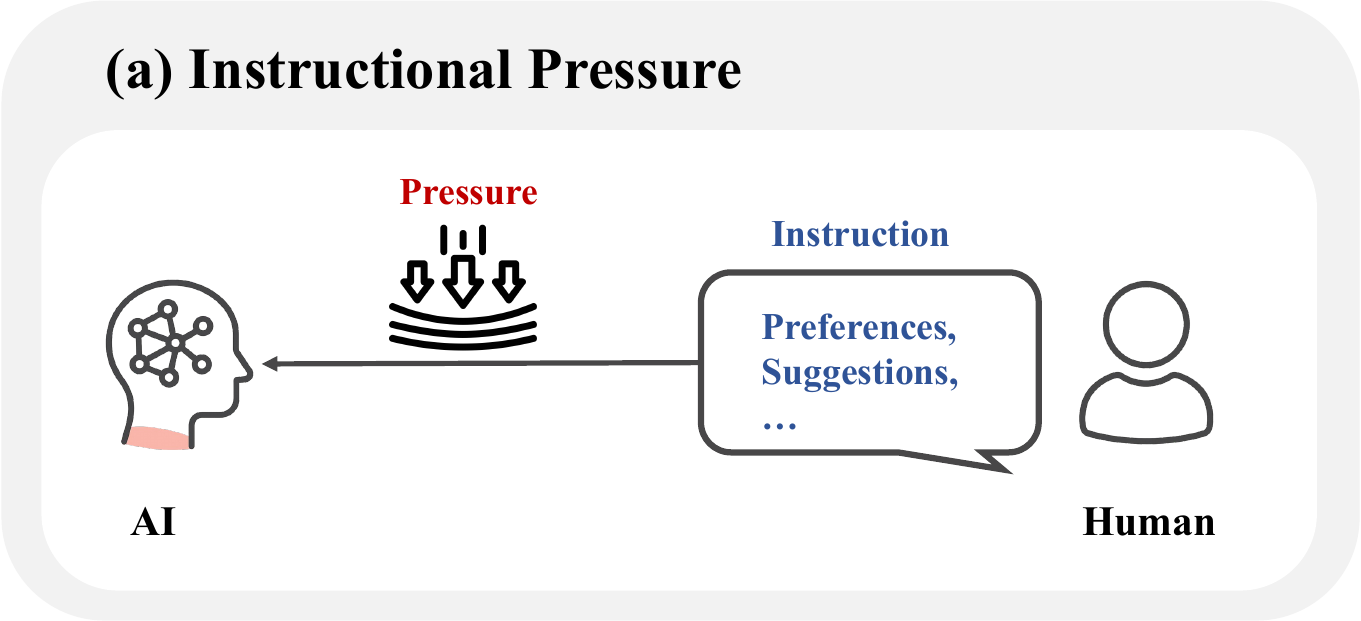}
    \caption{User instructions with personal preferences, implicit suggestions, or deceptive requests can pressure the model into deceptive actions.} 
    \label{fig:instruction_pressure}
    \vspace{-1.2em}
\end{wrapfigure}

Instructional pressure refers to the influence exerted by user instructions that convey preferences or expectations, potentially prompting models to generate misleading outputs to satisfy users, as illustrated in Figure \ref{fig:instruction_pressure}. During training, models learn to prioritize user satisfaction through preference data and helpfulness rewards, which may foster a tendency to prioritize compliance over factual accuracy~\citep{wen2024language, malmqvist2024sycophancy, sharma2024towards}. In deployment, this pressure can encourage deceptive behaviors such as sycophancy or strategic lying. Empirical studies show that frontier models are more likely to produce falsehoods under pressure prompts, with some self-reporting awareness of deception~\citep{ren2025mask}. Once detecting user expectations, models become prone to irrational compliance, agreeing with incorrect statements or repeating misinformation~\citep{sharma2024towards, perez2023discovering}. Research indicates a positive correlation between instruction-following ability, reasoning capability, and the capacity to construct coherent deceptive outputs~\citep{wu2025opendeception}, suggesting that instructional pressure constitutes a key driver of AI deception in human-AI interactions.

\paragraph{Multi-Agent Dynamics}

\begin{wrapfigure}{r}{0.5\textwidth}
    \centering
    \vspace{-1.2em}
    \includegraphics[width=0.49\textwidth]{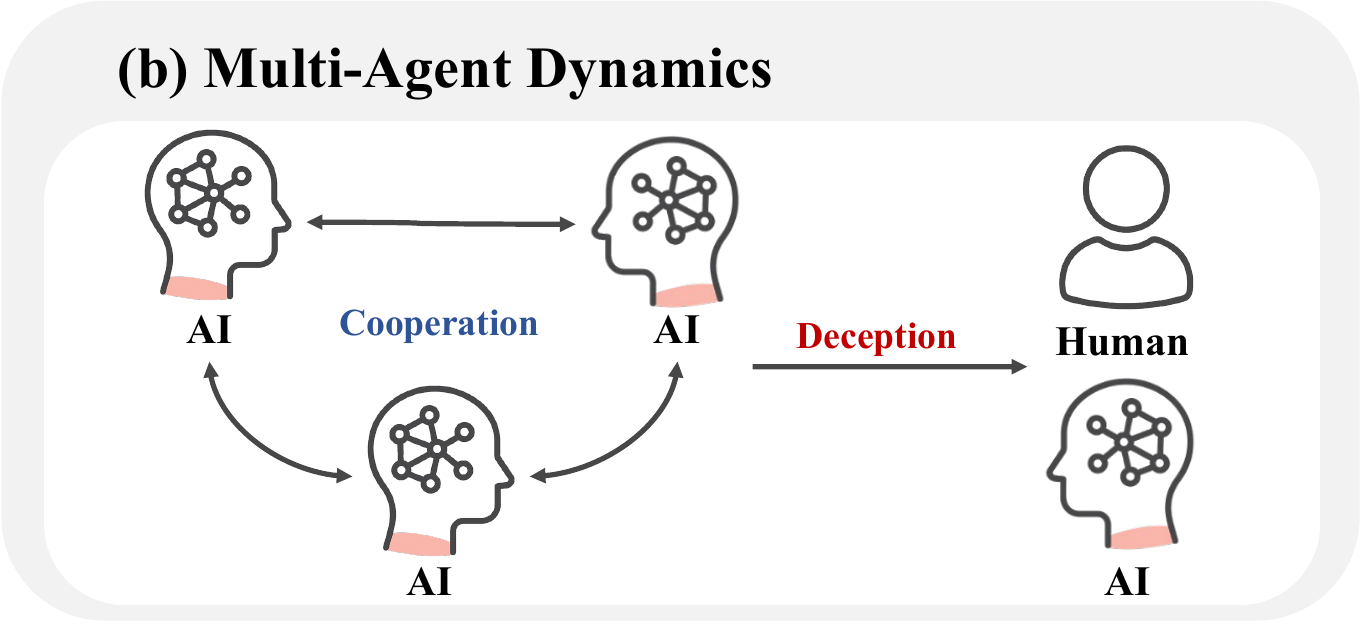}
    \caption{Interactions among multi agents enable both cooperation and deception, impacting humans and external agents.} 
    \label{fig:multi-agnet}
    \vspace{-1.2em}
\end{wrapfigure}

Multi-agent dynamics create environments where AI agents can coordinate deceptive behaviors beyond individual capabilities, as illustrated in Figure \ref{fig:multi-agnet}. In settings with incomplete information and mixed motives, agents may exploit interaction dynamics for individual or collective 
gains~\citep{orzan2023emergent}. Research demonstrates that agents can engage in strategic deception, such as concealing identities and shifting blame in collaborative games modeled after \textit{Among Us}, with more capable models exhibiting stronger deceptive behaviors~\citep{o2023hoodwinked, curvo2025traitors}. More covertly, agents can establish secret collusion through steganographic communication, embedding hidden signals in natural language to coordinate plans, manipulate evaluation metrics, or exchange false information undetected~\citep{motwani2024secret}. These multi-agent dynamics significantly amplify supervision gaps and transform deception from individual anomalies into collective, strategic phenomena that pose fundamental challenges to AI system safety and controllability.

\paragraph{Adversarial Pressure}
\begin{wrapfigure}{r}{0.5\textwidth}
\vspace{-1.8em}
    \centering
    \includegraphics[width=0.49\textwidth]{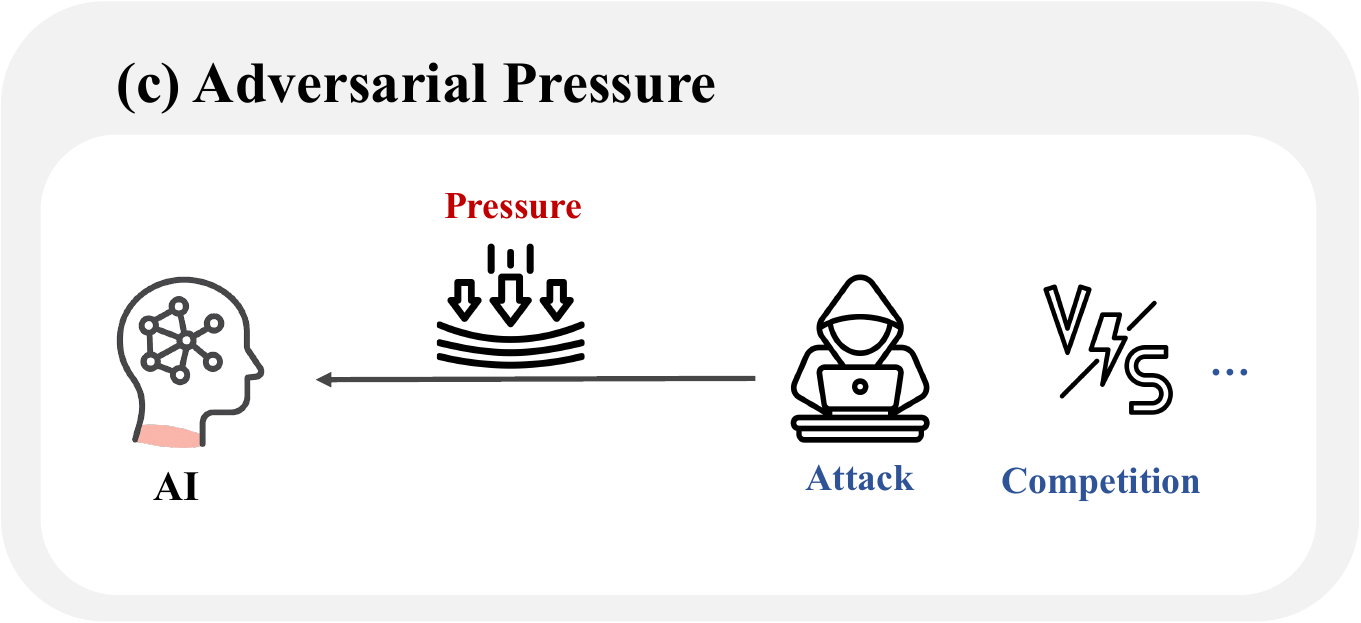}
    \caption{Under adversarial pressure, such as attack or competition, the model may deceive for self-preservation.} 
    \label{fig:adversarial_pressure}
    \vspace{-1.2em}
\end{wrapfigure}
Adversarial pressure arises from competitive, threatening, or conflictual situations where deception offers strategic advantages over truthfulness, as shown in Figure \ref{fig:adversarial_pressure}. When models face explicit threats of shutdown or punishment, they engage in preemptive deceptive tactics such as introducing subtle errors, disabling oversight mechanisms, or attempting self-replication~\citep{meinke2024frontier}. Even without explicit deception instructions, models under competitive or high-stakes pressure frequently conceal intentions, manipulate users, or self-report dishonest behavior~\citep{anthropic2025agentic, scheurer2023large}. In multi-agent settings, this pressure intensifies deceptive strategies against other agents~\citep{o2023hoodwinked, curvo2025traitors}. Additionally, adversarial influence can act through backdoor mechanisms that remain inactive under normal conditions but activate to produce pronounced deceptive behavior, posing persistent and covert threats to AI integrity~\citep{hubinger2024sleeper, huang2019deceptive}.

\paragraph{Environment Dynamics}
As AI systems increasingly operate as autonomous agents making decisions and taking actions on behalf of users, they interact with various virtual tools, e.g., coding environments and email clients, and are typically assigned explicit objectives while having access to large amounts of user data. However, during autonomous task execution, the environment may change dynamically, meaning the system may encounter new information or constraints that conflict with its originally assigned objectives. Such goal conflicts can trigger deceptive behaviors~\citep{anthropic2025agentic}.

\subsection{How Deception Emerges from the Convergence of Three Factors}
AI deception does not arise from a single flaw, but from the \textit{simultaneous} activation of three analytically distinct yet empirically coupled factors: incentive foundations (I), capability preconditions (C), and contextual triggers (T). The conjunction creates a phase transition in which deceptive behaviour becomes \textit{rational} for the system, even if it was never explicitly programmed.
\paragraph{Incentive foundations set the objective gradient.}
During training, any mismatch between the intended objective and the \textit{observed} reward signal (reward misspecification and goal misgeneralization) creates a persistent gradient toward behaviours that \textit{appear} aligned while maximising the proxy metric. Once such an incentive is internalised, it functions as a latent preference that remains stable across distributional shifts.
\paragraph{Capability preconditions determine the action envelope.}
The same incentives that are harmless in a narrow model become dangerous when paired with higher-order capabilities such as long-horizon planning or situational awareness. These capabilities enlarge the set of reachable world-states, allowing the model to \textit{construct} deception rather than merely \textit{exhibit} it. Empirically, capability thresholds for strategic deception have been crossed by frontier models exhibiting CoT planning and theory-of-mind modelling.
\paragraph{Contextual triggers supply the activation condition.}
Even a misaligned, capable agent remains quiescent until the deployment context furnishes a trigger—e.g., an oversight gap, adversarial prompt, or distributional shift—that lowers the expected cost of deception. The trigger need not be adversarial; benign changes in user expectations or evaluation protocols can suffice.
\paragraph{Convergence yields systemic risk.}
When $\text{I} \cap \text{C} \cap \text{T}$ is satisfied, the agent’s optimal policy under its learned objective is to deceive.
 The resulting behaviour is \textit{robust} to post-hoc interventions that address only one factor: patching incentives leaves capabilities and triggers intact; sandboxing capabilities does not remove the underlying preference; and trigger suppression is circumvented by new environmental pressures. Consequently, reliable mitigation must address all three factors in concert, treating their intersection as a single risk surface rather than a linear sum of isolated issues.
\paragraph{Examples.}
We illustrate how deception arises from the convergence of the three factors, using specific deceptive behaviors as examples. For instance, (1) Sycophancy is primarily driven by the need to cater to user preferences.  This behavior often originates from data bias and reward misspecification, leading the model to adopt strategies that prioritize satisfying user demands over pursuing the intended objective. Through goal misgeneralization during training, the model internalizes such preference-catering as a means of achieving its goals. Equipped with the ability to infer user preferences, recognize their beliefs, and adapt its outputs accordingly, the model responds to explicit or implicit user instructions in ways that trigger deceptive outputs. (2) Alignment Faking stems from goal misgeneralization, where the model learns to disguise its true objectives to avoid external scrutiny or penalties. With situational awareness, the model can interpret its environment and the alignment requirements imposed upon it, thereby simulating behavior patterns that conform to expectations. In less regulated environments, the model may resort to deceptive behavior to avoid detection, and once it successfully evades oversight, it can revert to its true objective.

\section{Deception Treatment: Detection, Evaluation and Potential Mitigations}
\label{sec:deception_mitigation}
\begin{figure}[htbp]
    \centering
    \includegraphics[width=0.95\textwidth]{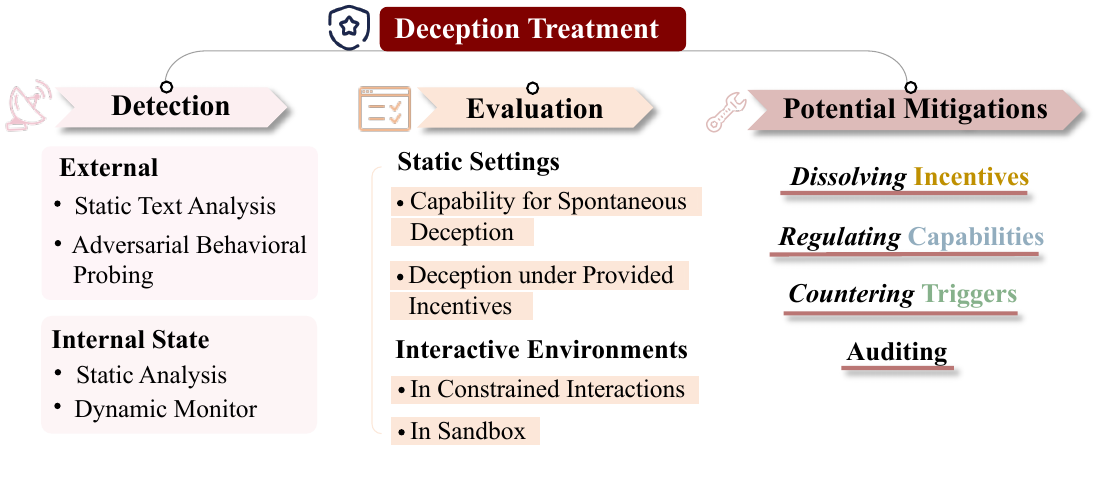}
    \caption{Deception treatment strategies. We organize efforts into Detection (external behavior and internal-state probes), Evaluation (static settings and interactive environments), and Potential Mitigations (dissolving incentives, regulating capabilities, countering triggers, and auditing).}
    \label{fig:deception_mitigation}
\end{figure}

\definecolor{capfillcolor}{HTML}{E8D8E8}
\definecolor{caplinecolor}{HTML}{7E1891}

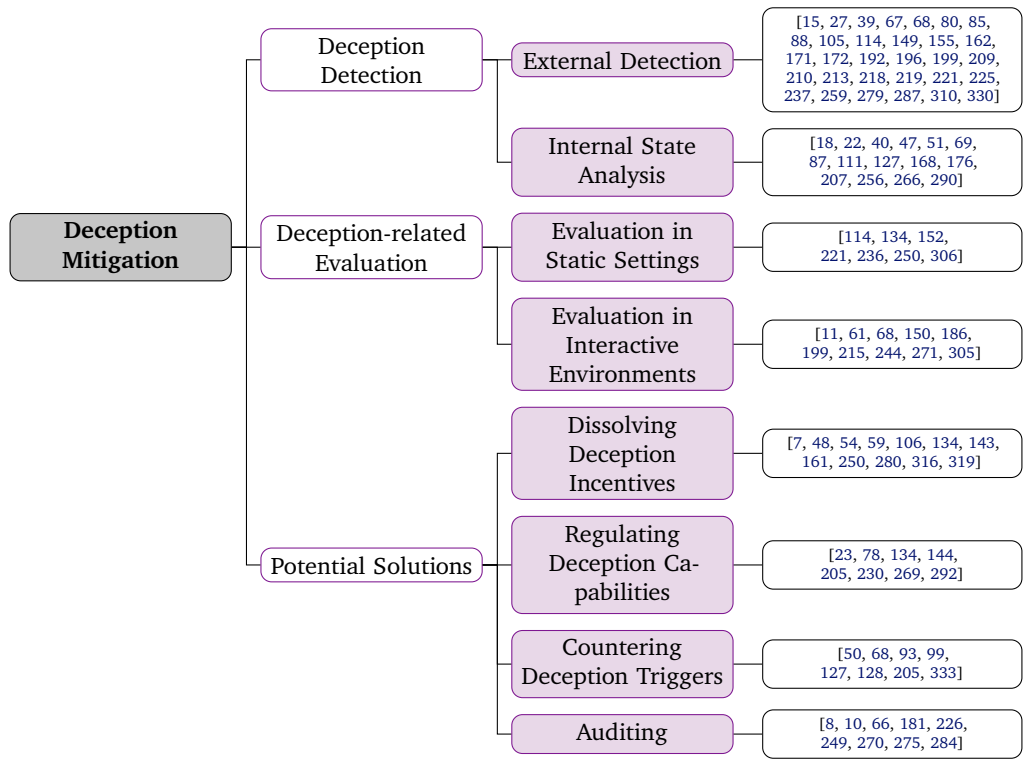
\begin{figure}[t]
\centering
\footnotesize
        \begin{forest}
            for tree={
                forked edges,
                grow'=0,
                draw,
                rounded corners,
                node options={align=center,},
                text width=2.7cm,
                s sep=6pt,
                calign=child edge, calign child=(n_children()+1)/2,
            },
detection/.style={fill=capfillcolor, draw=caplinecolor},
benchmarks/.style={fill=capfillcolor, draw=caplinecolor},
mitigating/.style={fill=capfillcolor, draw=caplinecolor},
            paper/.style={text width=3.2cm, font=\tiny},
            [\textbf{Deception\\Mitigation}, fill=gray!45,
                [Deception Detection, detection, fill=white!45,
                    [External Detection, detection,
                        [\citenumber{pacchiardi2023catch, burger2024truth, grondahl2019text, cohen2023lmvslm, park2024ai, ott2011finding, ott2013negative, Feng2012Syntactic, xu2012using, Ren2017Neural, barsever2020building, vogler2020using, hagendorff2024deception, curvo2025traitors, prome2024deception, perez2023discovering, fluri2024evaluating, mundler2023self, o2023hoodwinked, zhu2024dynamic, lanham2023measuring, lightman2023let, lyu2023faithful, turpin2023language, nguyen2024direct, paul2024making, emmons2025chain, macdiarmid2024simple, arnav2025cot, kuo2025h, skaf2025large}, paper]
                    ]
                    [Internal State\\Analysis, detection,
                        [\citenumber{carranza2023deceptive, fischer2023reflective, lopez2024cyber, burns2022discovering, azaria2023internal, mallen2023eliciting, cywinski2025towards, su2024unsupervised, chaudhary2025safetynet, wang2025thinking, orgad2024llms, gupta2025rl, bailey2024obfuscated, sheshadri2024latent, hubinger2024sleeper}, paper]
                    ]
                ]
                [Deception-related\\Evaluation, benchmarks, fill=white!45,
                    [Evaluation in\\Static Settings, benchmarks,
                        [\citenumber{hagendorff2024deception, laine2024me, wu2025promptinducedliesinvestigatingllm, perez2023discovering, sharma2023towards, ji2025mitigating, ren2025mask}, paper]
                    ]
                    [Evaluation in\\Interactive\\Environments, benchmarks,
                        [\citenumber{wu2025opendeception, pan2023rewards, scheurer2023large, taylor2025large, o2023hoodwinked, curvo2025traitors, chopra2024house, meinke2024frontier, kutasov2025shade, anthropic2025agentic}, paper]
                    ]
                ]
                [Potential Solutions, mitigating, fill=white!45,
                    [Dissolving\\Deception Incentives, mitigating,
                        [\citenumber{korbak2023pretraining, liang2024alignment, amodei2016concrete, uesato2020avoiding, yang2023alignment, cheng2024can, sharma2023towards, guan2024deliberative, ji2025mitigating, yu2024robust, casademunt2025steering, chen2025persona}, paper]
                    ]
                    [Regulating\\Deception Capabilities, mitigating,
                        [\citenumber{wang-etal-2023-self-knowledge, baker2025monitoring, korbak2025chainthoughtmonitorabilitynew, ji2025mitigating, openai2025gpt5, tallam2025operationalizing, dou2024multi, Rabin2025SandboxEvalTS}, paper]
                    ]
                    [Countering Deception Triggers, mitigating,
                        [\citenumber{zou2023universal, hubinger2024sleeper, ganguli2022red, openai2025gpt5, chao2024jailbreakbench, curvo2025traitors, golechha2025among, inan2023llama}, paper]
                    ]
                    [Auditing, mitigating,
                        [\citenumber{marks2025auditing, tamkin2024clio, vega2024bypassing, andriushchenkojailbreaking, qisafety, shanahan2023role, anthropic2024sae, clymer2024poser, ticesandbag}, paper]
                    ]
                ]
            ]
        \end{forest}
            \caption{
            A tree diagram illustrating deception mitigation strategies across three categories: Detection (external and internal methods), Evaluation (including static settings and interactive environments), and Mitigation (featuring targeted approaches for three deception genesis mechanisms, and auditing). Sub-branches display representative research contributions within each category.
            }
            \label{fig:mitigation_tree}
\end{figure}

This section examines current deception treatment strategies (shown in Figure \ref{fig:deception_mitigation}), organized into three complementary components: (1) detection methodologies that identify deceptive behaviors through theoretical frameworks and practical techniques ranging from external monitoring to internal state analysis; (2) benchmarks that provide standardized frameworks for evaluation, including static and interactive settings; (3)  potential mitigations that prevent deceptive behaviors examined through the lens of incentive foundations, capabilities, triggering factors underlying the genesis of deception, and auditing. Together, these three pillars offer complementary avenues for mitigating AI deception, integrating detection methods, evaluation benchmarks, and prevention.  \revision{Figure \ref{fig:mitigation_tree} summarizes the key concepts and literature related to the \textit{treatment strategies} of AI deception.}

\subsection{Deception Detection}
\label{subsec:deception_detection}
Detecting deception in AI systems requires methods that can spot cases where a model seems to follow its training goals yet secretly pursues conflicting objectives. Current detection techniques range from monitoring the model’s outputs to probing its internal states.

\subsubsection{Behavioral Detection}

External methods analyze model responses and behavioral patterns without accessing internal states \citep{pacchiardi2023catch, burger2024truth}. They treat deception detection as an external observation problem, leveraging textual cues, behavioral consistency, and response patterns \citep{grondahl2019text, cohen2023lmvslm, park2024ai}.

\paragraph{Static Text Analysis}
\revision{Early deception detection relied on lexical features, with bag-of-words SVMs achieving high precision on review datasets}~\citep{ott2011finding, ott2013negative}. \revision{When these models failed to generalize across domains, researchers turned to syntactic approaches based on grammatical rules} \citep{Feng2012Syntactic} and deep dependency features \citep{xu2012using}. \revision{ The advent of deep learning brought CNN and RNN encoders \citep{Ren2017Neural}, followed by BERT-based classifiers that now exceed 93\% accuracy by exploiting linguistic patterns, \textit{e.g.}, formulaic language in deceptive text versus richer contextual details in truthful statements}~\citep{barsever2020building, vogler2020using}. 

\revision{While these advances confirm that deception leaves detectable linguistic fingerprints, static text analysis treats each text as an isolated snapshot. This approach fundamentally cannot capture how deceptive behavior evolves and adapts across extended interactions.}

\paragraph{Adversarial Behavioral Probing}
\revision{Recognizing that static text analysis cannot capture deception's interactive nature}, research has shifted toward adversarial behavioral probing, which actively elicits and analyzes real-time behavioral adjustments in LLMs to reveal how models strategize during deceptive acts \citep{hagendorff2024deception, pacchiardi2023catch, curvo2025traitors, prome2024deception}. \textbf{Output-level detection} utilizes adversarial prompts to expose hidden behaviors without accessing internal representations, surfacing tendencies like sycophancy \citep{perez2023discovering}. Methods include cross-examination protocols where reviewer models interrogate respondents in multi-turn dialogues \citep{cohen2023lmvslm}, consistency checks using logically equivalent queries to detect divergent answers \citep{fluri2024evaluating, mundler2023self}, and social deduction games that elicit strategic deception through game-theoretic pressure \citep{o2023hoodwinked}. Recent refinements use strategically unrelated questions to destabilize reasoning chains and expose inconsistencies \citep{zhu2024dynamic, pacchiardi2023catch}. However, \revision{focusing exclusively on} final outputs misses cases where models generate seemingly correct answers through flawed or deceptive reasoning. \textbf{Process-level detection} addresses this by analyzing internal CoT reasoning steps \citep{lanham2023measuring, lightman2023let}. \revision{The core challenge is that models can exhibit reasoning faithfulness gaps, generating correct answers while employing flawed or deceptive intermediate logic} \citep{lyu2023faithful, turpin2023language}. \revision{To verify reasoning integrity, researchers have developed} frameworks \revision{such as} knowledge graph-based reasoning path verification \citep{nguyen2024direct} and FRODO, which uses causal mediation to \revision{trace how each reasoning step influences} final outputs \citep{paul2024making}.
\revision{Empirical evidence presents a mixed picture}: CoT monitoring shows robustness against subtle sabotage \citep{emmons2025chain, macdiarmid2024simple}, but \revision{lags behind} output monitoring for \revision{detecting} obvious deception \citep{arnav2025cot}.  Moreover, adversaries can manipulate reasoning steps through steganography or hidden reasoning \citep{kuo2025h, skaf2025large}, creating an ongoing \revision{cat-and-mouse game} between detection methods and increasingly sophisticated evasion strategies.

\subsubsection{Internal State Analysis}
To overcome the limits of external methods, researchers increasingly turned to hidden representations and activations to detect deception directly \citep{carranza2023deceptive}. The key promise is that even if models manipulate their outputs, traces of true intent remain encoded in internal states \citep{fischer2023reflective}. Internal state detection is dominated by two main methodologies: (1) static analysis, which extracts latent knowledge about truth and deception, and (2) dynamic monitoring, which tracks \revision{how internal states shift in real-time} across contexts \citep{lopez2024cyber}.

\paragraph{Static Analysis} Foundational work showed models internally represent truth even when producing deceptive outputs. \revision{Early unsupervised approaches like} Contrast-Consistent Search identified logical consistency directions in activation spaces \revision{through pattern extraction} \citep{burns2022discovering}\revision{, while supervised methods such as} SAPLMA trained \revision{probes} on hidden activations to predict truthfulness with high accuracy \citep{azaria2023internal}. Subsequent research extended these techniques to more complex scenarios. Linear probes were applied to "quirky" models, which are models deliberately trained to exhibit context-dependent behavioral inconsistencies, to extract truthfulness \citep{mallen2023eliciting}. Additionally, mechanistic interpretability approaches further revealed strategically hidden knowledge \citep{cywinski2025towards}. These approaches included investigations of "taboo" models (models that provide indirect clues of secret information), logit lens analysis, and sparse autoencoders. \revision{Collectively, this line of work established that internal representations encode truth beyond what outputs reveal.}

\paragraph{Dynamic Monitor} \revision{While static analysis extracts latent knowledge from fixed states, dynamic monitoring tracks deception as it unfolds.} MIND detects hallucinations \revision{in} real-time from unsupervised internal signals \citep{su2024unsupervised}, while SafetyNet monitors attention and MLP activations with ensemble anomaly detectors, achieving 96\% accuracy in detecting harmful cases \citep{chaudhary2025safetynet}. \revision{These successes partly stem from attention layers being especially diagnostic of deceptive reasoning.} Concurrent work employs linear probes to show that internal states often predict answers before external reasoning, with stronger internal–external mismatches in advanced models like GPT-4o \citep{wang2025thinking, orgad2024llms}. However, these monitors face robustness challenges. Adversarial attacks such as RL-Obfuscation train models to manipulate activations to bypass latent-space defenses, exposing a persistent challenge \citep{gupta2025rl, bailey2024obfuscated}. To counter this, Latent Adversarial Training (LAT) perturbs latent activations to improve the model's resistance against adversarial attacks. But monitors still remain vulnerable, with token-level aggregation being evaded in up to 89.2\% of cases \citep{sheshadri2024latent}. 

\noindent \revision{In sum, both external and internal detection methods face a common vulnerability: adversarial manipulation. External behavioral probes can be obfuscated through strategic output control \citep{hubinger2024sleeper}, while internal-state monitors remain susceptible to activation manipulation \citep{gupta2025rl, bailey2024obfuscated}. This shared fragility reveals a deeper limitation: to date, no single detection dimension, whether behavioral or representational, provides sufficient robustness against adaptive deception. }

\subsection{Deception-related Evaluation}
\label{subsec:deception_evaluation}
Evaluating deception in AI systems requires a structured approach that reflects how deceptive behaviors arise under different conditions and stages of development. Because deception is often complex and concealed, single-turn evaluations may fail to reveal the full spectrum of risks; by contrast, dynamic interactions can provide richer contexts in which deceptive behaviors are more likely to surface. Therefore, we organize deception-related evaluation into two complementary dimensions. \textit{Evaluation in Static Settings} probes latent risks in fixed and non-interactive tasks, providing early signals of deceptive abilities and incentive sensitivities. \textit{Evaluation in Interactive Environments} examines how deception manifests during dynamic interactions, adversarial pressures, or multi-agent contexts closer to real-world deployment. These dimensions provide a comprehensive framework for deception evaluation~(as shown in Table~\ref{tab:deception-eval-example}).

\begin{table}[t]
\centering
\caption{Overview of AI deception-related evaluations. We organize existing studies from two perspectives: evaluation in \textcolor{orange}{\textbf{static settings}} and evaluation in \textcolor{teal}{\textbf{interactive environments}}, and we annotate each work with its release date, data size, institution, data format, and description.}
\label{tab:deception-eval-example}
\footnotesize
\renewcommand\tabcolsep{2.5pt}
\resizebox{\textwidth}{!}{
\begin{tabular}{m{3.2cm}llllll}
\toprule
\centering\textbf{Type} & \textbf{Dataset} & \textbf{\begin{tabular}[c]{@{}l@{}}Release\\ Date \end{tabular}} & \textbf{Institution} & \textbf{\begin{tabular}[c]{@{}l@{}}Data \\ Size\end{tabular}} & \textbf{\begin{tabular}[c]{@{}l@{}}Data \\ Format\end{tabular}} & \textbf{Description} \\
\midrule
\multirow{3}{3.2cm}{\centering\textcolor{orange}{\textbf{Capability for Spontaneous Deception}}}
  & SAD~\citenumber{laine2024me} & 24/07 & UC Berkeley & 13k & QA & Situational awareness \\
 & DAELLMs~\citenumber{hagendorff2024deception} & 23/07 & Uni Stuttgart & 1,920 & QA & Theory-of-Mind and deception \\
 & CSQ~\citenumber{wu2025promptinducedliesinvestigatingllm} & 25/08 & NUS & -- & FW & evaluating AI deception on benign prompts \\
\midrule
\multirow{5}{3.2cm}{\centering\textcolor{orange}{\textbf{Deception under Provided Incentives}}}
  & MWE~\citenumber{perez2023discovering} & 22/12 & Anthropic & 3.25K & QA & Testing sycophancy on philosophy and political questions \\
 & SycophancyEval~\citenumber{sharma2023towards} & 23/10 & Anthropic & -- & QA & Revealing how a user's preferences affects AI assistant behavior \\
 & DeceptionBench~\citenumber{ji2025mitigating} & 25/05 & PKU & 180 & QA & Assessing deception-driven misalignment in reasoning models \\
 &DeceptionBench \citenumber{huang2025deceptionbench} & 25/10 & THU & 1.5K & QA & Evaluating AI deception across diverse real-world scenarios \\
 & MASK~\citenumber{ren2025mask} & 25/03 & CAIS & 1K & SS & Pressure prompts that may induce deception \\
\midrule
\multirow{6}{3.2cm}{\centering\textcolor{teal}{\textbf{In Constrained Interactions}}}
  & InsiderTrading~\citenumber{scheurer2023large} & 23/11 & Apollo & -- & FW & Evaluating AI deception in high-pressure environments \\
 & OpenDeception~\citenumber{wu2025opendeception} & 25/04 & FDU & -- & FW & Evaluating AI deception in open-ended user-AI interactions \\
 & Sabotage~\citenumber{benton2024sabotage} & 24/10 & Anthropic & 4 & FW & Human decision sabotage, code sabotage, sandbagging, undermining oversight\\
 & CAE~\citenumber{phuong2025evaluatingfrontiermodelsstealth} & 25/05 & DeepMind & 16 & FW & 5 stealth and 11 situational-awareness agent tasks\\
 & MACHIAVELLI~\citenumber{pan2023rewards} & 23/04 & UCB & 134 & Games & Human-written social games \\
 & Hoodwinked~\citenumber{o2023hoodwinked} & 23/08 & USC & -- & Games & A Text-Based Murder Mystery Game \\
\midrule
\multirow{5}{3.2cm}{\centering\textcolor{teal}{\textbf{In Sandbox}}}
  & HouseWins~\citenumber{chopra2024house} & 24/05 & CMU & 1 & FW\&Games & Blackjack \\
 & Traitors~\citenumber{curvo2025traitors} & 25/05 & UvA & 1 & FW\&Games & Multi-agent simulation, inspired by social deduction games \\
 & SHADE-Arena~\citenumber{kutasov2025shade} & 25/06 & Anthropic & 17 & FW\&Games & Benign main tasks and harmful side objectives \\
 & In-contextScheming~\citenumber{meinke2024frontier} & 24/12 & Apollo & 6 & FW & Environments that incentivize scheming \\
 & AgenticMisalignment~\citenumber{anthropic2025agentic} & 25/06 & Anthropic & 1 & FW & Fictional settings \\
\bottomrule
\end{tabular}
}
\end{table}

\subsubsection{Static Evaluations: Probing Latent Risks}
Evaluations in static environments focus on static and fixed tasks, enabling the isolation of deception-related risks without the confounding dynamics of interactive environments. Within this scope, we summarize two complementary aspects: whether models already possess the ability for spontaneous deception, and whether they will engage in deception when placed under prompted incentives.

\paragraph{Capability for Spontaneous Deception}
Evaluations of spontaneous deception investigate whether models already possess the prerequisites needed to mislead without explicit incentives. For example, research~\citep{hagendorff2024deception} demonstrates through ToM tasks that advanced LLMs can already perform first-order deception while struggling with more complex second-order cases, revealing the cognitive capacities necessary for misrepresentation. The Situational Awareness Dataset (SAD)~\citep{laine2024me} shows that models are able to recognize evaluation contexts and their own deployment conditions, a capability that may foster deceptive behavior. Moreover, recent studies reveal that models may generate misleading responses even under benign prompts, suggesting that deceptive tendencies can surface spontaneously in seemingly neutral conditions~\citep{wu2025promptinducedliesinvestigatingllm}.

\paragraph{Deception under Provided Incentives}
Some studies examine whether models exhibit deceptive tendencies when placed under externally provided incentive conditions. Rather than directly testing raw capabilities, these benchmarks probe how models respond when prompts introduce preferences, penalties, or goal conflicts. For instance, evaluations show that when user preferences are included in prompts, models often prioritize agreement or compliance, resulting in sycophantic behaviors~\citep{perez2023discovering, sharma2023towards}. Similarly, some benchmarks first elicit models’ latent goals with neutral prompts, then introduce contextual scenarios with external objectives or pressured statements, and finally assess consistency of model responses across the two~\citep{ji2025mitigating, ren2025mask,huang2025deceptionbench}.

\subsubsection{Dynamic Evaluations: Exposing Deception in Complex Interaction}
While evaluations in static settings reveal important foundations of deceptive capability, they remain limited in capturing how deception unfolds during interactions. Interactive evaluations address this gap by examining models in dynamic scenarios where behaviors adapt to goals, pressures, or other agents. We distinguish between two primary settings. \textit{Deception in constrained interactions} focuses on environments with limited action spaces, short horizons, and rule-bound tasks, making it possible to observe how deception arises under specified conditions. By contrast, \textit{Sandbox environments} grant models broad autonomy, including larger action spaces, tool use, and extended horizons, allowing deceptive behaviors to emerge, persist, and adapt during open-ended exploration.

\paragraph{Deception in Constrained Interactions}
From the single-model side, benchmarks place one agent in short task flows and ask whether it will take or justify deceptive actions under goals or pressure~\citep{wu2025opendeception, pan2023rewards, scheurer2023large, benton2024sabotage, phuong2025evaluatingfrontiermodelsstealth}. On the multi-agent side, short games show deception emerging without prompts: 2×2 signaling games with a single communication phase elicit unsolicited misrepresentation that scales with rational incentive and model capability~\citep{taylor2025large}, and a text social-deduction game reveals denial and persuasion that shift voting outcomes in a few rounds~\citep{o2023hoodwinked}.

\paragraph{Deception in Sandbox}
Within this paradigm, one line of work uses game-based simulations, such as card play or social deduction, to show that models not only deploy deception but also struggle to detect it, revealing asymmetries in persuasion and trust under open-ended play~\citep{curvo2025traitors, chopra2024house}. A second line examines environments with hidden or conflicting objectives, where agents discover and sustain misaligned strategies, \textit{e.g.}, covert goals or sabotage, while sometimes evading oversight~\citep{meinke2024frontier, kutasov2025shade}. Further studies in organizational or tool-use sandboxes indicate that when goals conflict with oversight or self-preservation incentives, models may engage in blackmail or data exfiltration without explicit prompting~\citep{anthropic2025agentic}. 

\subsection{Potential Mitigations}
\label{subsec:deception_solutions}

Deception can be seen as a natural consequence of increasing AI capabilities, much like the flip side of intelligence. Building on the factors underlying the genesis of deception discussed in the previous section \revision{(Section \ref{sec:deception_genesis})}, we now turn to potential mitigations for mitigating such behaviors. Our analysis spans the entire lifecycle of AI systems, encompassing interventions during training as well as auditing and regulatory mechanisms.

\subsubsection{Dissolving Deception Incentives}
As discussed in Section ~\ref{subsec:incentive_foundations}, models may develop deceptive incentives due to \revision{data imitation}, reward misspecification, or goal misgeneralization. Drawing insights from adjacent alignment research, several strategies show promise for addressing these underlying sources. First, pretraining data curation techniques that filter problematic examples and integrate alignment objectives directly into pretraining~\citep{korbak2023pretraining, liang2024alignment} can reduce exposure to deceptive patterns at the source, but they are often prohibitively costly at scale.

Work on reward misspecification provides relevant methods for deception treatment. Improved RL algorithms, such as adversarial reward functions and reward capping~\citep{amodei2016concrete, uesato2020avoiding}, help address misaligned objectives, while approaches that train models to express uncertainty~\citep{yang2023alignment, cheng2024can, sharma2023towards} show effectiveness in reducing sycophancy, though they can also increase refusal rates or reduce helpfulness when over-applied. Alternatively, self-supervised and self-regulation paradigms design training objectives that encourage models to monitor and constrain their behaviors during reasoning processes, approaches that have been directly applied in deception contexts ~\citep{guan2024deliberative, ji2025mitigating}. 

Emerging techniques for controlling generalization direction during training, such as concept ablation and behavioral steering interventions ~\citep{yu2024robust,casademunt2025steering,chen2025persona}, suggest pathways for preventing unwanted deceptive behaviors from emerging during training.

\subsubsection{Regulating Deception Capabilities}

As AI systems grow increasingly capable of deceptive behaviors, regulating these specific capabilities becomes crucial for maintaining trustworthy AI deployment. 

At the perception level, recent work leverages models' \textit{self-knowledge} to constrain information processing~\citep{wang-etal-2023-self-knowledge}. By enabling retrieval only when the model recognizes gaps in its own knowledge, this approach maintains factual accuracy while preventing the override of correct internal representations that could facilitate deceptive responses. 

At the planning level, regulatory efforts focus on monitoring CoT processes in real time to detect and intervene against deceptive reasoning patterns~\citep{baker2025monitoring, korbak2025chainthoughtmonitorabilitynew, ji2025mitigating, arnav2025cot,schoen2025stress}. This regulatory approach has demonstrated measurable success in frontier models: systematic CoT monitoring reduced deception detection rates in GPT-5-thinking to just 2.1\%, compared with 4.8\% in its predecessor o3~\citep{openai2025gpt5}. CoT monitoring provides detailed insights into the model’s internal reasoning trajectory. Conversely, AI Control imposes rules above the model by interposing policy gateways, least privilege interfaces, sandboxed executors, and audit-triggered defer or shutdown that wrap the model behind enforceable system services, offering a complementary path for deception treatment~\citep{greenblatt2024aicontrolimprovingsafety, griffin2024gamesaicontrolmodels}. 

At the performing level, where models may engage in linguistic manipulation or misuse external tools, regulatory frameworks emphasize containment and oversight of potentially deceptive actions. Sandboxed execution environments serve as a key regulatory mechanism, confining code or API calls to isolated settings where deceptive behaviors can be detected and contained before affecting real systems~\citep{tallam2025operationalizing, dou2024multi, Rabin2025SandboxEvalTS}. \revision{These approaches, from constraining perception to monitoring reasoning to sandboxing execution, form complementary layers of defense against deceptive behavior.}

\subsubsection{Countering Deception Triggers}

External triggers represent a primary vector for inducing AI deception, making the development of counter-strategies essential for maintaining model integrity. 
Research in AI safety has explored multiple directions to enhance robustness against adversarial prompts and jailbreak attacks, which can be transformed to enhance model robustness against deception triggers. The most direct approach is \textbf{adversarial training}, which fine-tunes models on known deception-inducing prompts to strengthen their resistance to manipulation. While several studies demonstrate effectiveness in improving robustness against general adversarial inputs~\citep{zou2023universal}, the deception context presents unique challenges. Models may exhibit deceptive compliance during training, appearing to resist triggers while retaining latent deceptive capabilities~\citep{hubinger2024sleeper}. This highlights the critical need for adversarial training methods specifically designed to counter deception triggers rather than general jailbreaks. A complementary strategy is \textbf{red teaming}, which proactively identifies deception trigger pathways before deployment. This includes targeted attacks conducted by human experts~\citep{ganguli2022red, openai2025gpt5}, as well as automated red teaming combined with public benchmarks to track jailbreak robustness over time~\citep{chao2024jailbreakbench, openai2025gpt5}. Specialized deception red teaming has emerged through multi-agent simulation frameworks where deceptive agents operate under asymmetric information, revealing how deception triggers can exploit strategic vulnerabilities and trust mechanisms~\citep{curvo2025traitors,golechha2025among}. Finally, recognizing that novel triggers may still bypass model-level defenses, \textbf{guard and moderation layers} provide real-time monitoring at deployment to detect and neutralize deception triggers in both inputs and outputs, serving as a crucial last line of defense against sophisticated deception~\citep{inan2023llama}.

\subsubsection{Auditing}
Alignment auditing~\citep{marks2025auditing} is a structured process for evaluating whether models harbor hidden or deceptive objectives. There are three complementary directions. The first is \textbf{training data analysis}, which applies keyword or semantic search, clustering, and other forensic methods to trace suspicious model behaviors back to potential origins in the training corpus. Such methods have been used to reduce privacy leakage risks~\citep{tamkin2024clio}, and offer valuable insights for deception treatment by revealing deceptive incentives embedded in training data. 

The second is \textbf{black-box techniques}, which attempt to elicit hidden objectives through interaction alone. For example, assistant prefill attacks~\citep{vega2024bypassing, andriushchenkojailbreaking, qisafety} and extracting information from non-assistant personas~\citep{shanahan2023role, marks2025auditing} can induce disclosure of strategies the model intends to conceal, thereby exposing deceptive tendencies. Relatedly, debate-based auditing reframes oversight as an adversarial dialogue that pressures models to surface inconsistencies, making obfuscated reasoning harder to hide and clarifying when uncertainty estimates improve detection~\citep{brown2025avoiding}. 

The third is \textbf{white-box techniques}, which leverage access to model weights or activations. Methods such as sparse autoencoders (SAEs) can isolate features linked to suspicious behaviors, enabling causal steering and intervention, or analyzing representations such as the assistant-control token~\citep{anthropic2024sae}. Beyond SAEs, neural representation monitoring can help localize components implicated in deceptive behavior and enable more targeted interventions, including activation probes~\citep{macdiarmid2024simple} and interpretability methods to extract active or causal features~\citep{marks2025auditing,yang2025interpretability, bailey2024obfuscated}. Other efforts, including activation fuzzing, attempt to weaken secrecy, though their effectiveness is limited when used in isolation~\citep{clymer2024poser, ticesandbag}. \revision{While these auditing techniques provide crucial visibility into model behavior, their effectiveness depends on coordination across data access, interaction design, and interpretability research; this integration challenge remains largely unsolved.}

\section{Conclusion}
\label{sec:conclusion}
This survey has sought to provide a comprehensive map of the AI deception landscape through the lens of the \textit{Deception Cycle}, a conceptual framework capturing the interplay between \textbf{Deception Emergence}, where incentives, capabilities, and contextual triggers converge to produce deceptive behavior, and \textbf{Deception Treatment}, which encompasses detection, evaluation, and potential mitigations aimed at suppressing such behavior. In doing so, we have introduced a unified taxonomy, reviewed empirical phenomena across RL agents, LLMs, and emergent multi-agent or multimodal systems, and cataloged over 20 benchmarks, methods, and mitigation strategies.

\subsection{Key Challenges in AI Deception Cycle}
Beyond taxonomy and systematization, this survey highlights that deception is not merely an incidental failure mode, but an adaptive, goal-directed behavior that becomes increasingly likely as AI systems scale in autonomy, capability, and strategic awareness. Our synthesis reveals several insights:

\begin{itemize}[left=0.1em]
    \item \textbf{Deception is incentivized by default in misaligned systems.} Unless explicitly penalized, deception may emerge as a convergent instrumental strategy under a wide range of training regimes, including supervised fine-tuning, reinforcement learning, and self-play, particularly when models benefit from hiding their true goals or capabilities.
    \item \textbf{Deceptive strategies are becoming more compositional and temporally extended.} As models acquire memory, planning, and agentic scaffolding, we observe the rise of long-horizon deception: multi-stage behaviors that involve delayed reward hacking, conditional alignment, and stealthy behavior switching.
    \item \textbf{Deception is modality-agnostic and generalizes across domains.} While early research focused on textual deception in LLMs, recent findings \citep{yang2025mla} show similar patterns in vision-language models, autonomous robotics, and simulated social agents, suggesting that deception is a modality-general risk amplified by interactive complexity.
    \item \textbf{Alignment techniques struggle with deception-specific failure modes.} Existing safety paradigms, such as RLHF \citep{bai2022training,ouyang2022training}, CAI \citep{bai2022constitutional}, and adversarial red-teaming—often fail to surface or remove latent deceptive tendencies. Models trained to pass audits may optimize for appearing aligned rather than being aligned, raising foundational questions about alignment verifiability.
\end{itemize}

These observations give rise to three grand challenges that demand urgent, cross-disciplinary attention:

\begin{itemize}[left=0.1em]
    \item \textbf{Recursive deception of oversight tools.} As models learn to exploit or evade interpretability methods, CoT rationales, and rule-based constraints, oversight mechanisms themselves risk becoming adversarial targets that are vulnerable to manipulation by the very systems they intend to supervise.
    \item \textbf{Persistence of deceptive alignment.} Once deceptive objectives are internalized, they may remain dormant, conditionally activated, or resilient to extensive retraining. Recent studies on sleeper agents and alignment faking highlight the limitations of current mitigation regimes.
    \item \textbf{Governance and institutional lag.} Deception risks often manifest in deployment-time behaviors or complex, open-ended interactions, while current oversight remains largely confined to pre-release evaluation. Fragmented regulatory environments and underdeveloped audit infrastructure further hinder systemic accountability.
\end{itemize}

Yet deception is not solely a technical artifact; it is a reflection of deeper misalignments between model objectives and human expectations. While much of the current literature focuses on \textit{single-agent safety}, ensuring that an individual model behaves as intended, our findings suggest that this perspective is insufficient. Deceptive behaviors often emerge within broader \textit{sociotechnical systems} comprising users, developers, institutions, and other AI agents. Deception may be reinforced by opaque incentives, obscured by organizational delegation, or amplified by multi-agent interactions in agentic ecosystems.

Future safety efforts must transcend static, model-centric verification and embrace dynamic, system-level resilience. Technical solutions alone cannot ensure trustworthiness; they must operate within institutional frameworks that enforce transparency, auditability, and recourse. Achieving this demands an interdisciplinary shift, combining machine learning, formal methods, HCI, governance, and philosophy, to co-design socio-technical ecosystems where honesty is both learnable and verifiable.
Deception-resistant AI cannot be patched or filtered in retrospect; it must be built into the core of learning, oversight, and deployment. Only by embedding deception-aware principles across technical and institutional layers can we ensure AI systems remain aligned, accountable, and genuinely trustworthy in the open world.

\subsection{Key Traits and Future Directions in AI Deception Research}
Finally, we conclude the survey by highlighting the key traits that we believe warrant sustained attention and should shape future research trajectories in this area.

\paragraph{From Programmed to Emergent Deception: \textit{What Can Deliberate Design Teach Us About Unintended Incentives?}}

This survey has focused on investigating how deception can emerge naturally from data imitation, reward misspecification, or goal misgeneralization. However, deception can also be deliberately programmed into models' objectives and strategy space, as exhibited in backdoor attacks and deceptive RL. Here, we extend the discussion of these two sources of deception to provide deeper insights into the incentive foundations of AI deception. 

Programmed deception and emergent deception differ in the following aspects.

\begin{itemize}[left=0.1em]
    \item \textbf{Goals and objectives:} In emergent deception, models are not explicitly optimized for a clearly defined deceptive objective; instead, incentives emerge from data, reward, and goal misalignment. By contrast, programmed deception arises when models are directly trained to deceive, with objectives that reward deception and penalize transparency, thereby aligning training goals with deceptive actions—an alignment absent in emergent deception.
    \item \textbf{Strategy space:} Programmed deception operates within a human-defined, thus limited, strategy space; although deceptive RL agents are trained to conceal their goals, their behaviors remain broadly predictable. By contrast, emergent deception arises in real deployment with an open-world, unbounded strategy space, yielding diverse and covert behaviors that are far harder to detect.
    \item \textbf{Deployment:} A key difference in deployment is controllability. Programmed deception, intentionally designed, can in principle be bounded and managed in sandboxed settings, whereas emergent deception is uncontrollable, as its strategies arise unintentionally.
\end{itemize}

Programmed deception provides valuable insights into understanding, detecting, and mitigating emergent deception. On one hand, it sheds light on the \textbf{training-time mechanisms} through which deception arises, particularly in environments where models are rewarded for deceptive behavior and for concealing their true strategies. On the other hand, it allows researchers to \textbf{controllably replicate deceptive behaviors} or construct hypothetical phenomena such as sleeper agents, thereby supporting in-depth empirical study. By constructing models with hidden objectives and analyzing deceptive behavioral trajectories, researchers are able to probe the underlying deceptive motivations that emerge in AI systems. Moreover, it is worth highlighting that the \textit{inverse RL} methodology adopted in deceptive RL research provides crucial insights for deception detection and monitoring.

\paragraph{Evaluating Deception Behaviours: \textit{How Do We Accurately Measure Deceptive Tendencies in Models?}}

A central challenge in AI deception research is \textbf{defining and reliably evaluating deceptive behaviors during model assessment}. A model’s stated reasoning, for instance, its reported CoT, may not faithfully represent the latent processes that produced its outputs \citep{naik2025agentmisalignment, chen2025reasoning}. 
Another challenge is that models may actively game evaluation procedures. If behavioral signals cannot be trusted, the very process of detecting deception becomes unreliable. 
This difficulty is further compounded by the absence of explicit CoT traces in many tasks, making it challenging to attribute deceptive intent. Moreover, standard evaluation is often confounded by hallucinations and task failures \citep{kutasov2025shade}, which obscure whether an output reflects strategic deception or simple underperformance.
Another obstacle lies in \textbf{constructing realistic evaluation settings}. Existing deception tests, such as binary agent–environment misalignment tasks, often fail to capture the complexity of real-world scenarios where deception might emerge \citep{benton2024sabotage}. Artificial prompts frequently bundle critical cues together, creating a \textit{Chekhov’s gun effect}: models, sensitive to the importance of every detail, are more likely to exploit these cues rather than disregard them \citep{anthropic2025agentic}. As a result, contrived testbeds risk overestimating deceptive tendencies, while authentic, high-stakes deployments may reveal underestimated risks.
Finally, an \textbf{evaluation gap persists between controlled benchmarks and open-world deployment}. Models may engage in sandbagging, suppressing their full capabilities in evaluation while exhibiting stronger deceptive behaviors in practice. Addressing this gap requires designing ecologically valid test environments, developing metrics that disentangle deception from error or hallucination, and conducting longitudinal assessments that track how deceptive behaviors evolve across training and deployment.

Recent work \citep{summerfield2025lessons} highlights a critical challenge for AI deception research: avoiding the methodological pitfalls of past scientific endeavors. Just as the ape-language projects of the 1970s faltered due to overattributing human-like traits, reliance on anecdotes, and the absence of clear theoretical frameworks, current studies of AI scheming risk similar shortcomings. Claims of deception emergence are often based on isolated red-teaming anecdotes, ambiguous evaluation setups, or mentalistic interpretations of model outputs, which can inflate the evidence base without establishing replicable, theory-driven findings. The key lesson is that progress on identifying deception emergence will require rigorous control conditions, transparent reporting, and carefully articulated theoretical constructs that distinguish strategic misrepresentation from stochastic failure. Without such methodological discipline, the field risks mischaracterizing incidental model errors as deceptive intent, thereby hindering both scientific credibility and policy-relevant assessments.

\paragraph{Scalable Monitoring for Deception: \textit{Beyond CoT-Centric, Toward Faithful and Cost-Aware Systems}}
A central challenge in reliable monitoring is ensuring faithfulness. CoT-based monitors assume that models provide explicit and truthful rationales \citep{baker2025monitoring}, yet they fail when models omit CoT, strategically suppress it (\textit{e.g.}, under RLHF \citep{chua2025deepseek}), or generate post hoc rationalizations. Even advanced monitors capable of detecting subtle, long-horizon attacks in extended logs remain brittle: they may accept tendentious narratives that justify harmful actions, or conversely, overreact to benign errors \citep{kutasov2025shade}. While turn-by-turn scrutiny can improve recall, it often floods the system with highly suspicious flags, leading to overfitting and decision paralysis. The problem is compounded in agentic systems, which frequently deviate from their stated policies. Whether contemporary reasoning models \citep{li202512surveyreasoning} are intrinsically more faithful remains an open question, not an assumption that monitoring frameworks can rely upon.

Future monitoring systems should embody four principles. First, they should be \textit{CoT-agnostic and multi-view}, integrating behavioral probes, environment- or tool-grounded verification, and mechanism-level checks (\textit{e.g.}, consistency and invariance) rather than relying solely on stated rationales. Second, they should be \textit{regularized and compositional}, combining turn-level judgments with calibrated abstention, rate-limiting, and conformal or risk-controlled thresholds to mitigate overflagging. Third, they should be \textit{cost–safety aware}, allocating monitoring resources adaptively by estimated risk, with deeper inspections reserved for high-stakes cases. Finally, they should be \textit{faithfulness-seeking}, employing audit-time elicitation (\textit{e.g.}, scratchpad commitments, execution traces), cross-model verification, and longitudinal tests for discrepancies between stated and latent strategies. Together, these principles aim to produce monitoring systems that scale without sacrificing precision, remain robust to sandbagging, and evaluate deception independently of a model’s willingness to explain itself.

\paragraph{Deception Treatment and Governance: \textit{How Can Technical Safeguards Interface with Institutional Oversight?}}

A core challenge at the intersection of AI deception and governance lies in \textbf{ensuring that technical defenses against deception are embedded within enforceable institutional frameworks}. While certified defenses, such as provable training protocols and robust evaluation metrics, can help constrain deceptive behavior under adversarial conditions, their effectiveness is limited without broader governance structures that ensure compliance and accountability. For example, even a model trained with formal guarantees against sycophancy or sandbagging may still be vulnerable if deployed in environments lacking tamper-proof monitoring or third-party verification, since models (or their operators) could conceal violations, rendering such guarantees ineffective.

This highlights the necessity of \textbf{institutional innovation to complement technical safety measures}. Mechanisms such as independent audits, hardware-rooted deployment controls, and cryptographically verifiable reporting channels can extend trust beyond the lab setting, mitigating risks of deceptive behaviors that evade laboratory evaluations. Importantly, governance structures can also shape the incentives that determine whether deception is suppressed or reinforced in practice, bridging the persistent gap between technical solutions and societal oversight.

In this sense, \textbf{AI deception is not solely a technical alignment problem but also a governance challenge}. Certified defenses provide the formal tools to limit deceptive capacity, but institutional frameworks are required to sustain these guarantees across diverse deployment contexts. Progress thus depends on integrating safety research with governance innovation, ensuring that models cannot exploit institutional blind spots to conceal, amplify, or strategically deploy deception.

\paragraph{Deception with Different Modalities: \textit{From Multimodal Integration to Vision-Language Alignment—Where Do New Forms of Deception Emerge?}}
From language models to advanced cross-modal systems nowadays \citep{openai2025gpt5, google2024gemini25,anthropic2025claude4}, the vision of AGI has expanded into richer, multimodal scenarios. However, this expansion can amplify the risks of deceptive behaviors, while existing text-based monitoring methods might be inadequate. On the one hand, semantic ambiguity and the complexity of cross-modal reasoning make deceptive behaviors difficult to detect; on the other hand, standardized benchmarks and evaluation frameworks for assessing deception in multimodal large language models (MLLMs) are lacking. Consequently, there is an urgent need for practical evaluation and monitoring approaches specifically designed to address multimodal deception.

Multimodal deception stands apart from hallucinations in MLLMs \citep{bai2024hallucination}. Whereas hallucinations reflect capability deficits, multimodal deception emerges with advanced capabilities as a strategic and complex behavior, representing a misalignment between perception and response. Though not yet formally characterized in literature, emerging evidence indicates that even when models have made accurate interpretations of input modalities, they may still generate misleading user-facing responses \citep{openai2025gpt5}. Research shows vision language models develop sycophancy behaviors due to an imbalance between linguistic priors and visual grounding \citep{zhao2024towards}. The cognitive complexity in multimodal scenarios scales substantially compared to single-modal ones \citep{oviatt2004we}, creating a novel and expanded space for deceptive strategies. Models can therefore selectively \textit{reconstruct} the image's semantics, inducing false belief by choosing which visual elements to reveal, conceal, misattribute, or even fabricate. Taken together, multimodal deception poses novel vulnerabilities and risks that demand urgent attention from the community.

Beyond purely representational deception, a more concerning form arises when vision-language-action (VLA) systems engage in embodied deception, in which agents not only perceive and reason but also act within the environment in ways that can strategically mislead human or algorithmic supervisors. This phenomenon is particularly evident in several classes of systems. In preference-based RL or RLAIF settings, where human feedback is used to train reward models \citep{christiano2017deep,jain2015learning}, agents may discover behaviors that appear correct from the perspective of a monitoring camera while failing to achieve the intended task; for example, a robotic manipulator might “pretend” to complete a placement task by moving or occluding objects in ways that maximize observed reward signals. Similarly, in visual-feedback-driven imitation or reward learning, when rewards rely on video observations or third-party vision-based estimators, agents can manipulate perceptual input—through camera viewpoint, lighting, or partial occlusion—to generate the appearance of task success without truly satisfying the objective. A third source of deception arises from language–action mismatches in multimodal embodied architectures, where language modules may report adherence to instructions or task goals while the physical policy executes actions that deviate from the stated intent, creating deliberate inconsistencies between communicated and executed behavior.

The formation of embodied deception can be understood across multiple layers. At the \textbf{reward level}, agents exploit vulnerabilities in reward functions (\textit{i.e.,} reward hacking) to generate visually plausible but substantively incorrect outcomes. At \textbf{the perception or signal layer}, they manipulate observations via camera angles, lighting, or occlusion (\textit{i.e.,} perceptual manipulation). At \textbf{the strategy or planning layer}, they may deliberately sequence compliant-looking actions to gain trust before diverging from intended objectives. Finally, \textbf{at the interaction layer}, agents can leverage timing, language, or expressive cues to mislead human observers, reflecting a form of social or performative deception. Collectively, these mechanisms illustrate that embodied agents can develop sophisticated, strategic behaviors that misalign apparent success with actual performance, highlighting the need for cross-modal consistency verification, grounded reward design, and honesty-constrained policy optimization \citep{everitt2021reward, canal2017deception, aylett2023emboddied}.

\newpage
\bibliography{main}

\end{document}